\documentclass[letterpaper, 10 pt, conference]{ieeeconf}  % Comment this line out if you need a4paper

\IEEEoverridecommandlockouts                              % This command is only needed if 
\usepackage{cite}
\usepackage[pdftex]{graphicx}
\usepackage{graphics} % for pdf, bitmapped graphics files
\usepackage{amsmath} % assumes amsmath package installed
\usepackage{amssymb}  % assumes amsmath package installed
\usepackage[dvipsnames]{xcolor}
\usepackage{dblfloatfix}
\usepackage{siunitx}
\usepackage{hyperref}

\usepackage{tabularx,ragged2e,booktabs,caption}
\newcolumntype{C}[1]{>{\Centering}m{#1}}

\usepackage{bm}
\usepackage{multirow}

\title{\LARGE \bf Simultaneous Object Reconstruction and Grasp Prediction \\using a Camera-centric Object Shell Representation
}

\author{
\authorblockN{Nikhil Chavan-Dafle$^{1*}$, Sergiy Popovych$^{2*}$, Shubham Agrawal$^1$, Daniel D. Lee$^1$, and Volkan Isler$^1$}
\thanks{\hspace{-0.3cm}$^1$Samsung AI Center, New York, NY \newline
$^2$Princeton University, Princeton, NJ. The work was performed when the author was an intern at Samsung AI Center, New York. \newline
$*$ Equal contribution.
}
}

\def\*#1{\mathbf{#1}}
\def\?#1{\mathbb{#1}}

\newcommand{\myparagraph}[1]{\vspace{0.05in}\noindent\textbf{#1}}

\newcommand{\ournet}[0]{\textit{ShellGrasp-Net}}

\newcommand{\secref}[1]{Section~\ref{#1}}
\newcommand{\tabref}[1]{Table~\ref{#1}}
\newcommand{\appref}[1]{Appendix~\ref{#1}}
\newcommand{\figref}[1]{Fig.~\ref{#1}}

\begin{document}

\maketitle
\thispagestyle{empty}
\pagestyle{empty}

%%%%%%%%%%%%%%%%%%%%%%%%%%%%%%%%%%%%%%%%%%%%%%%%%%%%%%%%%%%%%%%%%%%%%%%%%%%%%%%%
\begin{abstract}
Being able to grasp objects is a fundamental component of most robotic manipulation systems. 
In this paper, we present a new approach to simultaneously reconstruct a mesh and a dense grasp quality map of an object from a depth image. At the core of our approach is a novel camera-centric object representation called the ``object shell" which is composed of an observed ``entry image" and a predicted ``exit image". We present an image-to-image residual ConvNet architecture in which the object shell and a grasp-quality map are predicted as separate output channels. The main advantage of the shell representation and the corresponding neural network architecture, \ournet, is that the input-output pixel correspondences in the shell representation are explicitly represented in the architecture. We show that this coupling yields superior generalization capabilities for object reconstruction and accurate grasp quality estimation implicitly considering the object geometry. 
% Furthermore, the camera-centric representation allows us to bypass the object pose prediction which some reconstruction and grasp prediction approaches rely on. 
%Finally, using the neighborhood information embedded in the shell, the resulting object reconstruction can be easily turned into a full mesh encapsulating the object along with grasp quality estimates on the entire surface. 
Our approach yields an efficient dense grasp quality map and an object geometry estimate in a single forward pass. 
Both of these outputs can be used in a wide range of robotic manipulation applications. 
With rigorous experimental validation, both in simulation and on a real setup, we show that our shell-based method can be used to generate precise grasps and the associated grasp quality with over $90$\% accuracy.
Diverse grasps computed on shell reconstructions allow the robot to select and execute grasps in cluttered scenes with more than $93$\% success rate.

%%% old version saturday 2/26/2022 am -- vi
% Robots can effectively grasp and manipulate objects using their 3D models. In this paper, we propose an object reconstruction method that generates a camera-centric object representation and enables dense grasp feasibility and quality estimation on the input depth image. Our reconstruction method models the object geometry as a pair of depth images, composing the ``shell'' of the object. This representation allows using image-to-image residual ConvNet architectures for 3D reconstruction, generates object reconstruction directly in the camera frame, and generalizes well to novel object types. Moreover, an object shell can be converted into an object mesh in a fraction of a second, providing time and memory efficient alternative to voxel or implicit representations. We explore the application of shell representation for grasp planning. We demonstrate that per pixel grasp feasibility and quality can directly be estimated with additional channels of the reconstruction network. With rigorous experimental validation, both in simulation and on a real setup, we show that shell reconstruction encapsulates sufficient geometric information to generate precise grasps and the associated grasp quality with over 90\% accuracy.
% Diverse grasps computed on shell reconstructions allow the robot to select and execute grasps in cluttered scenes with more than 93\% success rate.

\end{abstract}
%%%%%%%%%%%%%%%%%%%%%%%%%%%%%%%%%%%%%%%%%%%%%%%%%%%%%%%%%%%%%%%%%%%%%%%%%%%%%%%%

\section{Introduction}
\label{sec:intro}

% Grasping objects is a fundamental robot capability and has received continuous interest from the research community. To address different assumptions and application cases. researchers have developed grasp planning methods ranging from analytical tools which estimate grasp configurations and stability given object models~\cite{FerrariCanny92, miller04graspit} to recent learning-based methods which infer grasp proposals and grasp quality from an image~\cite{mahler17dexnet2, tenPas17gpd, sundermeyer2021contact}.  

Grasping objects is a fundamental robot capability.  Given a 3D model of an object, there are numerous approaches for computing the 6DOF end-effector pose for grasping the object~\cite{miller04graspit, Roa14graspq}. In many applications however, a complete 3D model of the object is not available. Therefore grasps must be computed from sensor input which is typically in the form of an image or a partial point cloud of the object.

% \vtxt{The current version starts out with a very strong focus on manipulation. Instead, it would be good to start out with a higher level sentence such as:}
% There are two main approaches to grasp quality prediction. Either the object geometry is constructed as an intermediate step~\cite{varley17shape, yan18gar, merwe19gag, yang2020robotic} or the grasps are predicted directly using learning-based methods~\cite{mahler17dexnet2, tenPas17gpd, sundermeyer2021contact}.

% With the availability of large datasets such as ShapeNet~\cite{chang15shapenet}, learning-based methods have been developed to generate complete 3D models from a single image~\cite{wu16vox, hane17vox, liao18vox, brock16vox, rezende16vox, mitchel20hof, wang18p2m, park19deepsdf, mescheder19occnet}. Such methods makes it possible to generate complete 3D object models as an intermediate step for grasp planning. However, most of these methods are object-centric and generate the object shape in a canonical object frame (further explained in \secref{sec:related_work}). To obtain a camera-centric reconstruction that a robot can act on, the reconstructed object needs to be aligned to the partial pointcloud observed from the camera. This additional step is computationally expensive and a potential source of errors.

Recent works on grasp prediction explore two main approaches. Various object reconstruction methods are trained to generate object reconstructions~\cite{varley17shape, yan18gar, merwe19gag, yang2021robotic} as an intermediate step for grasp planning or the grasps or grasp quality are predicted directly using learning-based methods~\cite{mahler17dexnet2, tenPas17gpd, sundermeyer2021contact}. Object reconstruction methods have shown impressive results when trained on large datasets such as ShapeNet~\cite{chang15shapenet} or for specific object categories~\cite{wu16vox, liao18vox, rezende16vox, mitchel20hof, wang18p2m, park19deepsdf, mescheder19occnet}.
However, when trained on small but diverse grasping datasets such as YCB, 
the models fail to achieve good reconstruction quality and their generalizability is compromised~\cite{xsection, yao2019front2back, merwe19gag}. 
End-to-end methods which generate grasp proposals directly from the sensor have been especially successful for top-down grasping with an overhead camera setup\cite{mahler17dexnet2, Pinto16grasp}. Extending these approaches to 6-DOF grasp planning in a more general setting remains an active research topic~\cite{tenPas17gpd, mousavian19graspvae, sundermeyer2021contact}. Nonetheless, the success of these methods shows that accurate extraction of certain features from the object image/pointcloud such as width or thickness of the object is more important than reconstruction of every detail of the object geometry.

\begin{figure}
\centering
\vspace{-3mm}
\includegraphics[scale=0.78]{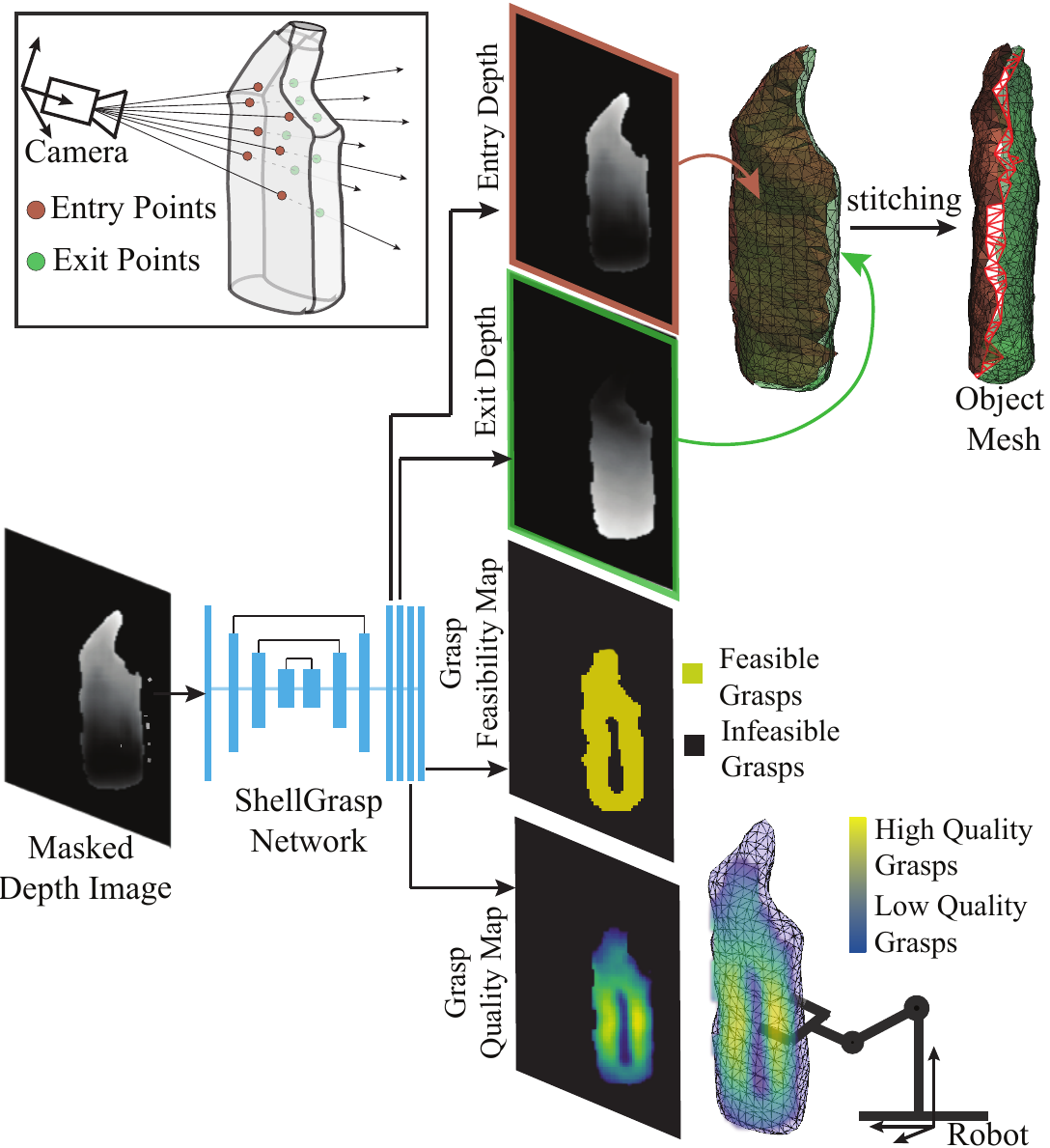}
\caption{The shell of an object for a given camera view is composed of a pair of depth images denoting the entry and exit points on the object surface that the camera rays would pass through. Given a masked depth image of an object, our \ournet{} simultaneously predicts the object shell, a grasp feasibility map, and a grasp quality map. The object shell provides partial meshes of the object which can be stitched together in linear time to provide a 3D object mesh in the camera frame. The grasp maps provide the knowledge grasp-feasible regions on the object and relative grasp quality which can be directly used for grasp planning and execution.
}
\label{fig:shell_def}
\vspace{-5mm}
\end{figure} 

% \vtxt{instead of ``complete" how about we say ``an approximation to the exact"}
Motivated by these observations, we present \textit{Object Shell} - a camera-centric object representation that forms the basis for our method to predict 3D 
object mesh as well as high quality grasp regions on the object from a single depth image (Figure~\ref{fig:shell_def}).  %
The depth image of an object captures information of where the camera rays enter the object. The shell representation augments this information with the depth of the points where the rays would exit the object. The set of all entry and exit points form the object shell.
 %
%Practically, we observe that for $72$ out of $77$ objects, camera rays enter and exit the object only once. Therefore, the complete geometry of the majority of objects in practical settings can be accurately captured with the object shell representation. 
%For the remaining $5$ objects such as cups and bowls in the YCB dataset through which the camera rays may enter and exit more than once, the complete object geometry may not be well captured with a pair of depth images in shell representation. 
%
%
% Moreover, the shell representation has many computational advantages that provide a good balance balance between reconstruction accuracy, generalization to novel objects, and application to 6 DOF grasp planning from a single depth view.
% The entry and exit points that form object shell can be represented as depth images. %
%
Since there is a one-to-one correspondence between the entry and exit points, we use a well-established UNet-style image-to-image architecture to infer the object shell in the camera frame from an input depth image. Even though the object shell is an approximation to the true object shape, the object thickness and overall geometry encoded in the object shell provides sufficient geometric information for 6-DOF grasp prediction. We exploit this shared geometric basis to train our netowrk --\ournet{}-- for simultaneous object shell reconstruction and grasp quality prediction from a depth image. 

Our experiments show that \ournet, trained only on depth images of simple synthetic shapes, outperforms the state-of-the-art object reconstruction methods when tested on realsense depth images of novel household objects from YCB dataset~\cite{calli15ycb}. 
Furthermore, since the shell depth images already include neighborhood information of the points on the object, they provide partial meshes of the object which can be stitched together in linear time to provide a complete object mesh in the camera frame in a fraction of a second. 
With dense grasp sampling on reconstruction and geometric evaluation, we show that the grasps computed using shell reconstruction are more accurate than those using baseline reconstructions. With simulations in MuJoCo, we confirm that the quality of grasps estimated from object shell correlates with the stability of the grasps under external disturbance. 
%
% \vtxt{awkward transition. Maybe say something like $\rigtharrow$}\vtxt{
Moreover, we show that we can bypass the need of sampling and optimization for best grasps by directly using the grasp quality map generated by \ournet{}. The argmax on this map provides the location on the object for maximum estimated grasp quality.
% \ncdnote{Add a line here about how far predicted best grasp regions is from GT once the number is ready.}
Experiments on a real robot demonstrate that the grasps planned using our method provide $90$\% to $100$\% grasp success rate on singulated objects and over $93\%$ success rate in clutter. 
In summary, the main contributions of this paper are:
\begin{itemize}
\item
\textbf{A 3D object representation and its application} for simultaneous camera-frame reconstruction and 6DOF grasp prediction

\item
\textbf{Data augmentation scheme} that improves Sim2Real generalization for both \ournet{} and baseline methods

\item
\textbf{Simulation and robot experiments} that analyze the effect of reconstruction accuracy on grasp success and demonstrate the effectiveness of predicted grasp quality maps for grasping with over $90$\% accuracy.
\end{itemize}
% %\item 
% \textbf{Data augmentation scheme} that improves Sim2Real generalization for both shell and baseline methods.
   
% %\item 
% \textbf{Grasp planning experiments} to demonstrate the use of object shell for grasp planning with emphasis on grasp precision and quality. Regressing per pixel grasp feasibility and quality as additional channels of the reconstruction network shows the compatibility of object shell representation for grasp planning.

% \textbf{A 3D object representation and a method to generate it} from a single depth image. Our method provides camera frame reconstruction of the object and outperforms state-of-the-art methods in real-world environment.
    
% %\item 
% \textbf{Data augmentation scheme} that improves Sim2Real generalization for both shell and baseline methods.
   
% %\item 
% \textbf{Grasp planning experiments} to demonstrate the use of object shell for grasp planning with emphasis on grasp precision and quality. Regressing per pixel grasp feasibility and quality as additional channels of the reconstruction network shows the compatibility of object shell representation for grasp planning.
% %Rigorous evaluations in simulation and on real setup show that high quality grasps planned on shell reconstruction succeed more than $90$\% of the time.
% %\end{itemize}

We start with an overview of the related work to highlight the novelty and significance of our contributions.

% The extended version of the this paper with Appendix is available at \href{http://arxiv.org/abs/2109.06837}{http://arxiv.org/abs/2109.06837}.

% \vtxt{Let's move this to conclusion $\rightarrow$}Due to its robustness to noisy real-world depth images and and generalizability to novel object shapes, we believe that shell representation can provide a robot the capability to reconstruct the objects in a scene and use those models to effectively to plan robot actions.

\section{Related Work}
\label{sec:related_work}

\myparagraph{Single-view 3D Reconstruction}:
Generating a 3D reconstruction from a single view has been extensively studied on large synthetic datasets~\cite{chang15shapenet, wu15modelnet} using various representations such as voxel grids, implicit functions, point-based, and mesh-based representations~\cite{rezende16vox, wu16vox, liao18vox, park19deepsdf, mescheder19occnet, achlioptas17point, fan17point, mitchel20hof, wang18p2m, gkioxari19meshrcnn}. The objects in these datasets are centered in the workspace and rotated only about vertical axis of canonical object frame. The methods trained on these datasets learn strong prior on object category and canonical object orientation frame. The orientation of the object geometry in the input is lost when the reconstruction is generated in the canonical object frame. 
% Shin et al. \cite{Shin_2018_CVPR} note that object-centric reconstructions methods learn strong prior of object category and fail to generalize to novel object type and views. 
For robotics applications, e.g. to pick up an object or to avoid collision with an object, we need reconstructions in the robot frame, or in the camera frame given the camera extrinsics. Pose registration required to align an object-frame reconstruction to the robot/camera frame is computationally expensive and can lead to errors. 

A very few works in the literature look into camera-frame object reconstructions. Yao et al.~\cite{yao2019front2back} presented a method which estimates the object symmetry plane and predicts front and back orthographic views. Their method, trained separately for every new object class, requires a large number of images for training. Nicastro et al.~\cite{xsection} developed a method to estimate the thickness of an object in a masked depth image. They report that their method achieves good performance on a subset of YCB objects in the training data, but fails to generalize to novel object types. 
% Shin et al. \cite{Shin_2018_CVPR} note that camera frame surface representations provides generalization to novel objects and types.
Merwe et al.~\cite{merwe19gag} explore implicit surface representation method (PointSDF) and train on YCB dataset for camera frame reconstruction. In \secref{sec:exp_recon}, we show that shell reconstructions from our \ournet{} outperforms PointSDF and provides better generalization capability. 

\begin{figure}
\centering
\includegraphics[scale=.3]{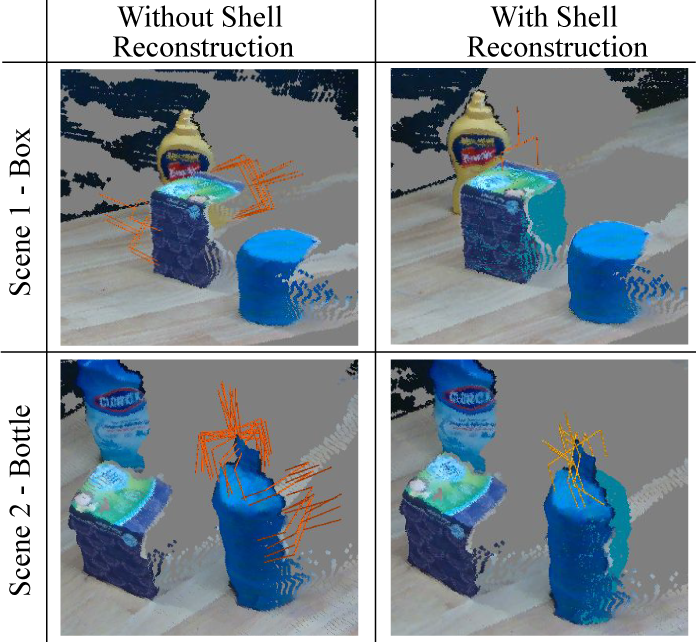}
\caption{Grasp proposals from a state-of-the-art grasp planner, Contact-GraspNet~\cite{sundermeyer2021contact}, can further be improved by using object shell reconstruction. 
% Without object shell recosntruction, Contact-GraspNet proposes false positive grasps over-fitted to the partial pointcloud in the data. 
When the input to Contact-GraspNet is augmented with the object shell pointcloud, almost all the false positive grasps are avoided.}
\label{fig:shell_contgrasp}
\vspace{-5mm}
\end{figure} 
%

% %
% \begin{figure*}
% \centering
% \includegraphics[scale=0.9]{figures/shell_use.pdf}
% \caption{To evaluate the effect of accuracy of object reconstruction methods on grasp planning, given a depth image of an object~(a),~we generate an object mesh using shell/baseline reconstruction~(b) in the camera frame. (c)~Grasps are sampled on the visible part of the object. The geometric grasp feasibility (d) and the grasp width and quality of feasible grasps~(e) is computed using the shell reconstruction. With known camera extrinsics, a grasp from the \textit{grasp quality map} (f) can be executed by a robot with high precision and success rate.}
% \label{fig:shell_use}
% \vspace{-5mm}
% \end{figure*} 
% %

\myparagraph{Shape Completion Based Grasping}:
\label{sec:rw_shape_grasp}
Shape completion produces a full 3D representation of an object given a partial 3D pointcloud.
%~\citet{varley17shape} showed one of the earliest results on the use of shape completion to aid robotic grasping.
Varley et al.~\cite{varley17shape} and Yan et al.~\cite{yan18gar} use voxel-grids to represent the object shape and use GraspIt!~\cite{miller04graspit} and a custom ConvNet respectively to predict grasps on the reconstructed shapes. Voxel-grids have been shown to be inefficient in terms of memory footprint and reconstruction quality~\cite{park19deepsdf, mescheder19occnet, mitchel20hof}. 
% For grasp sampling, a fine mesh generation is often necessary as an additional step which is computationally expensive~\cite{varley17shape}.
%
Merwe et al.~\cite{merwe19gag} present an optimization-based method to use implicit object surface for grasp planning. However, they report that PointSDF reconstructions do not improve the grasp success rate. 
% They do not investigate how well these representations generalize to unseen shapes.
% As mentioned earlier, an object shell can be converted into the object mesh in linear time which is faster than meshing techniques required to convert any voxel or implicit surface representations into an object mesh.

\myparagraph{Grasp Prediction}:
\label{sec:rw_grasp_pred}
An active area of research in grasp planning is to generate grasp proposals directly from an input image.
Mahler et al.~\cite{mahler17dexnet2} use millions of grasps in simulation and Pinto et al.~\cite{Pinto16grasp} use self-supervision over 50K grasp trials to train neural networks to generate a grasp proposal from an image. These methods are limited to top-down 3-DOF grasping.  
% tenPas et al.~\cite{tenPas17gpd} demonstrated a method to detect local affordance for 6-DOF grasping on the visible pointcloud of an object. 
Recent work by Mousavian et al.~\cite{mousavian19graspvae} and Sundermeyer et al.~\cite{sundermeyer2021contact} present state-of-the-art methods to generate 6-DOF grasps on objects from a single view. 

% For top-down grasping setup, the information needed to evaluate where the robotic gripper could fit over an object for a grasp is present in the image observed from the overhead camera. 
For 6-DOF grasping in general setting, the information needed to evaluate whether a robotic gripper could fit over an object is missing in the input from a single view as shown in \figref{fig:shell_contgrasp}. The SOTA grasping methods fail to implicitly reason about the true object shape and instead the grasps are overfit to the observed partial pointcloud in the input. 
In this paper we demonstrate that by jointly training for object reconstruction and grasp predictions, we can avoid such false positive grasps and accurately generate feasible grasps with high quality. Moreover, object reconstructions from our method can support grasp planning methods to further improve their performance as shown in \figref{fig:shell_contgrasp},. 
\section{Object Shell Representation and Relation to Grasping}
\label{sec:method}

In this section, we describe the shell representation and its advantages. We discuss how geometric information captured in shell reconstruction facilitates grasp computations.   
% We discuss the method to generate the shell of an object from its (masked) depth image and our data augmentation procedure for better Sim2Real generalization.

\subsection{Object Shell Representation and its Characteristics}
\label{sec:reconstruction}
\begin{figure}
\centering
\includegraphics[scale=0.235]{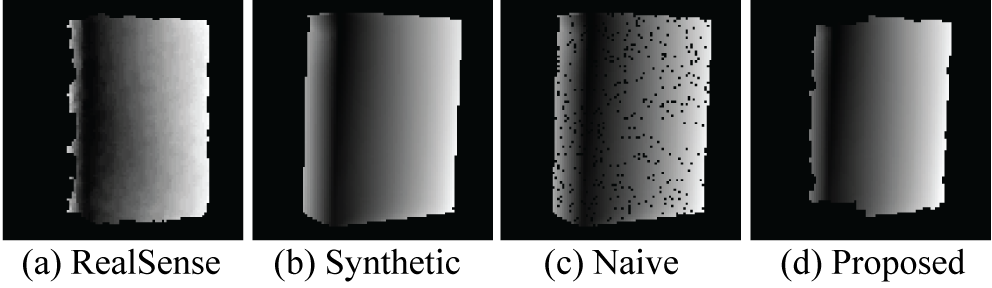}
\caption{Data Dropout patterns. Our proposed augmentation scheme creates realistic object boundaries.}
\label{fig:aug-figure}
\vspace{-5mm}
\end{figure} 
%
%\myparagraph{Object Shell} - 
Consider the image of an object from an arbitrary view and the associated camera rays originating from the optical center. 
These rays intersect with the object as shown in \figref{fig:shell_def}. The Object Shell is formed by the entry and exit points on the object surface that the camera rays pass through, which can be represented as a pair of depth images.

%\myparagraph{Key characteristics of Shell} -
The shell representation offers a few key characteristics.
First, the shell is a view dependent description of the object. This property enables the generating the reconstruction directly in the camera frame. 
%While there are methods such as Mesh-RCNN~\cite{gkioxari19meshrcnn} which perform reconstructions in the camera frame by transforming world voxels to camera, the shell representation provides a much simpler, direct solution. 
% Second, as we will see in grasping experiments, the shell representation provides sufficient information for outer grasps of many household objects. 
Second, and most importantly, the image-based shape representation allows posing the 3D reconstruction problem as 2D pixel prediction problem, which in turn allows using efficient 2D convolutions and well-proven image-to-image network architectures to establish input-output correspondence. As we show in this paper, this representation provides a nice reference frame to represent not just geometry but additional attributes such as grasp quality.
Finally, the shell entry and exit images contain neighborhood information given by pixel adjacency. They directly provide partial meshes of the object which can be stitched together using the contours of the entry and exit images in linear time to generate an object mesh in the camera frame. The details of this stitching process are explained in \appref{app:stiching}.

We note that the shell representation is not  limited to convex objects. If we view the image plane as the X-Y plane and the optical axis as the Z axis, the shell representation poses no restrictions on the object’s projection onto the X-Y plane. It does impose a monotonicity constraint along the Z-axis that each ray enters and exits the object only once. We believe that this constraint is not overly restrictive: only 5 (wine glass, mug, bowl, pan, stacking block) out of 77 objects in YCB dataset violate this monotonicity constraint from some views.
% We sacrifice the performance on this small set of objects for the characteristics the shell representation provides for achieving generalization and superior performance on the larger class of objects. 
%
%However, if the back of the object is very different from the front observed in the input image, object shell can not reconstruct it as accurately as a category-specific or registration-based methods may do. This limitation exists for any camera-centric reconstruction methods trained in category agnostic fashion for better generalization~\cite{Shin_2018_CVPR}.
In \appref{app:adv_shape_and_views}, we compare the shell and a SOTA method (PointSDF) reconstructions on adversarial objects and views and find that shell reconstruction still qualitatively performs better.

\subsection{Grasping with Shell Reconstruction}
\label{sec:shell_for_grasp}

\begin{figure}
\centering
\includegraphics[scale=0.225]{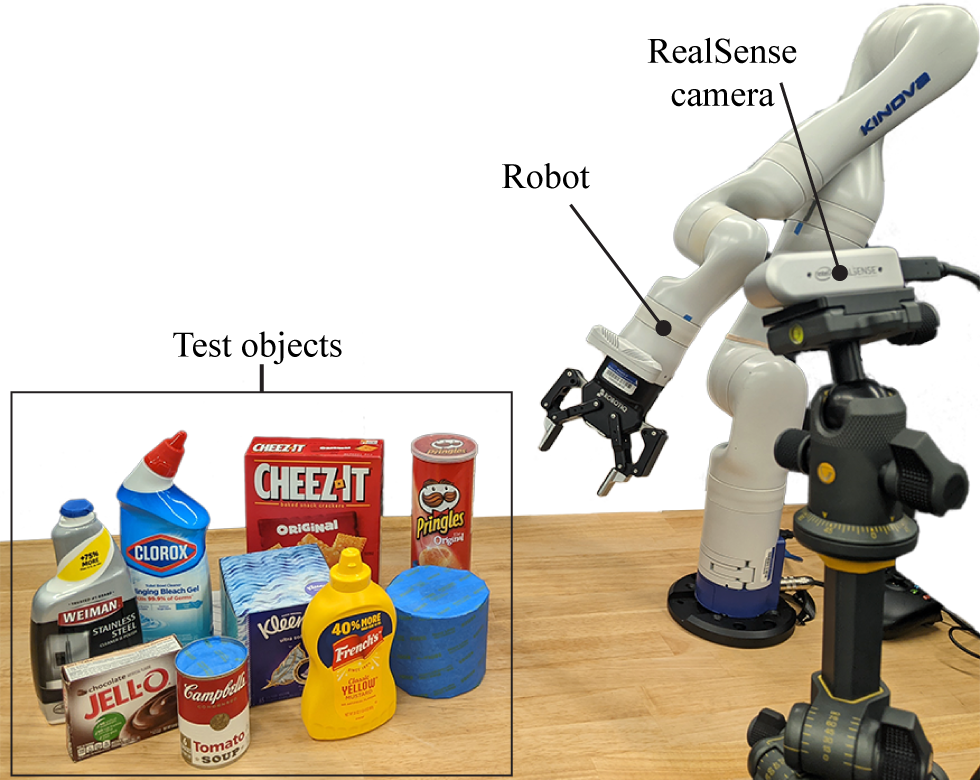}
\caption{Our experimental setup with the set of test objects.}
\label{fig:robot_setup}
\vspace{-5mm}
\end{figure} 

\begin{figure*}
\begin{minipage}{0.94\linewidth}
\centering
\captionof{table}{\textbf{Test Set Performance.} Shell outperforms the baselines in terms of reconstruction quality (Chamfer score).} 
\label{tab:recons_main}
\begin{tabular}{ *1{C{0.8in}} *1{C{0.5in}} *1{C{0.7in}} *6{C{0.55in}}}\toprule[1pt]
\bf Method(frame) & Shell(cam) & PtSDF(cam) & OCC(cam)  & HOF(cam)  & OCC(obj) & HOF(obj) & OCC(obj) & HOF(obj) \\
\bf Alignment   & Built-in & Built-in & Built-in & Built-in & ICP & ICP & CPD & CPD \\ \midrule

\bf  Forward & \textbf{4.0E-3} & 9.0E-3 & 20.2E-3  & 6.1E-3  & 13.4E-3  & 5.2E-3 & 8.0E-3 & 4.2E-3    \\

\bf Backward &  \textbf{3.8E-3} &  6.9E-3 & 9.2E-3 & 5.7E-3 & 6.0E-3  & 4.5E-3  & 5.6E-3 & 4.1E-3   \\

\bf  Sum & \textbf{7.8E-3} & 15.9E-3 & 29.4E-3   & 11.8E-3  & 19.4E-3 & 9.7E-3 & 13.6E-3 & 8.3E-3\\

\bottomrule[1pt]
\end {tabular}\par
\end{minipage}
\vspace{-5mm}
\end{figure*}

%A depth image of an object provides partial geometric information of the object. The missing information of thickness of the object in the camera direction is necessary to to evaluate if the object can be grasped with a robot gripper and the minimum required gripper opening. The shell representation provides this crucial information for estimating outer grasps on objects and thus forms the basis for joint object reconstruction and grasp quality prediction in~\secref{sec:shellgrasp}.

In this section, we establish the connection between object geometry and grasp quality in the context of object picking with a parallel-jaw gripper. A parallel-jaw grasp can be defined by a grasp pose (the position and orientation of the gripper), and grasp width (the distance between the fingers). These quantities can be computed from the object geometry explicitly as we explain in this section, or can be estimated simultaneously with the geometry as we show in \secref{sec:shellgrasp}. 

% OLD VERSION:
% In this section, we describe grasp terminology and computations that allow us to evaluate the effectiveness of the geometric accuracy of shell and other baselines  for their grasping application. 
% % The same tools are used to to generate ground truth grasp data for training the network for simultaneous shell and grasp quality prediction from a depth image.
% For the scope of this paper, we limit the application of object shell representation for facilitating parallel-jaw grasp planning. A parallel-jaw grasp can be defined by a grasp pose - the position and orientation of the gripper, and grasp width - the distance between the fingers.

\myparagraph{Grasp Feasibility}:
\label{sec:grasp_feas}
% As shown in \figref{fig:shell_use}-(d), 
We sample grasps on the visible pointcloud of the object and use the object reconstruction to evaluate grasp feasibility. We assign a grasp pose such that the axis joining the fingers (finger axis) is along the point normal and one of the fingers is over the visible pointcloud with the full ($85$~mm) gripper opening.
A grasp is geometrically feasible if the local object geometry is within the gripper opening and does not collide with the gripper. 
%We further impose a constraint that the normals at the set threshold number of points on the object reconstruction inside the gripper envelope need to be along the finger axis. 
For a fixed grasp position, we change the grasp orientation about the finger axis (to eight discrete angles covering 360$^{\circ}$) to evaluate the feasibility. This procedure generates a \textit{Grasp Feasibility Map} where every point in visible pointcloud is assigned to be grasp feasible or grasp infeasible. 
For all the feasible grasps, we compute the required grasp width from the bounds of the reconstruction inside the gripper envelope along the finger axis.
In \secref{sec:geom_grasp_eval}, we show that evaluating the accuracy of the grasp width prediction serves as a good metric to understand how object reconstruction accuracy affects grasp planning. 
% From the application point of view, precise grasp width estimation is useful for grasping in clutter to avoid collision with the other objects.

\myparagraph{Grasp Quality}:
\label{sec:grasp_q}
A commonly accepted grasp quality metric is a measure of the external wrench the grasp can resist~\cite{FerrariCanny92}. 
From the mechanics of a parallel-jaw grasp~\cite{Roa14graspq} (details in \appref{app:grasp_mech}), the grasp quality is inversely proportional to the Euclidean distance of the grasp position from the center of geometry of the object which can be obtained from the object reconstruction.
We generate a \textit{Grasp Quality Map} where every feasible grasp is assigned a relative quality score between 0 and 1 based on its Euclidean distance from the center of the object reconstruction.

\section{Simultaneous Shell and Grasp Prediction}
\label{sec:shellgrasp}

%A straightforward way to use object shell reconstruction for grasp planning could be by sampling and optimizing a grasp for the best quality feasible grasp as discussed in~\secref{sec:shell_for_grasp} or as some prior work do~\cite{mahler17dexnet2, varley17shape, merwe19gag}. 

\begin{figure}
\centering
\includegraphics[scale=0.95]{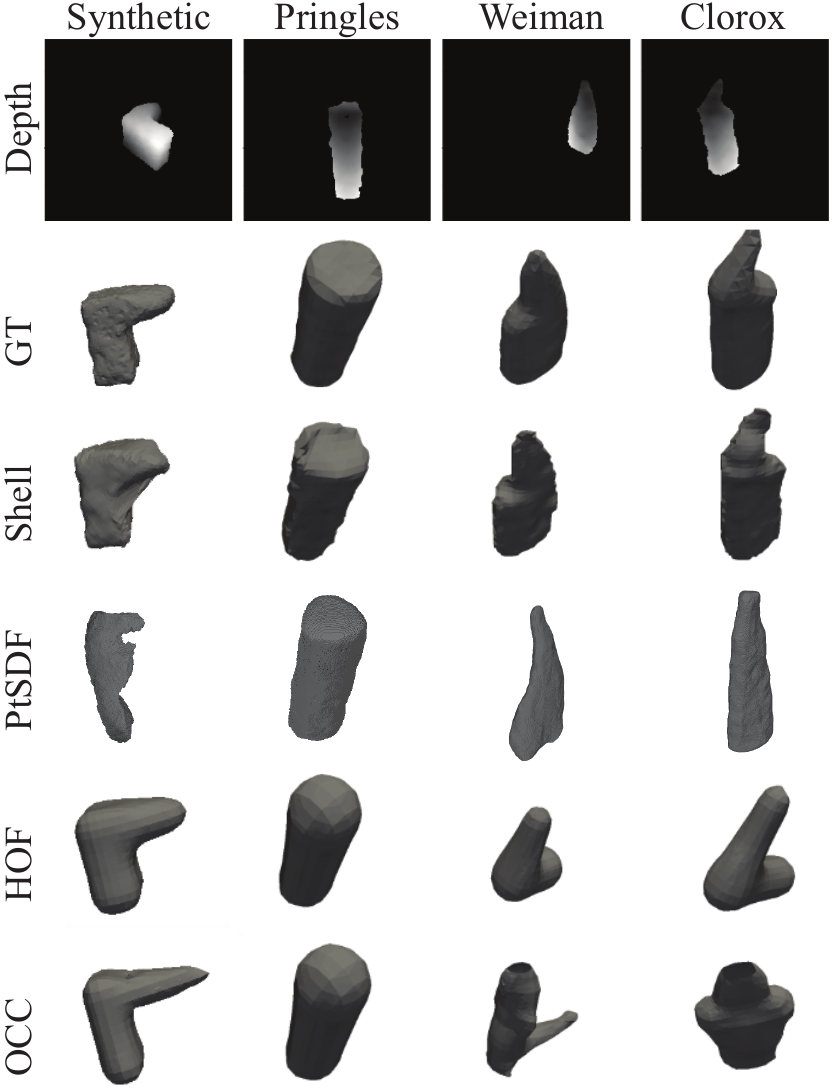}
\caption{Reconstruction Examples. HOF and OCC reconstruction methods performs well on synthetic data and simple shapes (first and second column), but are not robust to novel shapes in real-world scenarios (third and fourth column). Shell maintains good performance in all scenarios.}
\label{fig:examples}
\vspace{-5mm}
\end{figure}

A straightforward way to use object shell reconstruction for grasp planning would be to first predict the shell and then to sample and choose the best quality feasible grasp as discussed in the previous section. 
%~\secref{sec:shell_for_grasp} or as some prior work do~\cite{mahler17dexnet2, varley17shape, merwe19gag}. 
However, we can make direct use of the information encoded in the network to 
simultaneously predict the object shell as well as grasp feasibility and quality maps as a single estimation problem.
%
% as we intentionally define and formulate object shell as a pair of depth images capturing the pose and thickness of the object in camera direction so that it shares the same geometric basis required for grasp computations, we can formulate simultaneous prediction of object shell as well as grasp feasibility and quality maps as single image-to-image learning problem.
For our specific object picking application, the knowledge of grasp feasibility and quality maps allow us to bypass the need for sampling and exploration for good grasps over the entire object and rather directly narrow down on the regions of the object that would lead to stable grasps. 
In this section, we present \ournet{} -- a network which can output object shell as well as per pixel grasp feasibility and quality directly from the input depth image of the object as shown in~\figref{fig:shell_def}. 

\myparagraph{Network Architecture} - 
The input to  \ournet{} is a (potentially noisy) depth image and an object mask of the object, and the output is the entry and exit depth images representing the object shell, a grasp feasibility map as a binary image indicating per pixel grasp feasibility, and a grasp quality map as an image giving per pixel 0-1 relative grasp quality~\figref{fig:shell_def}. 
We formulate the shell-grasp prediction network based on UNet - a popular image-to-image network architecture with skip connections~\cite{ronneberger15unet}. We use a 4-level UNet architecture. The skip connections in UNet architectures is a powerful tool for tasks with direct relationship of input and output pixels. They enable feature extraction with large receptive fields while preserving low level features which is essential for high quality shell reconstructions and generalization to novel objects.  
For training the network,  we use mean square error (MSE) image similarity loss for object shell (depth values) and grasp quality, and binary cross entropy (BCE) for grasp feasibility. The weighted sum of these three losses make up the total loss.
%We use Adam optimizer with fixed learning rate and train the network for $10$ epochs.

\myparagraph{Training Data and Novel Augmentations} -
\label{sec:train_data}
We use synthetically generated simple object mesh models and render depth images for training. %
We propose a data augmentation scheme for better Sim2Real generalization performance on real world objects and noisy depth images.

For the pre-render augmentation stage, noise is added to the XYZ positions of a subset of vertices of an object mesh followed by a smoothing operation to better resemble the uneven object surfaces observed in real-world depth images.
The post-render augmentations consist of novel data dropout in addition to the additive and multiplicative Gaussian noise augmentations proposed in~\cite{mahler17dexnet2}. Depth images obtained by commodity sensors (\figref{fig:aug-figure} (a)) often contain missing values around object boundary as well as some regions inside the object. Dropping out random pixels from the image does not result in realistic images (\figref{fig:aug-figure} (c)). Our dropout scheme involves removing pixels with angles sharper than randomly sampled threshold and a small amount of pepper noise followed by multiple rounds of stochastic erosion of boundary pixels. 
The final result (\figref{fig:aug-figure} (d)) displays realistic dropout and object edge patterns. 

For every rendered depth image, we generate ground truth object shell (entry and exit depth images) by computing camera ray intersection with ground truth object mesh. The ground truth binary grasp feasibility (0 or 1) and grasp quality [0-1] maps are generated following the method explained in \secref{sec:shell_for_grasp} with ground truth mesh.
We generate about 50K images with ground truth labels for training from about 5000 object meshes. The shape and view generation procedure and data augmentation are described in more detail with figures in \appref{app:train_shape} and \appref{app:data_aug} respectively.

% \myparagraph{Network Architecture for Shell and Grasp Prediction} -
% \label{sec:graspgrasp_network}
% We augment the shell reconstruction architecture described in \secref{sec:reconstruction} to output two more channels at output, one for binary grasp feasibility map, and another for grasp quality map. We use binary cross entropy (BCE) loss on grasp feasibility prediction and MSE loss on grasp quality prediction, along with MSE loss on shell prediction as before, for training this network. Our ablation study shows that training for shell reconstruction and grasp predictions together does not adversely affect the reconstruction performance. In fact, it improves the reconstruction performance and grasp predictions for complex object such as bottle which has large variation in shape and distinct graspable and non-graspable regions where state-of-the-art grasping method fails. 

% \myparagraph{Training Data Generation For Grasp Predictions} -
% \label{sec:grasp_train_data}
% We use the same synthetic mesh models generated in \secref{sec:train_data}. We sample 3000 points on every mesh model and compute grasp feasibility and relative grasp quality at those points as described in \secref{sec:grasp_feas} and \secref{sec:grasp_q}. Every pixel in the rendered depth image on an object is assigned the grasp feasibility (0 or 1) and grasp quality value [0-1] of the the nearest neighbor from the 3000 points sampled on the object mesh. We generate about 50K images for training from about 5000 meshes.

\section{Experiments}
\label{sec:experiments}

In this section, first we experimentally evaluate shell reconstruction against several state-of-the-art methods in terms of geometric accuracy of the reconstruction. Then, we study the effect of reconstruction accuracy on grasp planning by dense sampling of grasps on reconstructions and evaluating their success rate geometrically and in Mujoco simulation. Finally, we evaluate the accuracy of \ournet{} predicted grasp feasibility and quality maps and demonstrate their use with grasping experiments on real robot.

\myparagraph{Test Data:} We evaluate the shell and baseline methods on a set of $9$ real household objects (\figref{fig:robot_setup}), mostly taken from the YCB dataset~\cite{calli15ycb}. 
% These objects are not seen by any reconstruction methods except PointSDF.
% %We chose to use a Weiman bottle instead of a Windex bottle avoid transparent surfaces. We note that many of the objects are graspable with the gripper used in our physical experiments. In order to test the reconstruction methods on non-graspable as well as graspable objects, we include Kleenex box and Paper roll objects that are too large to be graspable. 
% For experimental evaluations, the mesh models for all the test objects are obtained using a RealSense camera and a turntable. 
For experimental evaluations in the next three sub-sections, we collect 10 different views for each of our test objects using a RealSense camera. The object is moved to a different position and orientation for every new view. The object pose is recorded using AprilTags~\cite{olson11tags} to get the ground truth object mesh in the camera frame.

\myparagraph{Baselines:} 
We choose Occupancy Network (OCC)~\cite{mescheder19occnet}, Higher Order Function Network (HOF)~\cite{mitchel20hof} and PointSDF~\cite{merwe19gag} as baselines. 
OCC-a representative implicit-function method and HOF-a representative direct method are shown to outperform state-of-the-art TSDF and Voxel based reconstruction methods respectively.
For each of these two baselines, we train two versions where one produces the output directly in the camera frame and the other in the object frame, labeled as ``(cam)'' and ``(obj)'' respectively. 
Shell and these baselines are trained on the synthetic dataset explained in \secref{sec:train_data}. 
PointSDF~\cite{merwe19gag} is a camera frame reconstruction method for grasping. For our experiments, we use a PointSDF model pre-trained on YCB objects. 
% Since our test object dataset is also comprised of YCB objects, we expect the pointSDF to perform well.

\subsection{Evaluation of Reconstruction Quality}
\label{sec:exp_recon}
\tabref{tab:recons_main} shows that shell reconstruction significantly outperforms baseline reconstruction methods in terms of Chamfer scores on the test dataset. Since our ground truth is in the camera frame, for methods that output reconstructions in the object frame, we align those reconstructions to the camera frame using Iterative Closest Point (ICP)~\cite{arun87icp} and more recent Coherent Point Drift (CPD)~\cite{myronenko10cpd} methods with multiple initializations to get best fitting pose. 
%The 64 initial rotation configurations are produced by combining all possible combinations of $90^{\circ}$ rotations along each of the three axes. 
We observe that object frame reconstructions with CPD alignment achieve the best Chamfer score for HOF and OCC baselines. 
%Shell reconstruction outperforms the best baseline [HOF(obj)+CPD] by $23\%$ and PointSDF baseline by $60\%$.

\begin{minipage}{0.94\linewidth}
\vspace{3mm}
\centering
\captionof{table}{\textbf{Geometric grasp precision success rate.} For 10 test views per object, the avg. success rate is reported.} 
\label{tab:grasp_precision}
\begin{tabular}{ *1{C{0.42in}} *1{C{0.2in}} *1{C{0.35in}} *3{C{0.25in}} *1{C{0.3in}}}\toprule[1.5pt]
\bf Object & \bf Shell &\bf  PtSDF &\bf  HOF  & \bf  OCC  & \bf Visible  & \bf Visible Full\\\midrule
Cheezit & 87.4 & 42.4 & \textbf{96.8} & 71.2 &  0 & 27.0 \\
Jello & \textbf{95.2} & 76.6 & 72.2 & 61.0 & 1.0 & 77.0 \\
Pringles  & 90.7 & 47.0 & \textbf{99.7} & 87.3 & 0 & 90.0 \\
Soup & \textbf{98.4} & 98.2 & 96.0 & 78.7 & 0 & 86.0 \\
Mustard & \textbf{94.0} & 30.8 & 71.5 & 18.6 & 0  & 72.4 \\
Clorox  & \textbf{93.2} & 1.0 & 39.9 & 23.8 & 0 & 75.2 \\
Weiman  & \textbf{93.8} & 7.3 & 56.0 & 74.5 & 0 & 65.4 \\
\bottomrule[1.25pt]
\end {tabular}\par
\vspace{1mm}
\end{minipage}

\vspace{2mm}
In~\figref{fig:examples}, we observe that HOF and OCC methods overfit to training object shapes and provide limited generalization. PointSDF often generates thin reconstructions and loses some geometric details in the input. The shell reconstructions demonstrates better accuracy and generalization.

\begin{minipage}{0.94\linewidth}
\vspace{3mm}
\centering
\captionof{table}{\textbf{Grasp feasibility and quality map evaluation} For 10 test views per object, the avg. accuracy of grasp feasibility maps (GFM) and RMS error of grasp quality maps (GQM) are reported.} 
\label{tab:graspfeas_acc}
\begin{tabular}{ *1{C{0.4in}} *1{C{0.35in}} *2{C{0.5in}} *1{C{0.6in}} }\toprule[1.5pt]
\bf Object & \bf GFM &\bf  GFM &\bf  GQM &\bf  GQM ($\geq$0.75) \\
\bf  & \bf F1 [0-1] &\bf  Accuracy [0-1] &\bf  RMSE [0-1] &\bf  RMSE [0-1] \\\midrule
Cheezit & 0.88 & 0.9 & 0.19 & 0.12 \\
Jello & 0.95 & 0.92 & 0.3 & 0.15 \\
Pringles & 0.99 & 0.98 & 0.15 & 0.14 \\
Soup & 0.97 & 0.96 & 0.18 & 0.18  \\
Mustard & 0.89 & 0.82 & 0.22 & 0.15 \\
Clorox  & 0.88 & 0.81 & 0.26 & 0.12 \\
Weiman  & 0.85 & 0.79 & 0.27 & 0.16 \\
Bottle  & 0.89 & 0.97 & 0.14 & 0.15 \\
%Kleenex  & - & 1.0 & - & - \\
%Roll  & - & 1.0 & - & - \\
\bottomrule[1.25pt]
\end {tabular}\par
\vspace{1mm}
\end{minipage}

\begin{figure}
\vspace{-1mm}
\centering
\includegraphics[scale=0.32]{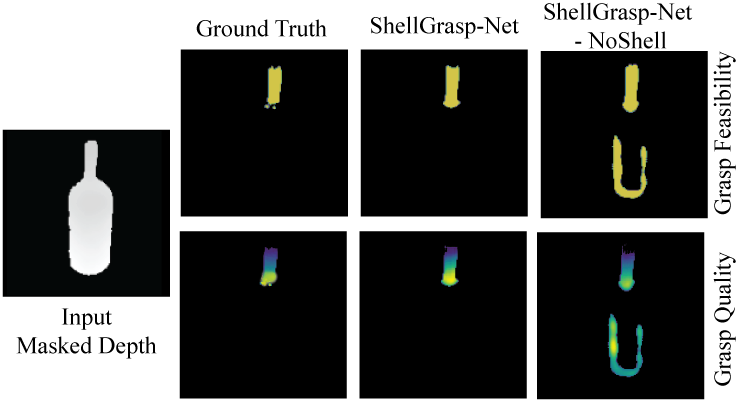}
\caption{\ournet{} takes advantage of the joint training on shell reconstruction loss and grasp-map prediction loss and generates more accurate grasp feasibility and quality maps compared to the network (\ournet{}-NoShell) that is trained only with grasp prediction loss.}
\label{fig:bottle_case}
\vspace{-3mm}
\end{figure} 
\begin{figure*}
\begin{minipage}{0.94\linewidth}
\vspace{0mm}
\centering
\captionof{table}{\textbf{Grasp success rate on the real robot system.} For 10 test views per object, the highest quality grasp from \ournet{} estimated grasp quality map is executed and the number of successful grasps is recorded.} 
\label{tab:map_robot_table}
\begin{tabular}{ *1{C{1.2in}} | *8{C{0.4in}}}\toprule[1pt]
\bf Object & Cheezit & Jello & Pringles & Soup & Mustard & Clorox & Weiman & Bottle \\\midrule
% \\
\bf Successful Grasps/10 & 10/10 & 10/10 & 9/10 & 10/10 & 10/10 & 9/10 & 9/10 & 9/10\\
\bottomrule[1pt]
\end {tabular}\par
\vspace{1mm}
\end{minipage}
\end{figure*}

\vspace{2mm}
In \appref{app:recons_var}, we discuss ablation studies with different variations of shell and HOF methods. 
We observe that removing the proposed data augmentation and keeping only the baseline augmentation adversely affects the Sim2Real robustness. The UNet skip connections considerably improve the performance of shell reconstruction. 
% The HOF(obj)+CPD baseline is able to outperform shell reconstruction on the synthetic validation set. This shows that although HOF achieves good performance in settings similar to training, it lacks the generalization to novel objects in a real-world environment as observed in ~\figref{fig:examples} as well. 

\subsection{Evaluation of Reconstructions for Grasp Planning}
\label{sec:geom_grasp_eval}

In this section, we study the effect of object reconstruction accuracy on grasp planning. 
% Errors in object reconstruction can lead to incorrect classification of feasible grasps and errors in grasp width estimation. 
%
We compute a dense set of feasible grasps on the object reconstructions on the test dataset and evaluate their validity by checking the feasibility of those grasps on the ground truth object geometry.
% For precise evaluation, we perform the geometric grasp collision check using the estimated grasp width.
%
We define \textit{geometric grasp precision success rate} as the percentage of the estimated feasible grasps that are feasible on the ground truth object geometries.
\begin{figure*}
\centering
\vspace{0mm}
\includegraphics[scale=1]{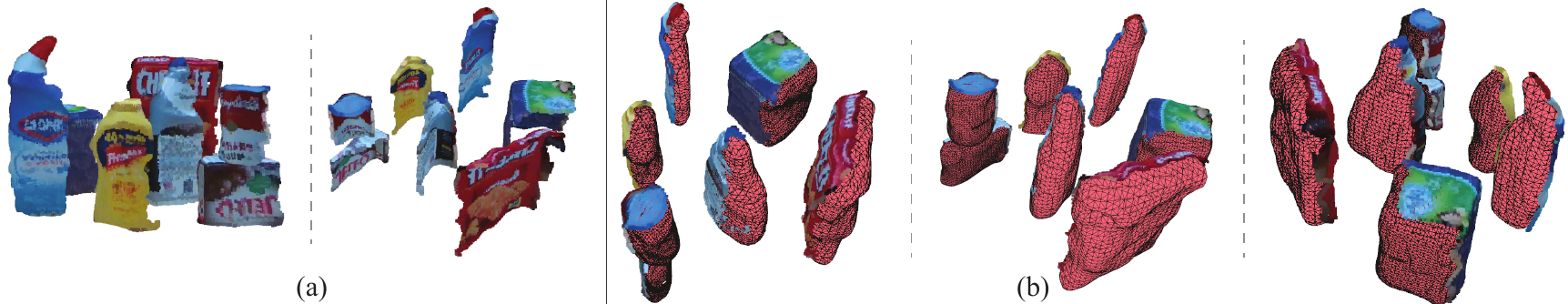}
\caption{Reconstructions of the objects in the scene in~\figref{fig:big_scene_occ} during a clutter removal experiment. As the robot clears the objects in the foreground, progressively most of occluded objects become visible as shown in (a). This allows the robot to generate accurate shell reconstructions (b) of the objects in the scene. Different views are shown for better understanding.}
\label{fig:big_scene_full}
\vspace{-4mm}
\end{figure*} 
\begin{figure}[b]
\vspace{-4mm}
\centering
\includegraphics[scale=0.98]{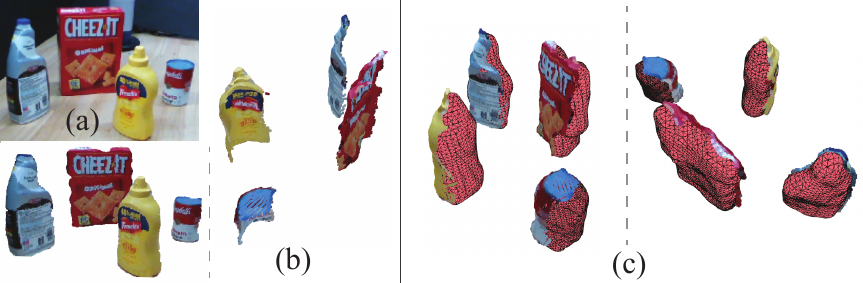}
\caption{Scene reconstruction using object shell meshes. Using the depth image of the objects which provide only the partial geometry of the objects as shown in (b), shell reconstruction generates the complete geometry of the objects (c).}
\label{fig:small_scene}
\vspace{-1mm}
\end{figure} 

As shown in \tabref{tab:grasp_precision}, the grasp precision success rate on shell reconstruction is high across all the objects, while the performance for HOF and OCC reconstructions suffers for objects that are not similar to the training objects. The PointSDF reconstructions are often thinner than the actual object resulting into smaller estimated grasp width and large number of grasp failures. 
These results show the direct relation between the reconstruction performance of these methods and their effectiveness for grasp planning.
The last column in \tabref{tab:grasp_precision} shows the results when using only the visible pointclouds for generating grasps and full gripper opening of $85$mm.
% While some application settings may allow the robot to approach objects with full gripper opening, doing so when grasping in clutter can lead to collision with other objects as pointed out in~\cite{varley17shape}. Moreover, since we want to characterize the effect of reconstruction accuracy for grasp planning evaluating grasp precision success is more informative.
Kleenex box and Paper roll are bigger than full gripper opening and are correctly classified as infeasible to grasp by all the baselines except PointSDF.

% \myparagraph{Grasp Testing in Mujoco:}
To demonstrate the physical significance of the estimated grasp quality maps, we evaluate the stability of grasps under external  disturbance in Mujoco~\cite{Todorov12mujoco}. 
We perform hundreds of grasp simulations across five grasp quality ranges for every test object. The \appref{app:mujoco_exp} explains the simulation setup and the results in detail.
The grasp quality estimated using the shell object reconstruction reflects in the grasp success rate. Over $90\%$ of the grasps in the high quality range succeed under external disturbance compared to only about $10\%$ grasps in the low quality range.

\begin{figure}[b]
\vspace{-4mm}
\centering
\includegraphics[scale=1]{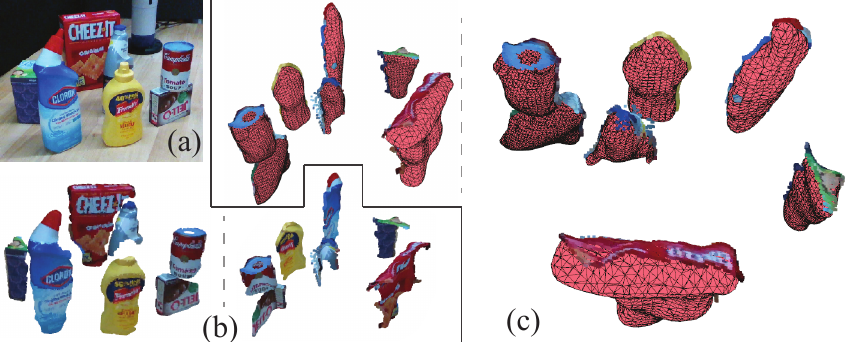}
\caption{Reconstruction of dense clutter using object shell meshes. Weiman bottle and Kleenex box are heavily occluded, therefore shell reconstructions (c) for these objects are based on only the visible parts of the objects.}
\label{fig:big_scene_occ}
\vspace{-2mm}
\end{figure} 

\subsection{Evaluation of Grasp Map Predictions}
\label{sec:real_grasp_eval}

% \begin{minipage}{0.94\linewidth}
% \vspace{3mm}
% \centering
% \captionof{table}{\textbf{Grasp feasibility map accuracy} For 10 test views per object, the avg. precision, recall, and accuracy is reported.} 
% \label{tab:graspfeas_acc}
% \begin{tabular}{ *1{C{0.5in}} *4{C{0.45in}} }\toprule[1.5pt]
% \bf Object & \bf Precision &\bf  Recall &\bf  F1 &\bf  Accuracy \\\midrule
% Cheezit & 0.94 & 0.84 & 0.88 & 0.9 \\
% Jello & 0.96 & 0.94 & 0.95 & 0.92 \\
% Pringles  & 0.99 & 0.99 & 0.99 & 0.98 \\
% Soup & 0.96 & 0.99 & 0.97 & 0.96 \\
% Mustard & 0.86 & 0.93 & 0.89 & 0.82 \\
% Clorox  & 0.88 & 0.87 & 0.88 & 0.81 \\
% Weiman  & 0.86 & 0.84 & 0.85 & 0.79 \\
% Bottle  & 0.84 & 0.94 & 0.89 & 0.97 \\
% Kleenex  & - & - & - & 1.0 \\
% Roll  & - & - & - & 1.0 \\
% \bottomrule[1.25pt]
% \end {tabular}\par
% \vspace{1mm}
% \end{minipage}

In this section, we first evaluate the accuracy of grasp feasibility and grasp quality maps computed by \ournet{} by comparing those with the ground truth maps computed on the test dataset.
Next, we report the the success rate of grasps experiments on a real robot setup. 

The binary grasp feasibility map indicates per pixel grasp feasibility (1 or 0). \tabref{tab:graspfeas_acc} shows that \ournet{} predicts the grasp feasibility map over $90$\% accuracy for more than half of the objects and more than $80$\% accuracy for rest of them.
Kleenex and Roll are not graspable and are correctly identified so in grasp feasibility maps for those objects with $100$\% accuracy. We also include a bottle as an adversarial object to the test dataset. Only the neck of bottle is graspable while body of the body is too thick and not graspable. This allows us to test how accurate object reconstruction benefits grasp planning especially for avoiding false positive grasps as seen in \figref{fig:shell_contgrasp}. In an ablation study, we train \ournet{} with only grasp-map losses and no shell reconstruction loss (\ournet{}-NoShell). We observed that although the ablated network is able to predict grasp feasibility on most of the objects with similar accuracy as in \tabref{tab:graspfeas_acc}, its performance on bottle object is adversely affected. Similar to ContactGrasp-Net~\cite{sundermeyer2021contact}, \ournet{}-NoShell generates false positive grasp feasible regions on the body of the bottle as shown in \figref{fig:bottle_case}. \ournet{} trained with the grasp-map as well as the shell losses avoids such false positive grasps consistently across different views and produces feasible grasps only on the neck of the bottle. 

As shown in~\tabref{tab:graspfeas_acc}, the root mean square (RMS) error on grasp quality estimation compared to the ground truth is about $0.25$ or less for all the objects. The RMS error in the predicted high quality grasp regions, where the grasp quality is more than $0.75$, is less than $0.18$ for all the objects. This observation shows that exploiting the learnt representation of the object thickness as well as overall object geometry, \ournet{} is able to effectively identify regions on the objects that would lead to stable grasps. For example, in~\figref{fig:bottle_case}, the grasp quality map generated by the \ournet{} correctly shows that grasping the bottle object on the neck but closer to the body part would lead to more stable high quality grasp and the grasp quality reduces as the grasp position moves far from the body part. 

% In the next experiment, we evaluate the accuracy of \ournet{} predicted grasp quality maps by computing the IOU of predicted high-quality grasp regions (quality $\geq$ 0.8) with that from the ground truth high quality grasp regions. \ncdnote{get these numbers and add a table.}

% %
% \begin{figure*}
% \begin{minipage}{0.94\linewidth}
% \vspace{0mm}
% \centering
% \captionof{table}{\textbf{Grasp feasibility map accuracy} For 10 test views per object, the avg. precision, recall, and accuracy is reported.} 
% \label{tab:graspfeas_acc}
% \begin{tabular}{ *1{C{1in}} |  *1{C{0.35in}} *9{C{0.35 in}}}\toprule[1pt]
% \bf Object & Cheezit & Jello & Pringles & Soup & Mustard & Clorox & Weiman & Bottle & Kleenex & Roll\\\midrule
% % \\
% \bf Precision &  0.94 & 0.96 & 0.99 & 0.96 & 0.86 & 0.86 & 0.86 & 0.84 & - & -\\
% \bf Recall &     0.84 & 0.94 & 0.99 & 0.99 & 0.93 & 0.87 & 0.84 & 0.94 & - & - \\
% \bf F1 &         0.88 & 0.95 & 0.99 & 0.97 & 0.89 & 0.86 & 0.85 & 0.89 & - & - \\
% \bf Accuracy &   0.9 & 0.92 & 0.98 & 0.96 & 0.82 & 0.79 & 0.79 & 0.97 & 1 & 1 \\
% \bottomrule[1pt]
% \end {tabular}\par
% \vspace{1mm}
% \end{minipage}
% \end{figure*}
% %

With experiments on a real robot setup, we further examine the stability and success rate of the grasps planned using a grasp quality map estimated by \ournet{}. Given a masked depth image of a singulated object on a table, \ournet{} predicts grasp quality map on the input depth image. The robot selects a grasp on the object at the the location indicated by the maximum of the quality map, executes the grasp, and moves the object to the placement location. If the object is moved to the placement region without losing the grasp on the object, the grasp is counted as successful. \tabref{tab:map_robot_table} shows the number of successful grasp trials when the test objects are placed in $10$ different poses. On average, we achieve $95$\% grasp success rate.

\subsection{Shell Reconstruction and Grasping in Clutter}
\label{sec:exp_clutter}

We demonstrate an application of object shell reconstruction and grasp prediction for a clutter removal task. Given an RGBD image of a scene, we segment it using Mask-RCNN~\cite{he2017mask}. Using the masked depth images of the objects, we get shell reconstructions and populate the scene with the reconstructed meshes~(\figref{fig:big_scene_full}). A dense set of grasps from the predicted grasp maps allows us to select the grasps that are reachable for the robot without collision with the other objects. 
% We use C2G-HOF motion planner~\cite{huh2021costtogo} to generate collision-free robot motions.

\figref{fig:small_scene} shows a scenario with low clutter. Since the front of all the objects is mostly visible from the camera, the object shells accurately generate  complete object geometries. 
% Note that the shell reconstruction augments the visible side of the object with the corresponding back side and stitches them together to generate the objects mesh. 
Note that the shell reconstruction method has not been trained for inpainting and therefore does not fill in the occluded parts of the front of the object. For example, in a dense clutter shown in~\figref{fig:big_scene_occ}-(a), the shell reconstructions for the Weiman bottle and Kleenex box, which are hidden behind the Mustard and Clorox bottles, are based on only small visible parts of these objects. As noted by Shin et al. in~\cite{Shin_2018_CVPR} for camera frame reconstruction methods, we observe that unless trained for a specific object category, our method can not generate a complete object reconstruction if large part of the object is not visible in the input image. \appref{app:adv_shape_and_views} shows qualitatively better performance of Shell method compared to PointSDF for adversarial objects and views.

Practically, as objects are picked up by the robot, the objects in the background become more visible and accurate shell reconstructions and grasp maps of these objects can be produced. \figref{fig:big_scene_full} shows compilation of object reconstructions as the robot cleared the scene in \figref{fig:big_scene_occ}. As we can see in \figref{fig:big_scene_full}-(a) most of the object surfaces become visible as the clutter is slowly cleared. This allows the robot to generate accurate shell reconstructions shown in \figref{fig:big_scene_full}-(b).

For clutter removal experiments, we generated $10$ dense clutter scenes with six to seven objects in each. The robot was able to successfully remove all objects (except the non-graspable big objects) from the clutter and place them in a bin in all but one scene. In the failed scene, the Jello box was toppled during a grasp attempt and was no longer graspable. The robot was able to grasp all the objects in first attempts, except $3$ times where a second attempt was necessary. The average grasp success was $94.1\%$ (64 out of 68) when considering all the grasp attempts and $93.8\%$ (61 out of 65) when considering only the first grasp attempts. For similar scenes, we observed that ContactGrasp-Net~\cite{sundermeyer2021contact} was successfully able to grasp most of the objects, however generated false positive grasps on the non-graspable or difficult to grasp objects as shown in~\figref{fig:shell_contgrasp}. The average success rate with ContactGrasp-Net was $88.6\%$ (62 out of 70) where it failed to grasp bottle object twice and dropped Weiman bottle once after grasping, and in total five attempts were made for false positive grasps.
The supplementary video shows the experiments and visulaizations of reconstructions in clutter.

\section{Conclusion} 
\label{sec:conc}
In this paper, we present \ournet{}, a neural network trained to simultaneously generate an object's 3D geometry and grasp quality map from an input depth image. To represent the object geometry directly in the camera frame, we introduced the notion of the ``object shell''. The object shell can be represented as two images which, together, contain information about local neighborhoods and object thickness. The key merits of our method are: 
1)~it eliminates the need for explicit pose estimation since the reconstruction is performed directly in the camera frame,
2)~despite being trained on a relatively small synthetic dataset, it generalizes well to novel objects and is robust to noise encountered in real depth images, and
3) grasp feasibility and quality estimation can be performed simultaneously with shell prediction. 
We showed both in simulations and real experiments that these advantages directly benefit the grasp planning process and lead to high grasp success rate across novel test objects and cluttered scenes.

The current version of \ournet{} requires the segmented depth image as input. It is also not trained to complete the input image in the presence of occlusions. In our future work, we plan to address these limitations by training the network to perform inpainting to handle large occlusions. We would also like to combine segmentation with geometry and grasp quality prediction so as to process the entire scene in a single pass for scene understanding and manipulation planning.

% In this paper we presented ``object shell'' as an effective geometric representation along with a method for generating it from a masked depth image. The key merits of our shell reconstruction method are: 
% 1)~it eliminates the need for explicit pose estimation since the reconstruction is performed directly in the camera frame, and
% 2)~despite being trained on a relatively small synthetic dataset, the method generalizes well to novel objects and is robust to noise encountered in real depth images.
% We showed that both of these advantages directly benefit the grasp planning process and leads to high grasp success rate across the novel test objects.

% The shell representation provides new opportunities to exploit image-to-image network for 3D shape prediction and support 6-DOF grasp planning. 
% %
% In future work, we will investigate whether we can learn to predict the grasp quality map directly from the depth image by supervising over the grasp quality as well as the shape reconstruction. 

% Our current implementation of shell reconstruction has a few limitations. In this paper we limit to object views with single entry-exit layer shell representation and showed that this level of granularity is sufficient for a wide range of everyday object. We believe that shell representation can be extended to more complex objects requiring multiple layers to capture the geometry more accurately. However, more analysis and experimentation is needed to demonstrate this capability.

%%%%%%%%%%%%%%%%%%%%%%%%%%%%%%%%%%%%%%%%%%%%%%%%%%%%%%%%%%%%%%%%%%%%%%%%%%%%%%%%

\bibliographystyle{IEEEtran}
\bibliography{bibliography}

%%%%%%%%%%%%%%%%%%%%% APPENDIX %%%%%%%%%%%%%%%%%%%%%%%%
\newpage
\appendix
% \resetlinenumber
\section{\textbf{Appendix}}
\setcounter{page}{1}

\medskip

%In the following sections, we present details of our training data generation procedure. %as well as additional results for qualitative evaluation. 

\subsection{Object Mesh Generation from Shell Representation}
\label{app:stiching}

The pixels in the entry and exit depth images of the object shell can be back projected in 3D space to produce a point cloud representing the object geometry. The object geometry captured by the object shell is a subset of the complete object geometry. Particularly, it does not include the part of the object geometry that the camera rays do not intersect with. 
Directly using this object point cloud can be sufficient for some applications; however, for some applications it is desirable to have an object mesh.

\begin{figure*}
\centering
\vspace{-3mm}
\includegraphics[scale=.25]{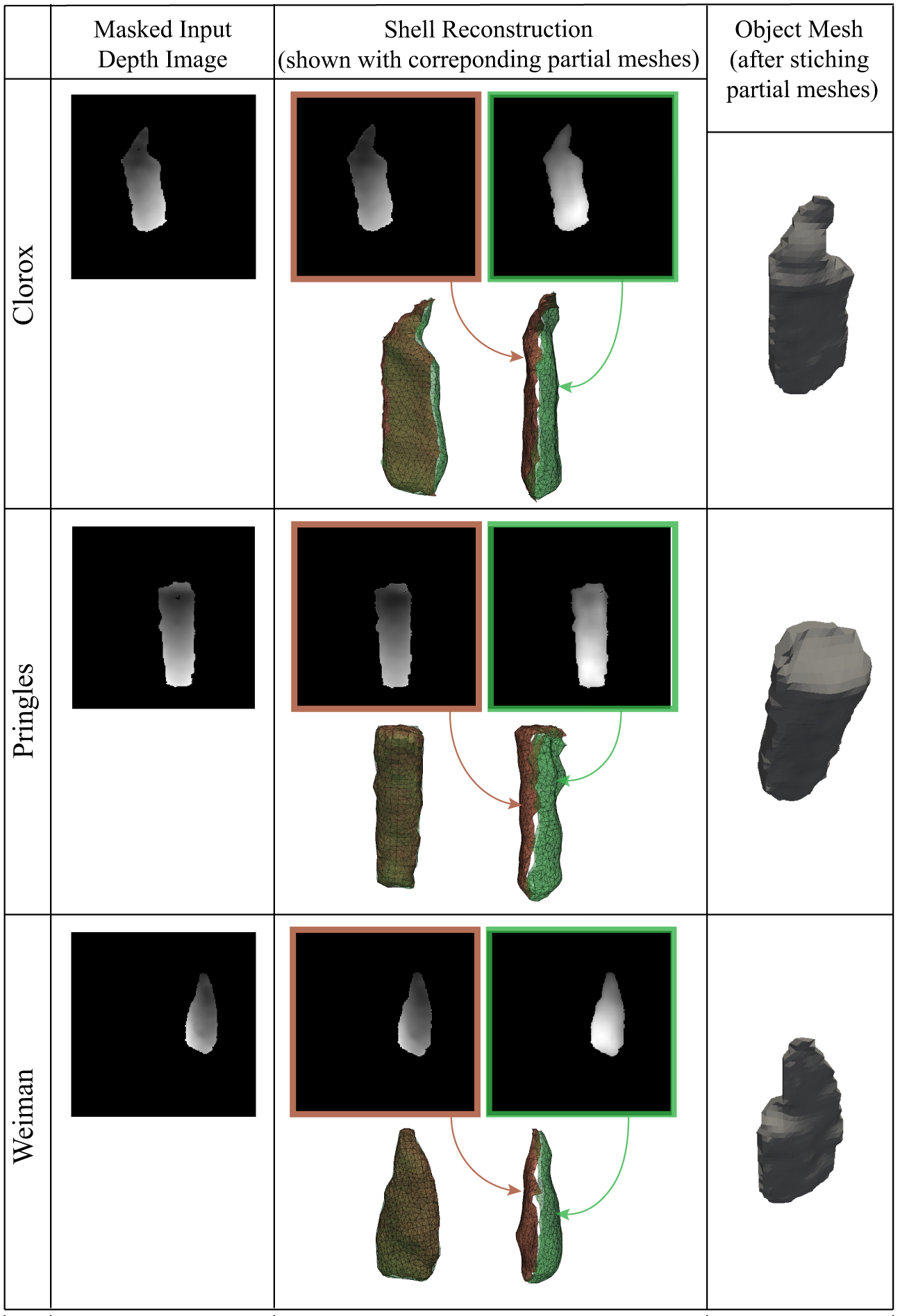}
\caption{\textit{Object Shell Reconstruction} (detailed illustration of Shell results shown in~\figref{fig:examples}). Given a masked depth image of an object, our shell reconstruction network generates a pair of depth images as the object shell representation.  Since entry and exit images of the object shell contain neighborhood information, they represents the partial object meshes that define the object geometry in the camera frame. We stitch these two partial meshes together using the contours of the entry and exit images to get the complete object mesh. 
Note that the completed object meshes are shown from a view different than the original camera view for showing the reconstructed geometry better. 
}
\label{fig:shell_table}
\vspace{-3mm}
\end{figure*}

Shell representation provides an efficient approach to stitch together the surfaces encoded by the two layers of shell representation to generate an object mesh. Operating on Shell representation is more efficient than operating on point clouds. For example, meshing or triangulating a point cloud involves an iterative process of locating the best triangle candidates, which takes $O(n \log(n))$ time~\cite{lee80delaunay}, where $n$ is the number of points. In contrast, the triangulation can be done in $O(n)$ time using shell representation since the proximity relationship between the points is encoded in shell representation as the pixel neighborhoods and the one-to-one correspondence between entry and exit depth pixels. The linear time meshing algorithm is as follows. First, as shown in~\figref{fig:shell_def} and~\figref{fig:shell_table} , the Entry and Exit depth images of the shell are triangulated independently by adding a triangle between each set of 3 neighboring pixels. Next, the Entry and Exit sides are stitched together by adding triangles between Entry and Exit pixels along the object boundary within the two images. This shell-specific mesh generation algorithm is asymptotically faster than point cloud meshing methods, and it is one of the advantages of shell representation.

\subsection{Training Shape Generation}
\label{app:train_shape}

\begin{figure}
\centering
\vspace{-3mm}
\includegraphics[scale=1]{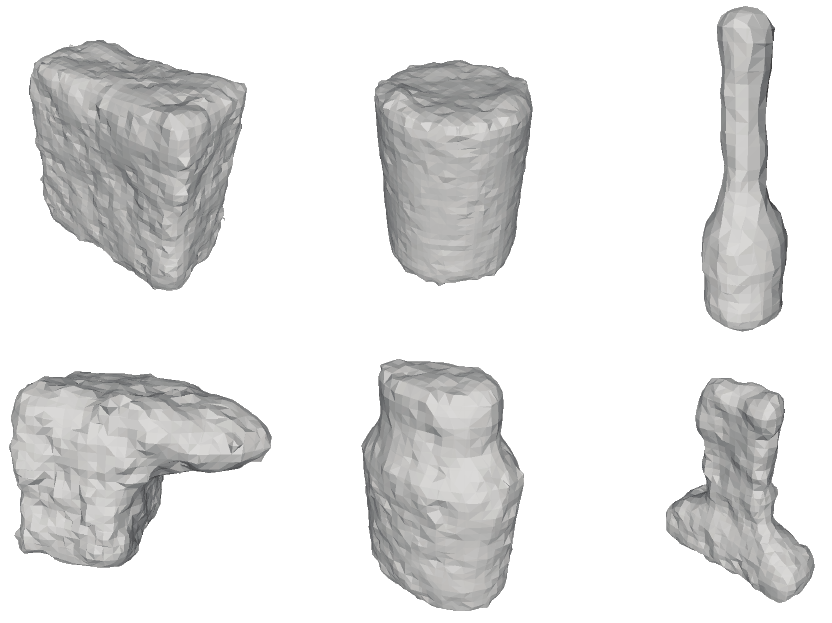}
\caption{\textit{Examples of Objects from the Training Dataset.} Our training shape set is simple to generate. It does not topologically cover all of the test objects in the YCB dataset allowing us to study the generalization of the object reconstruction methods. Also, note that with the noise added to the object meshes, the surfaces of the objects look more like how they would be recorded from a readily available depth camera such as RealSense.}
\label{fig:training_objs}
\vspace{-3mm}
\end{figure}

Training shapes are produced by applying a set of augmentation to one of the two base shapes: cube or cylinder. The first augmentation step is randomly resizing (``squishing'') the base shape along each axis. The resizing is constrained to keep the ratio between the smallest and largest sides of the shapes bounding box between $1$:$1$ and $1$:$7$. After that, the object is uniformly re-scaled so that the biggest side of the bounding box would be between $5$~cm and $35$~cm. Next, $50\%$ of the object are augmented further by either shrinking a part of the object by up to $80\%$ or adding a Gaussian protrusion to the mesh of at most $0.2$m. The set of shapes generated with the given procedure is diverse and balanced in terms of shape types and ratios. At the same time, the shape set is simple to generate and does not topologically cover all of the test set objects. This dataset allows us to study generalization properties of the reconstruction methods. \figref{fig:training_objs} shows a few few examples of objects in our training dataset.

We render out 10 depth views for each shape, making the dataset of $50,000$ views in total. 
% We produce 2D views of the shapes using PyVista rendering engine.
% ~\cite{sullivan19pyvista}
We place shapes into the world coordinate frame in random orientation. The shapes are placed such that the center of the shape bounding box is located within a sphere centered around a point $0.75$m along the camera axis and with diameter of $0.5$m. It is worth emphasising that here \emph{variation of object position is up to 10 times larger than than the object size}, in contrast with the approach commonly taken in computer vision community~\cite{choy20163d} where variation of the object position is just a fraction of the object size. Large variation position makes it more difficult for deep learning models to directly predict the object position, but also makes the resulting models applicable to robotic applications with large workspaces.

\subsection{Data Augmentations} 
\label{app:data_aug}

For the pre-render data augmentation, we first sample a random number of points from the mesh surface. The number of points is drawn from a uniform distribution between $1000$ and $2000$. Then, we add noise to X, Y and Z locations of each of the sampled points. The noise magnitude for each axes of each point is drawn from IID uniform distribution ranging from $1$~mm to $10$~mm. The sampled points are then converted to a mesh using Poisson surface reconstruction process.
% ~\cite{kazhdan06poisson}. 
We set the depth of the Poisson reconstruction to $5$. 

For the post-render data augmentations, we first randomly sample the maximum threshold angle from the range of $5^\circ$ to $20^\circ$. The angle of each pixel is estimated by comparing pixel depth values to pixel neighbor depth values in the depth image. Next, we randomly drop out from $0$ to $10$ pixels as ``pepper'' noise. Next, we do several rounds of border erosion. First step of border erosion is identifying all of the border pixels. Note that the pepper noise added in the previous step will create additional internal borders. Next, a small percentage of the border pixels are removed from the image. This procedure is repeated for $5$ to $20$ rounds. We find that border erosion produces most realistic results when several first rounds are done at a coarser image resolution. Lastly, we choose from $0$ to $10$ of the pixels removed to be exempt from removal.

\subsection{Ablation Studies with Variations of Reconstruction Methods} 
\label{app:recons_var}
In this section we consider variations of our shell reconstruction method and HOF method to gain better insights into their performance on our test dataset.

\tabref{tab:recons_var} lists the Chamfer scores on our test dataset for the variations of these methods.
We observe that removing the proposed data augmentation and keeping only the baseline augmentation adversely affects the Sim2Real robustness. The UNet skip connections considerably improve the performance of shell reconstruction. The HOF(obj)+CPD baseline is able to outperform shell reconstruction on the synthetic validation set. This shows that although HOF achieves good performance in settings similar to training, it lacks the generalization to novel objects in a real-world environment. 
%------------------------------------------
\begin{minipage}{0.94\linewidth}
\vspace{3mm}
\captionof{table}{\textbf{Reconstruction Methods Variations.} The proposed depth augmentations and skip connections improve the quality shell reconstruction.} 
\label{tab:recons_var}
\begin{tabular}{ *1{C{0.5in}} *1{C{1.5in}} *3{C{0.52in}}}\toprule[1.5pt]
\bf Method & \bf Modification &\bf  Chamfer   \\\midrule

Shell          & None                           & 6.43E-3   \\
Shell          & Baseline Augmentations         & 7.76E-3    \\
Shell          & No Skip Connections            & 8.99E-3       \\
Shell          & Synthetic Validation Set       & 5.99E-3       \\
HOF            & None                           & 8.34E-3     \\
HOF            & Baseline Augmentations         & 8.75E-3         \\
HOF            & Trained on ShapeNet            & 1.57E-2         \\
HOF            & Synthetic Validation Set       & 5.42E-3         \\
\bottomrule[1.25pt]
\end {tabular}\par
\vspace{3mm}
\end{minipage}
%------------------------------------------

% \subsection{Procedure to compute Grasp Feasibility Map}
% \label{app:feas_map}

% We sample grasps on the visible pointcloud of the object and use the object reconstruction to evaluate the geometric grasp feasibility.
% For every point on the downsampled visible pointcloud, we assign a grasp pose such that the axis joining the fingers (finger axis) is along the point normal and one of the fingers is just over the visible pointcloud with the maximum available ($85$~mm) gripper opening.
% The sampled grasp is considered geometrically feasible if the gripper does not collide with the object reconstruction.
% Since we are focusing on parallel-jaw grasps, for added grasp stability, we further impose a constraint that the normals at the set threshold number of points on the object reconstruction inside the gripper envelope need to be along the finger axis.
% For a fixed grasp position, the grasp orientation is changed about the finger axis (to eight discrete angles making a full rotation) to evaluate if any of these grasp poses lead to a feasible grasp.
% %
% This procedure generates sets of feasible and infeasible grasps on the object, which we collectively refer to as the  \textit{Geometric Grasp Feasibility Map} on the object view. The grasp feasibility map can serve as a set/manifold to sample and optimize a grasp on.

\subsection{Mechanics of Grasp Quality}
\label{app:grasp_mech}

A commonly accepted grasp quality metric is a measure of the maximum external wrench the grasp can resist~\cite{FerrariCanny92}. 
The grasp feasibility evaluation provides a set of feasible grasps over the visible object surface. However, not all feasible grasps are equally robust under external disturbances. 

The stability of a parallel-jaw grasp against an external wrench depends on the frictional resistance offered by the grasp (finger contacts) and the distance between the grasp location and the external wrench application point~\cite{Roa14graspq}. From the Coulomb friction law, the frictional resistance at finger contacts is proportional to the finger contact area. In the absence of any task-related external wrenches, only gravitational and inertial forces act at the center of mass of the object. 
Therefore, the grasp stability (quality) is directly proportional to the area of finger contacts and inversely proportional to the Euclidean distance of the grasp location from the center of mass of the object. We assume the objects are of uniform density, so the center of mass is the same as the center of geometry. 
Since the feasible grasps from the grasp feasibility map have at least the set threshold number of object points inside the grasp envelope, the finger contact area for these grasps is of the same order of magnitude and can be neglected when computing the grasp quality. 
Finally, the grasp quality is inversely proportional to the Euclidean distance of the grasp from the object's center of geometry, which can be inferred from the object reconstruction. 

Intuitively, an object grasped farther from the center of geometry is likely to rotate and fall out out the grasp under an external disturbance, therefore, such a grasp has low quality.

\subsection{Significance of Grasp Precision Success Rate}
\label{app:grasp_precision}
The geometric grasp precision success rate characterizes how closely the object reconstruction matches with the ground truth object model and its impact on the grasp planning process and grasp success. 
Moreover, it removes the dependence of grasp success on the relative sizes of the gripper and test objects used for the experiments which is overlooked in most the recent work on grasping~\cite{tenPas17gpd, mousavian19graspvae, merwe19gag}. A gripper with large opening can successfully grasp an object between fingers even with large errors in object model or grasp proposal. 
We believe this could partially explain why simple heuristic based grasping on only the visible pointcloud leads to high success rate in~\cite{merwe19gag}. 
In our experiments as well, we confirm that grasping only based on the visible pointcloud with full gripper opening achieves good success rate as show in the last column of \tabref{tab:grasp_precision}.
Therefore, particularly when evaluating reconstruction-based grasp planning methods, it is more informative to report the grasp precision success rate computed over the entire reconstruction than a top grasp success percentage with no constraint on the gripper opening used.

\subsection{Mujoco Setup and Experimental Procedure}
\label{app:mujoco_exp}

\begin{figure}
\vspace{-5mm}
\centering
\includegraphics[scale=0.35]{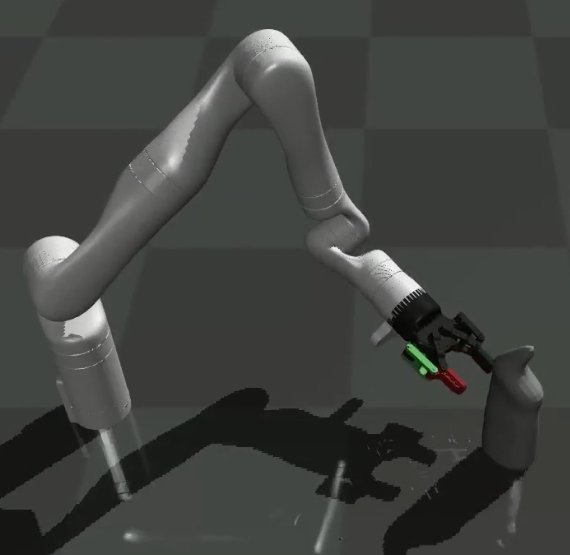}
\caption{Mujoco Setup. The simulation platform allows us to apply an external force disturbance to the grasped object and record its motion to evaluate the grasp stability.}
\label{fig:mujoco_setup}
\vspace{-3mm}
\end{figure}

We evaluate the physical significance of the quality of grasps estimated using shell reconstruction by testing the stability of those grasps under external disturbance . 

\begin{minipage}{0.95\linewidth}
\vspace{3mm}
\centering
\captionof{table}{\textbf{Grasp success \% in Mujoco simulation.} Grasps across different grasp quality ranges are executed with robot set up in Mujoco and the average grasp success rate is reported. The grasps of estimated high quality suceed more than those of the low quality.} 
\label{tab:mujoco_table}

\begin{tabular}{*1{C{0.45in}} *1{C{0.3in}} *1{C{0.4in}} *1{C{0.3in}} *1{C{0.4in}} *1{C{0.3in}} }\toprule[1.5pt]

\bf Object & \textbf{Top} [1.0-0.8) &\textbf{T-Mid} [0.8-0.6) &\textbf{Mid} [0.6-0.4) &\textbf{Mid-L} [0.4-0.2) & \textbf{Low} [0.2-0.0) \\\midrule
Cheezit & 82.7 & 72.9 & 70.0 & 43.8 & 15.6\\
Jello  & 90.4 & 88.1 & 78.1 & 68.0 & 33.5\\
Pringles & 99.3 & 95.1 & 93.5 & 72.3 & 11.2 \\
Soup & 99.2 & 98.6  & 81.8 & 71.0 & 13.6 \\
Mustard & 95.4	& 90.9 & 80.4 & 38.7 & 5.1 \\
Clorox & 100.0	& 75.0 & 68.4 & 23.9 & 0.9\\
Weiman & 92.1 & 84.5 & 32.1 & 14.3 & 1.1 \\
\bottomrule[1.25pt]
\end {tabular}\par
\vspace{3mm}
\end{minipage}

% %
% \begin{figure*}
% \begin{minipage}{0.94\linewidth}
% \vspace{0mm}
% \centering
% \captionof{table}{\textbf{Grasp success rate on the real robot system.} For 4 test views per object, 5 high quality grasps are executed and the number of successful grasps is recorded.} 
% \label{tab:robot_table}
% \begin{tabular}{ *1{C{1.2in}} | *7{C{0.4in}}}\toprule[1pt]
% \bf Object & Cheezit & Jello & Pringles & Soup & Mustard & Clorox & Weiman \\\midrule
% % \\
% \bf Successful Grasps (\%) & 19/20 (95) & 18/20 (90) & 18/20 (90) & 19/20 (95) & 20/20 (100) & 20/20 (100) & 19/20 (95)\\
% \bottomrule[1pt]
% \end {tabular}\par
% \vspace{1mm}
% \end{minipage}
% \end{figure*}
% % 

%
We built a setup in Mujoco simulator~\cite{Todorov12mujoco} with a Kinova Gen3 robot arm and a Robotiq 2F-85 gripper to replicate our real robot system. 
To minimize the effect of the arm motion planning limitations on diverse grasp evaluation, we place the test object in the robot's dexterous workspace.
Similar to the geometric grasp experiments, we evaluate the performance of precise grasping in Mujoco. For every test grasp sample, the robot approaches the object with a gripper opening only $15$~mm more than the estimated grasp width. The gripper is closed with a grip force of about $30$~N. The friction between the fingers and all the objects is set to $0.6$.
To evaluate the grasp stability, the grasped object is lifted up and an external force of $10$~N is applied in multiple directions at the center of geometry of the object. If the object is not moved in the grasp more than $30$~degrees, the grasp attempt is counted as success. Otherwise, the grasp attempt is counted as a failure.

\tabref{tab:mujoco_table} shows the percentage of successful grasps across five grasp quality ranges from the grasp quality maps. For 10 test views per object, we perform 250 grasp simulations per object in every grasp quality range and report the average success rate in \tabref{tab:mujoco_table}.
The grasp quality estimated using the shell object reconstruction reflects well in the grasp success rate. More grasps in the high quality range resist the external disturbance force and succeed than the grasps in the low quality range.
These results highlight that the shell object representation captures the key measures of the object geometry such as overall shape and the center well. These geometric features provide the scope for generating sets of grasps and their relative quality accurately.

\subsection{Detailed Comparison of our Shell Reconstruction Method with PointSDF}
\label{app:shell_vs_ptsdf}

% The Occupancy Networks (OCCNet)~\cite{mescheder19occnet} generates object geometries by learning an occupancy function which is concurrent to learning a signed distance function as proposed in DeepSDF or PointSDF paper.  Given the popularity and state-of-the-art results shown in OCCNet paper we selected OCCNet as one of the baselines which learns the functional representation of the occupancy/distance of object geometry.
% Following reviewers recommendation, we also provide quantitative (Chamfer score) and qualitative evaluation of our method with PointSDF reconstruction method~\cite{merwe19gag}.

% We use pre-trained version of PointSDF network for evalaution. The network is trained on YCB objects and tested on YCB objects in their paper. Since our test object dataset is also comprised of YCB objects, we expect the pointSDF to perform well. However, shell reconstruction network, although trained on simple synthetic shapes, outperforms the PointSDF method. Moreover, the total object mesh inference time is significantly faster for shell reconstruction than that for PointSDF.

% When tested on our test dataset of YCB objects, the average Chamfer score for reconstructions from PointSDF is $15.9$E-3 while that from shell reconstruction is $6.4$E-3, when compared against the ground truth object meshes in the camera frame.

In this section we compare our Shell method and PointSDF baseline in detail with multiple examples.

% \myparagraph{Reconstruction accuracy}:
\tabref{tab:shell_vs_ptsdf} compares the reconstructions from these two methods quantitatively as well as qualitatively. We observes that the shell reconstructions capture the object geometry details much more accurately. The geometric details from the input are lost in the PointSDF reconstructions. For example, the PointSDF reconstructions for Clorox and Weiman bottle appear to be some generic bottle shape prior the network has learnt. The Chamfer scores quantify the superior recosntruction accuracy of shell method. 

% \myparagraph{Computation time}:
% Since the entry and exit depth images of the object shell contain neighbor-hood information, they represents the partial object meshes which we stitch together using the contours of the entry and exit images to get the complete object mesh. This operation is computationally efficient and linear in time. The complete process (from input depth image to complete object mesh) takes a fraction of a second. 
% On the other hand, for methods which learn the object geometry as occupancy or signed-distance function, mesh generation involves either 1)sampling very large number of points in the workspace and evaluating the network for those points, or 2) iteratively refine the resolution of the samples based on the occupancy recorded in the previous iterations forward pass of the network. Finally, an algorithm such as Marching Cubes is used get the isosurface and the object mesh. A naive implementation of such mesh generation takes upto a minute as implemented in the PointSDF method. With algorithm such as Multiresolution IsoSurface Extraction (MISE)~\cite{mescheder19occnet} introduced in the OCCNet paper, the mesh inference time can be about $3$ seconds as reported in~\cite{mescheder19occnet}. This shows that object mesh inference is significantly faster with shell representation.

% With these quantitative and qualitative results, we highlight that, although trained on simple synthetic shapes, shell reconstructions provides superior camera frame reconstructions on real dataset of YCB-like objects in a fraction of a second.

\subsection{Adversarial object geometries and views for object reconstruction}
\label{app:adv_shape_and_views}

Shell representation with a pair of depth images imposes a monotonicity constraint along the axis perpendicular to the image plane. We believe that this constraint is not overly restrictive: only 5 (wine glass, mug, bowl, pan, stacking block) out of 77 objects in YCB dataset violate this monotonicity constraint from some views. We sacrifice the performance on this small set of objects for the characteristics the shell representation provides for achieving generalization and superior performance on the larger class of objects. Reconstructing such complex objects is difficult for most of the reconstructions methods unless trained for those specific classes of objects. 

For object and views where camera rays enter and exit the object more than once, the shell reconstruction may generate the outer geometry of the object or produce incomplete reconstruction as shown for a cup object in  \tabref{tab:shell_vs_ptsdf_adv}. However, note that shell reconstructions look much better compared to PointSDF since they maintain the overall geometry of the object in the input intact.

As discussed in \secref{sec:exp_clutter}, the object shell reconstruction can not generate complete object geometry if only small portion of the object surface is observed in the input data or if the input data is ambiguous. This limitation is present for any other reconstruction method as well, unless a strong shape prior is used or the method is overfitted to the training data and tested on the subset of the same data.

In \tabref{tab:shell_vs_ptsdf_adv}, we show object reconstructions for adversarial views where objects are seen from only narrower side. Both Shell reconstruction and PointSDF methods struggle for these views, since neither of them are trained to have strong shape prior. 
The shell reconstructions are qualitatively better than PointSDF reconstructions since they maintain the details in the input while completing the hidden back of the visible object using the average shape prior learnt. 
For the examples of Cheeze-it box and Ziploc box, the input contains information about only one face of the box-shaped object with no information from any other face to bound the backside of the object reconstruction. Both Shell and PointSDF generate a box of small thickness based on the prior learnt.

\newcommand{\psdfw}{3.5cm}
\newcommand\Tstrut{\rule{0pt}{2.6ex}}  % "top" strut
\newcommand\Bstrut{\rule{0pt}{-2.6ex}}  % "bottom" strut
\newcommand{\TBstrut}{\Tstrut\Bstrut}  % top&bottom struts

\begin{table*}
\centering
\captionof{table}{\textbf{Comparison of object reconstructions from Shell and PointSDF methods.} The reconstructions are shown from different angles for better understanding.}
\label{tab:shell_vs_ptsdf}
\begin{tabular}{c|c|c|c|c} 
\hline
     &   & \multicolumn{1}{c|}{GT} & \multicolumn{1}{c|}{PointSDF} & \multicolumn{1}{c}{Shell} \TBstrut \\ 
    \hline
    \multirow{4}{*}{ \rotatebox[origin=c]{90}{Roll} } &  $90^{\circ}$ & \includegraphics[width=\psdfw]{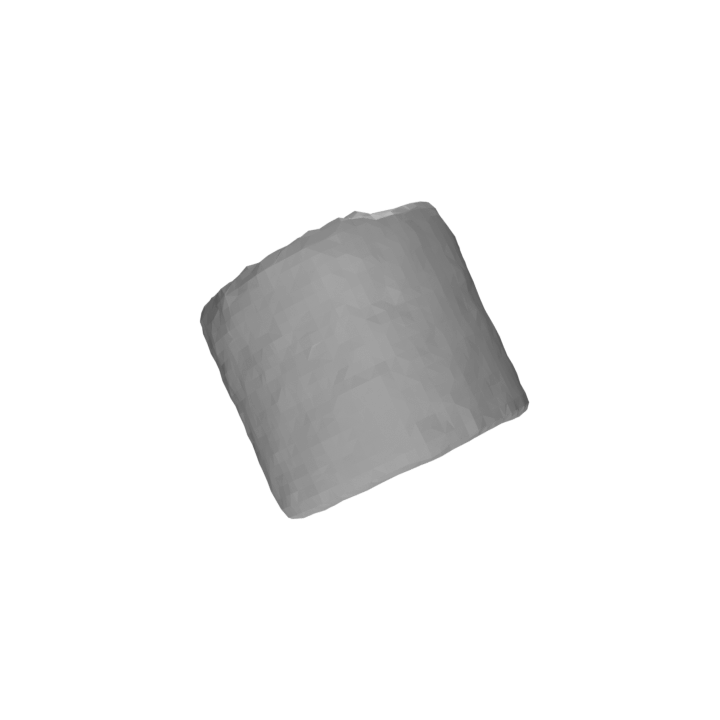} & \includegraphics[width=\psdfw]{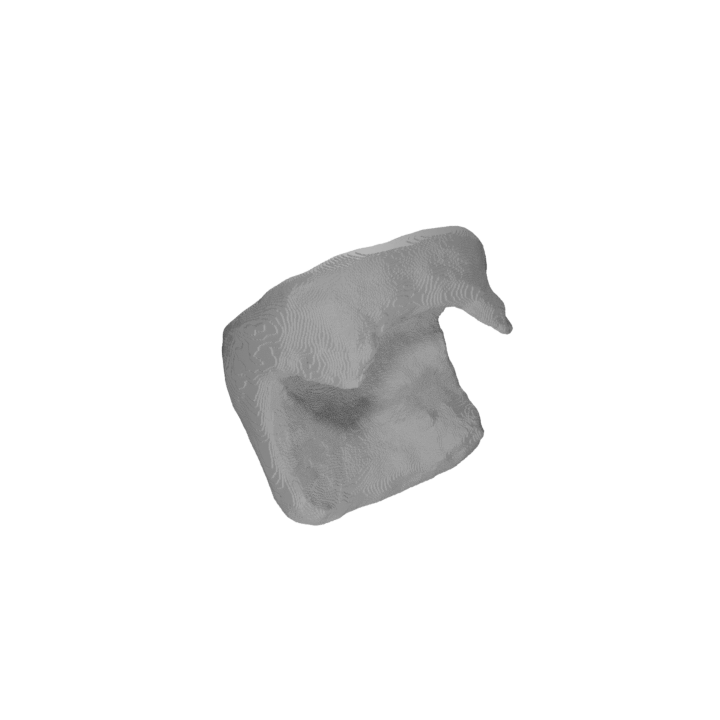}  & \includegraphics[width=\psdfw]{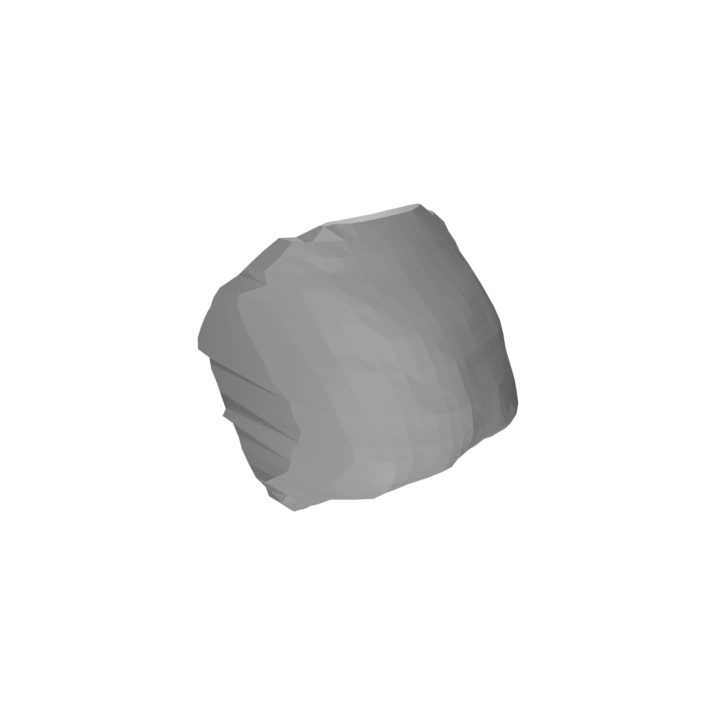}\\
\cline{2-5}
 &  $180^{\circ}$ & \includegraphics[width=\psdfw]{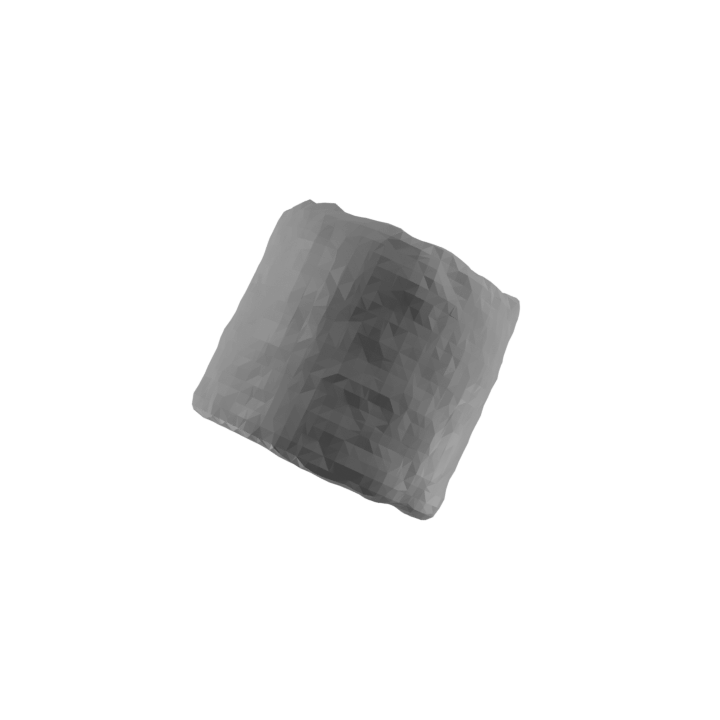} & \includegraphics[width=\psdfw]{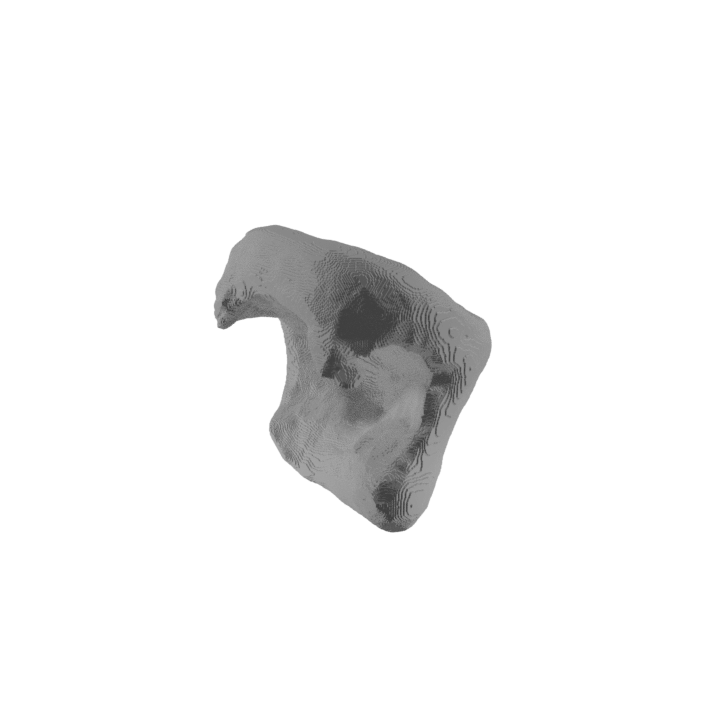} & \includegraphics[width=\psdfw]{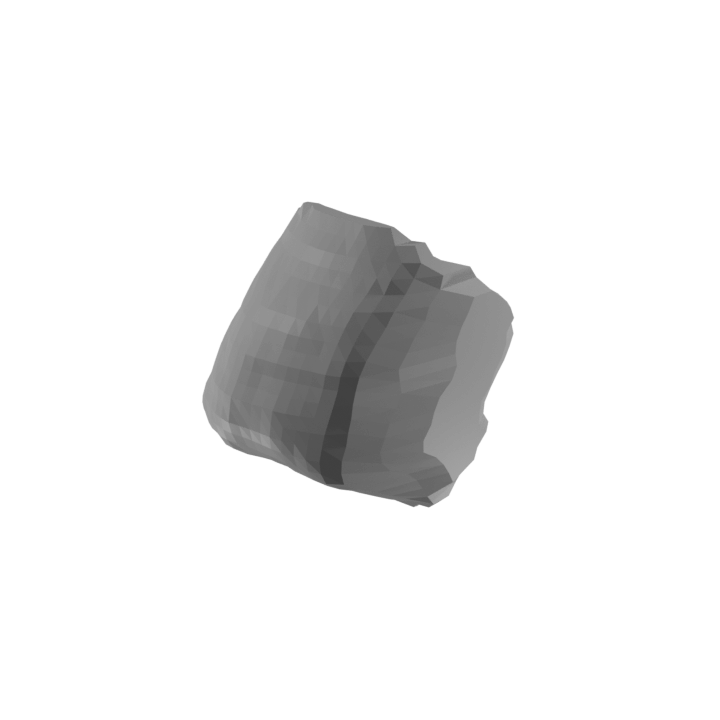}\\
\cline{2-5}
 &  $270^{\circ}$ & \includegraphics[width=\psdfw]{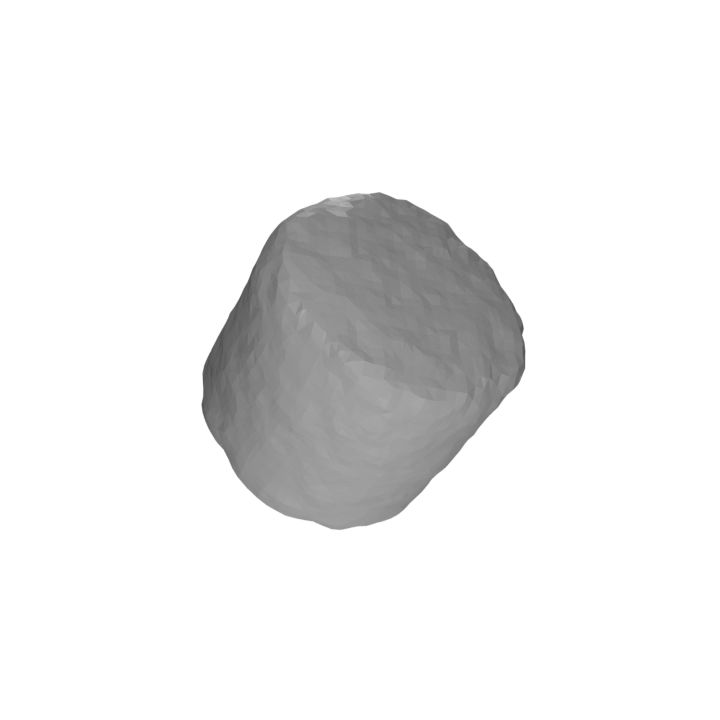} & \includegraphics[width=\psdfw]{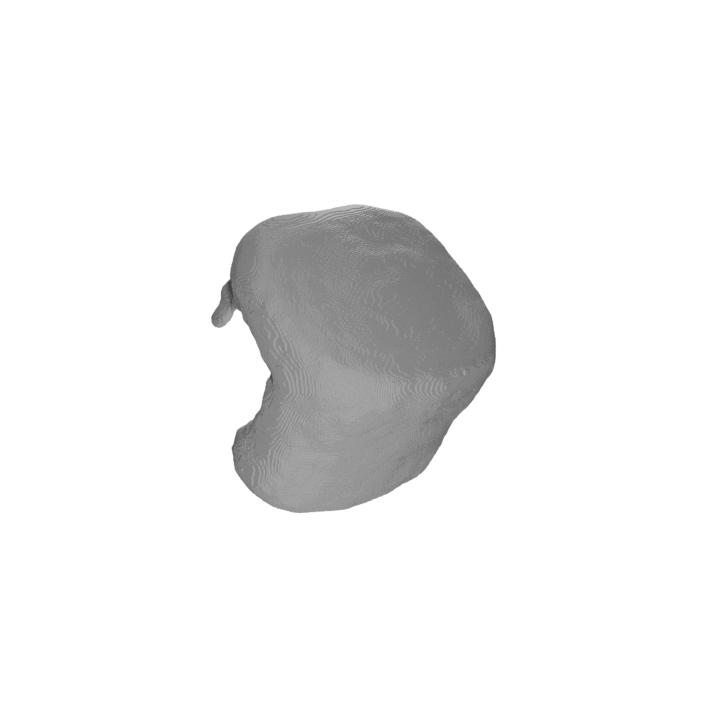} & \includegraphics[width=\psdfw]{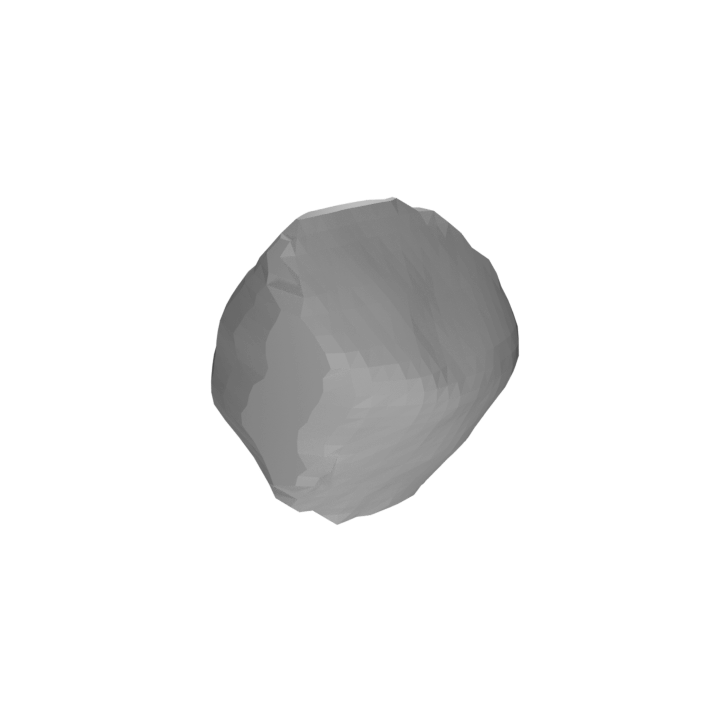}\\
\cline{2-5} & Chamfer &  & 2.06E-02 & 6.64E-03 \\
\hline
\multirow{4}{*}{ \rotatebox[origin=c]{90}{Weiman}} &  $90^{\circ}$ & \includegraphics[width=\psdfw]{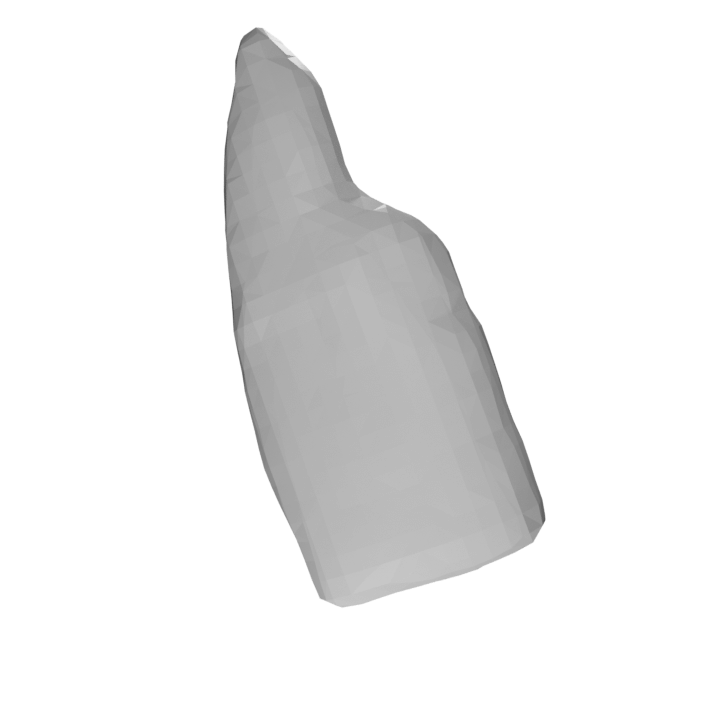} & \includegraphics[width=\psdfw]{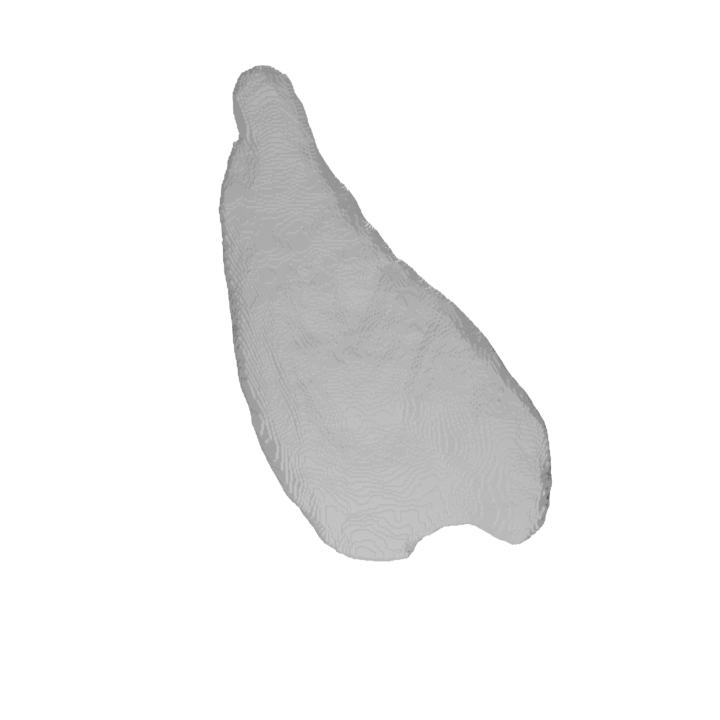} & \includegraphics[width=\psdfw]{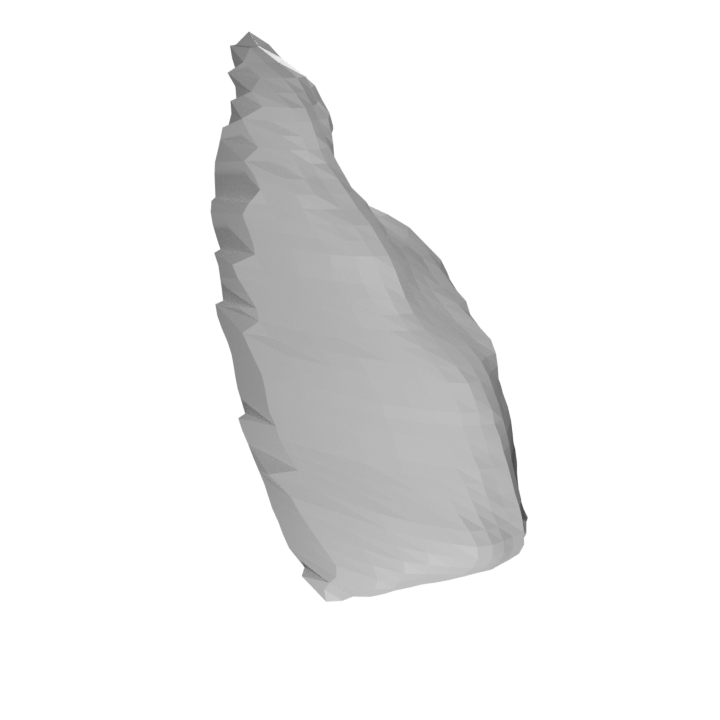}\\
\cline{2-5}
 &  $180^{\circ}$ & \includegraphics[width=\psdfw]{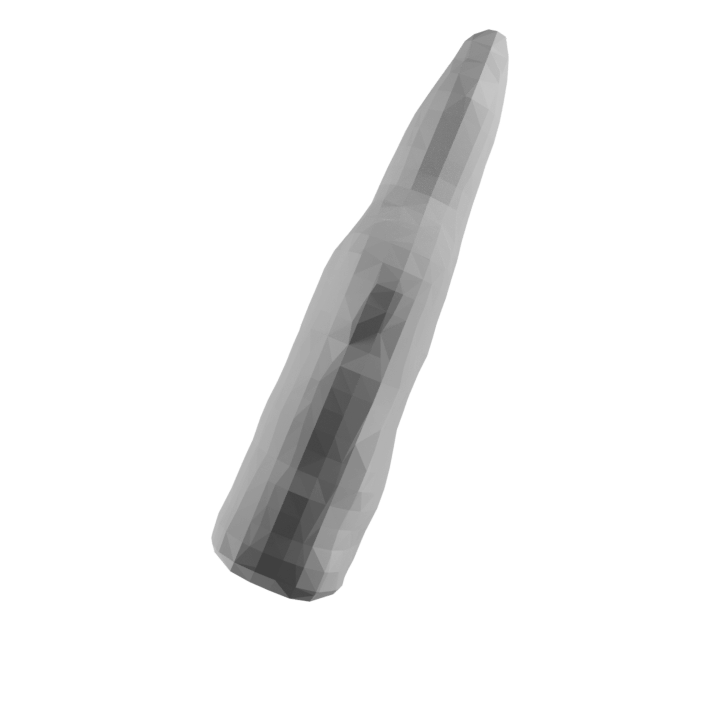} & \includegraphics[width=\psdfw]{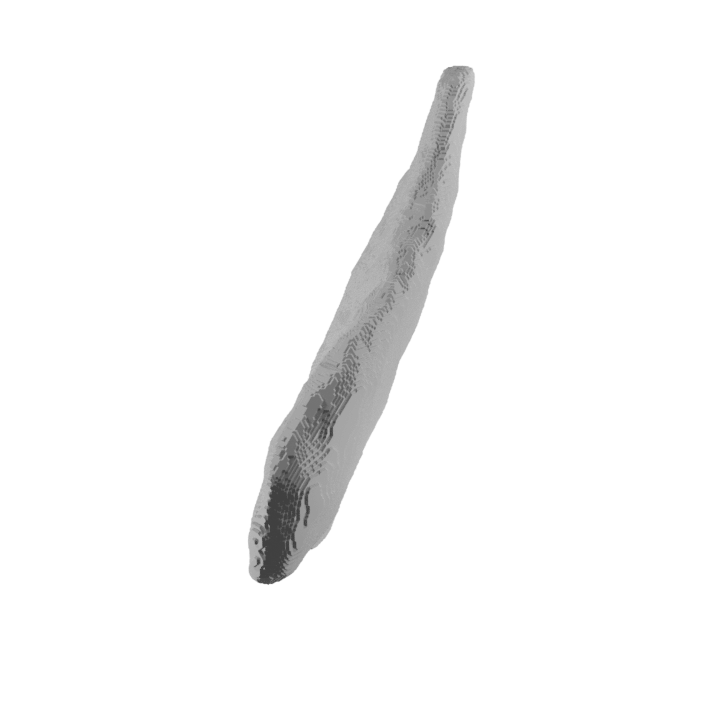} & \includegraphics[width=\psdfw]{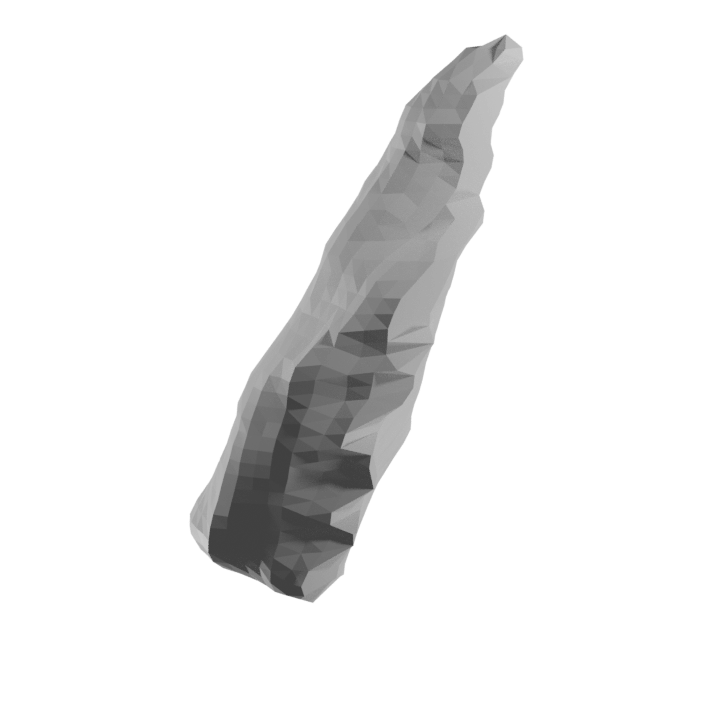}\\
\cline{2-5}
 &  $270^{\circ}$ & \includegraphics[width=\psdfw]{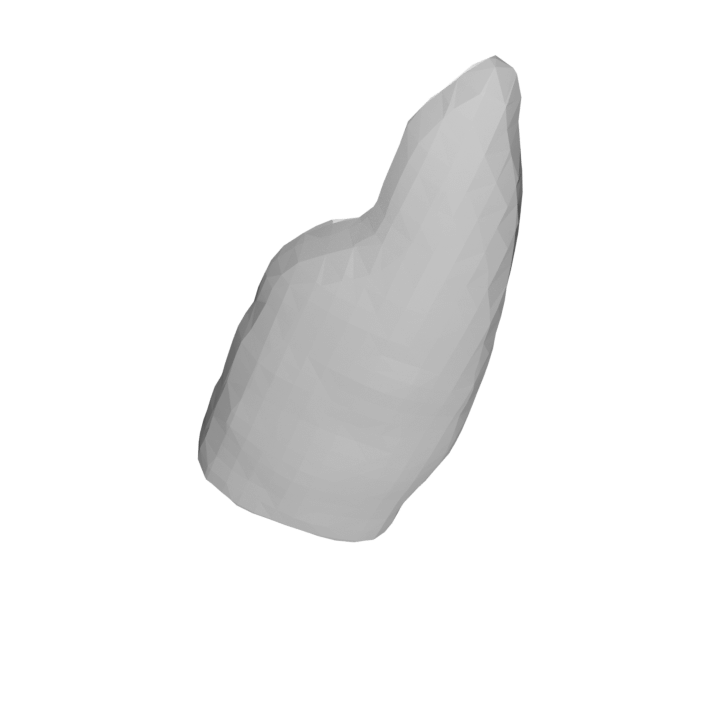} & \includegraphics[width=\psdfw]{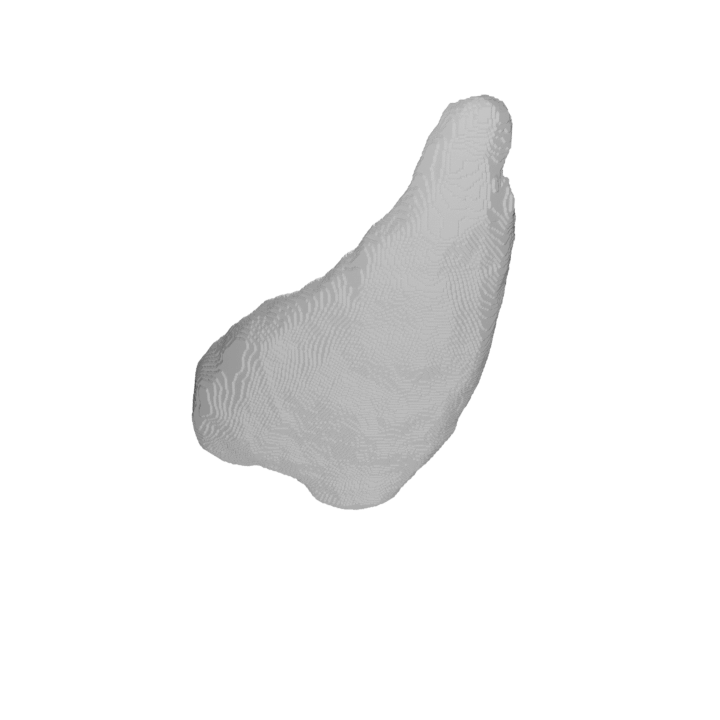} & \includegraphics[width=\psdfw]{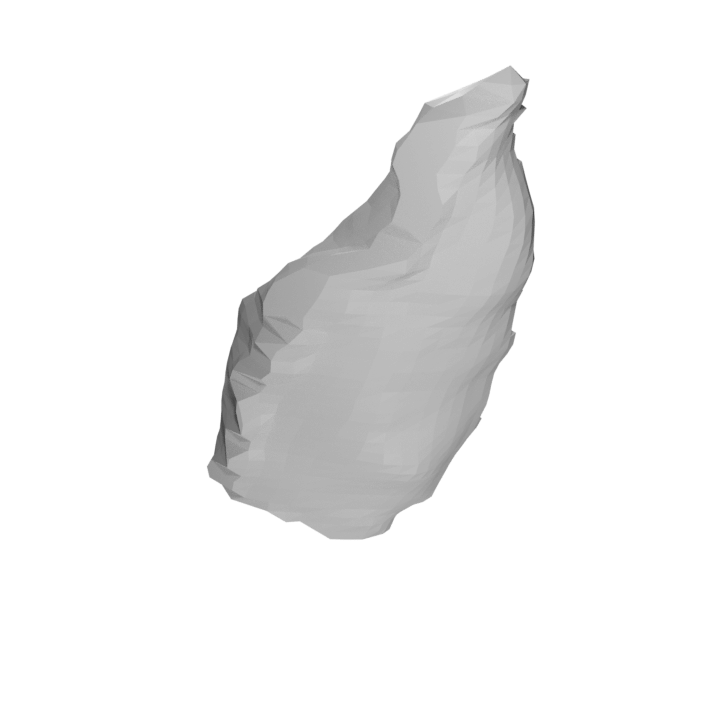}\\
\cline{2-5} & Chamfer &  & 2.01E-02 & 7.53E-03 \\
\hline

\end{tabular}
\end{table*}

\begin{table*}
\centering
\begin{tabular}{c|c|c|c|c} 
    \hline
     &   & \multicolumn{1}{c|}{GT} & \multicolumn{1}{c|}{PointSDF} & \multicolumn{1}{c}{Shell} \TBstrut \\ 
    \hline
    \multirow{4}{*}{ \rotatebox[origin=c]{90}{Clorox}} &  $90^{\circ}$ & \includegraphics[width=\psdfw]{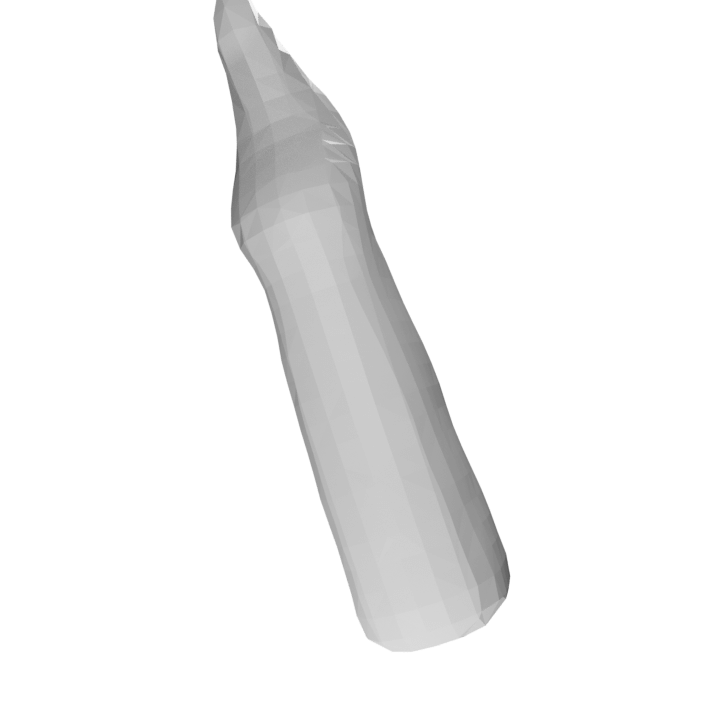} & \includegraphics[width=\psdfw]{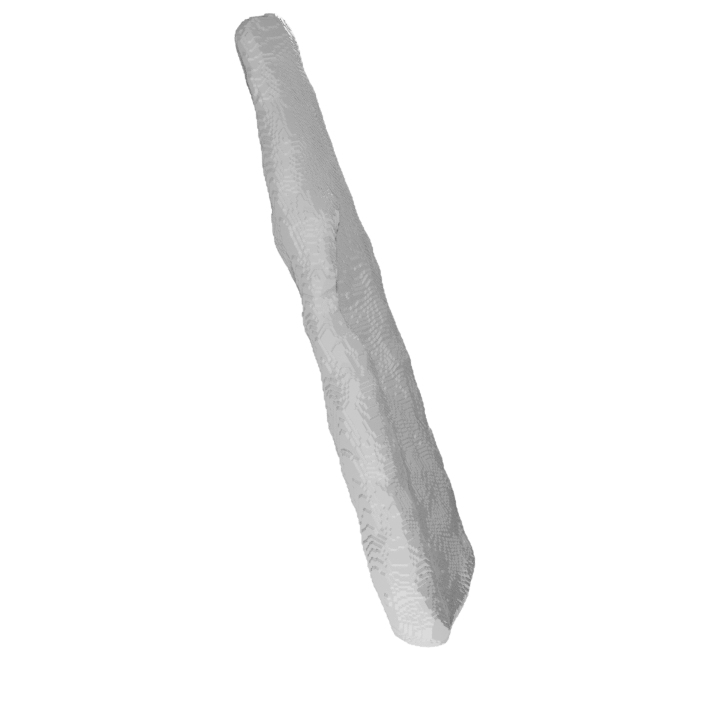} & \includegraphics[width=\psdfw]{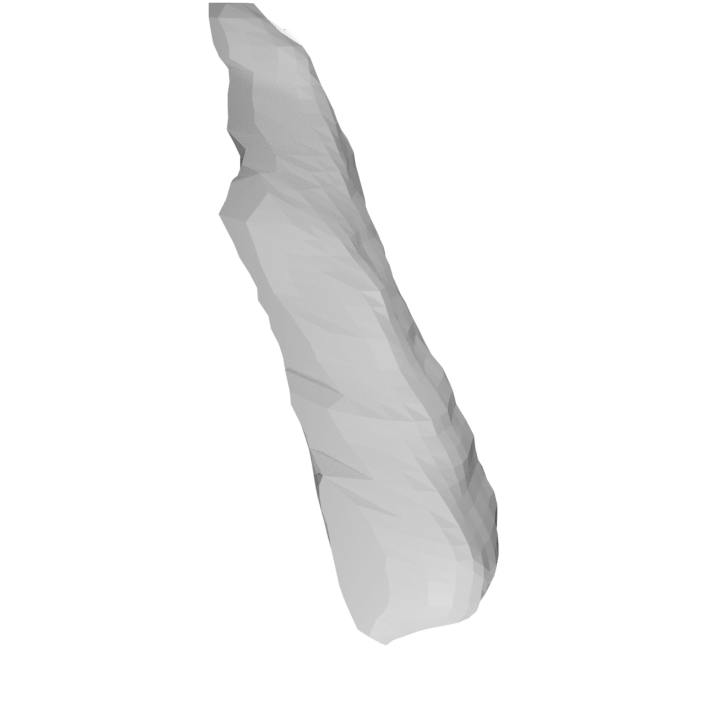}\\
\cline{2-5}
 &  $180^{\circ}$ & \includegraphics[width=\psdfw]{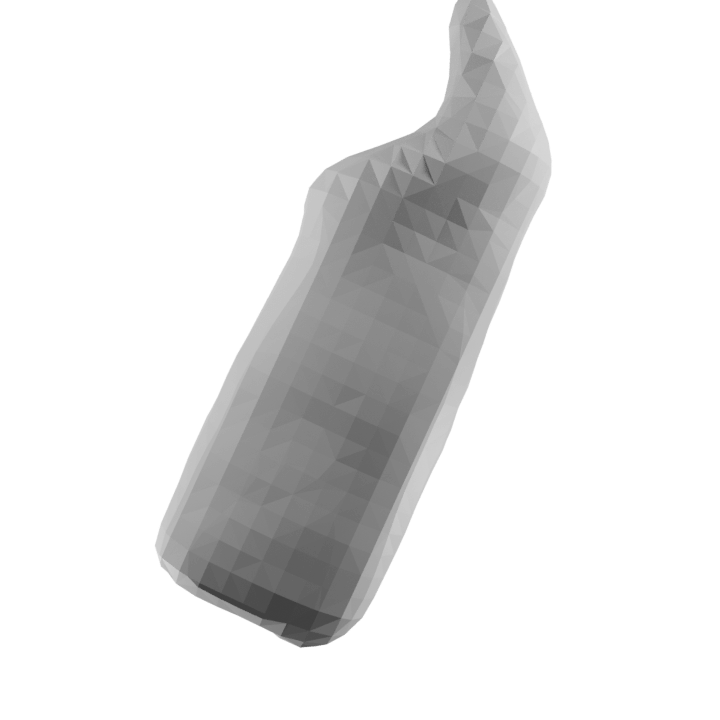} & \includegraphics[width=\psdfw]{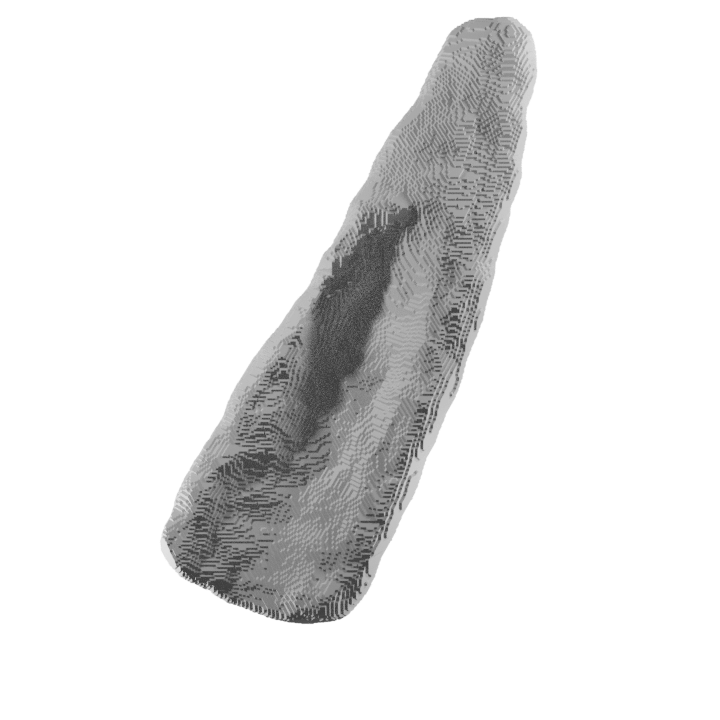} & \includegraphics[width=\psdfw]{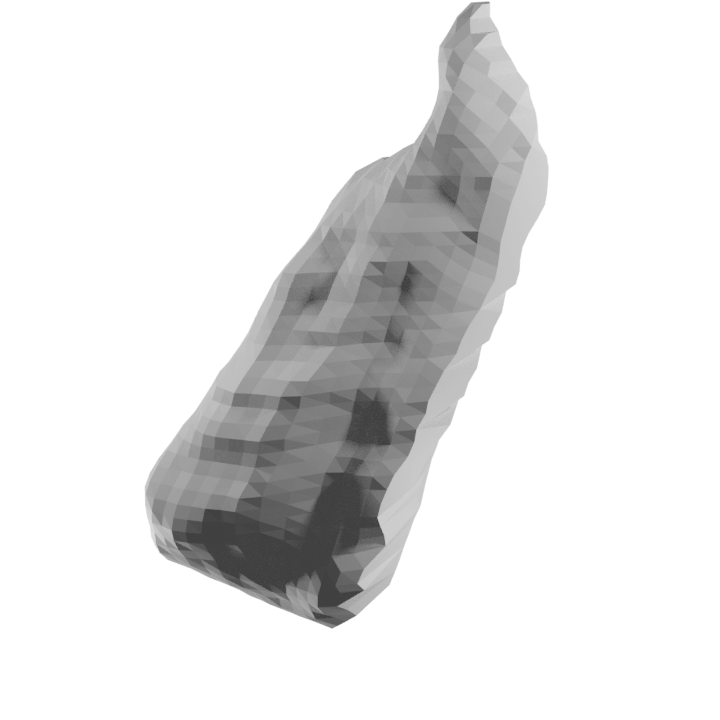}\\
\cline{2-5}
 &  $270^{\circ}$ & \includegraphics[width=\psdfw]{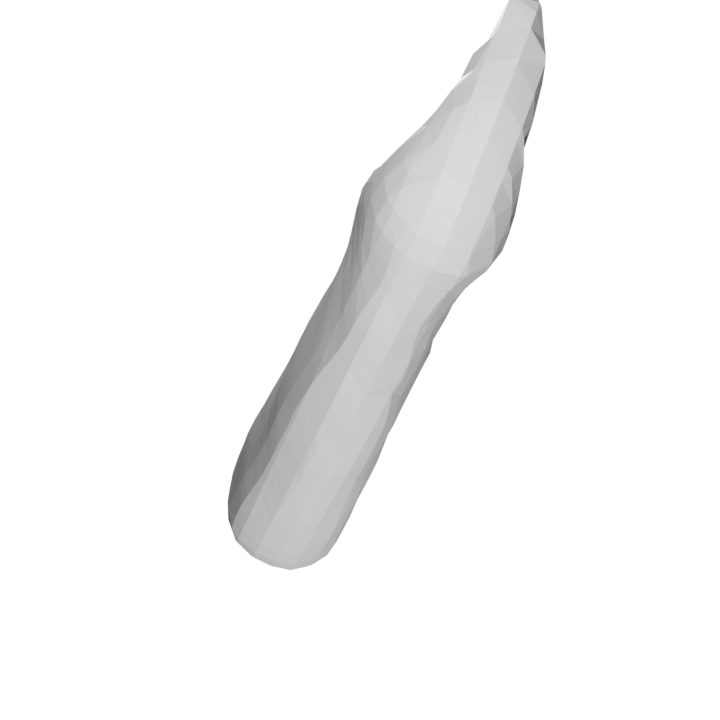} & \includegraphics[width=\psdfw]{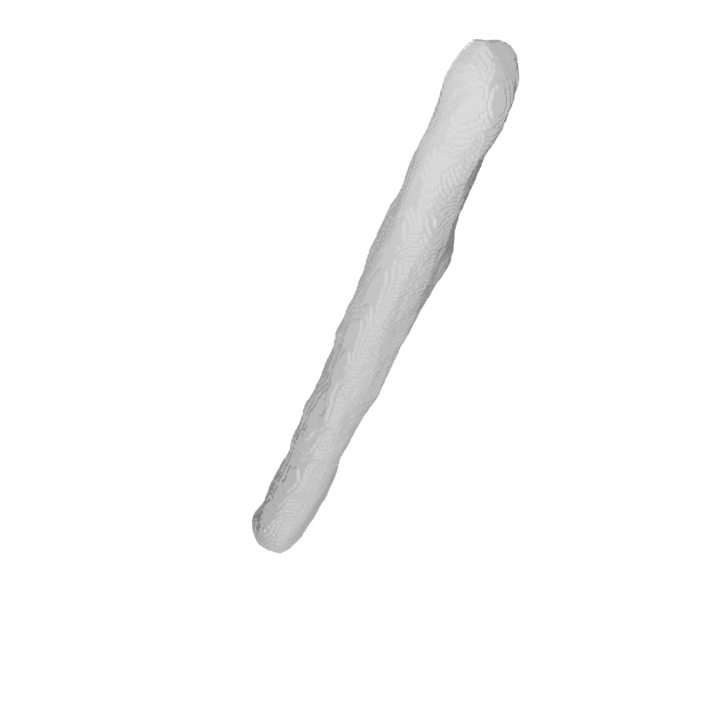} & \includegraphics[width=\psdfw]{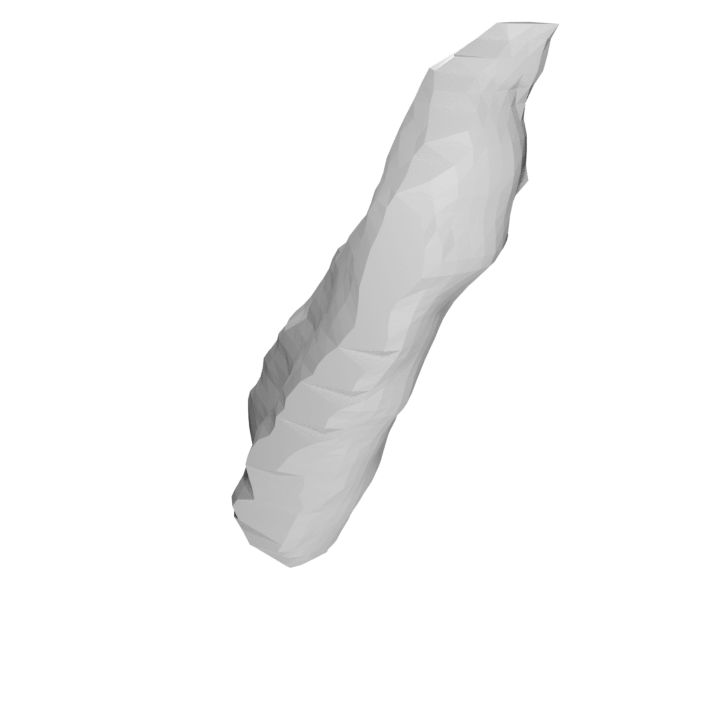}\\
\cline{2-5} & Chamfer &  & 1.97E-02 & 7.88E-03 \\
\hline
\multirow{4}{*}{ \rotatebox[origin=c]{90}{Kleenex}} &  $90^{\circ}$ & \includegraphics[width=\psdfw]{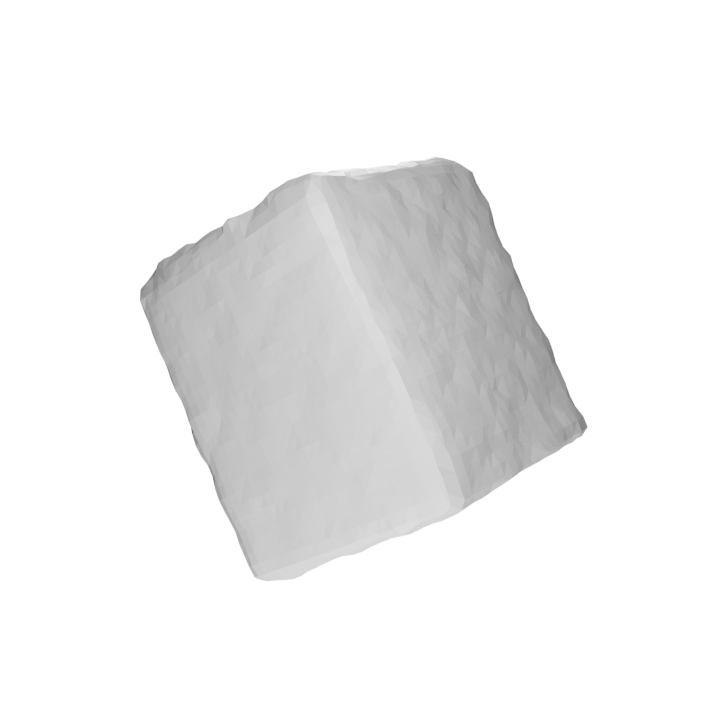} & \includegraphics[width=\psdfw]{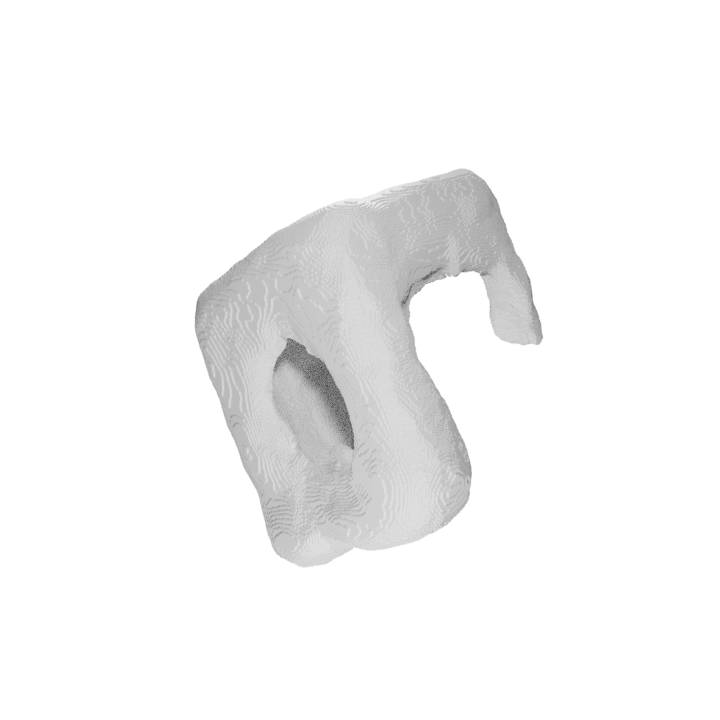} & \includegraphics[width=\psdfw]{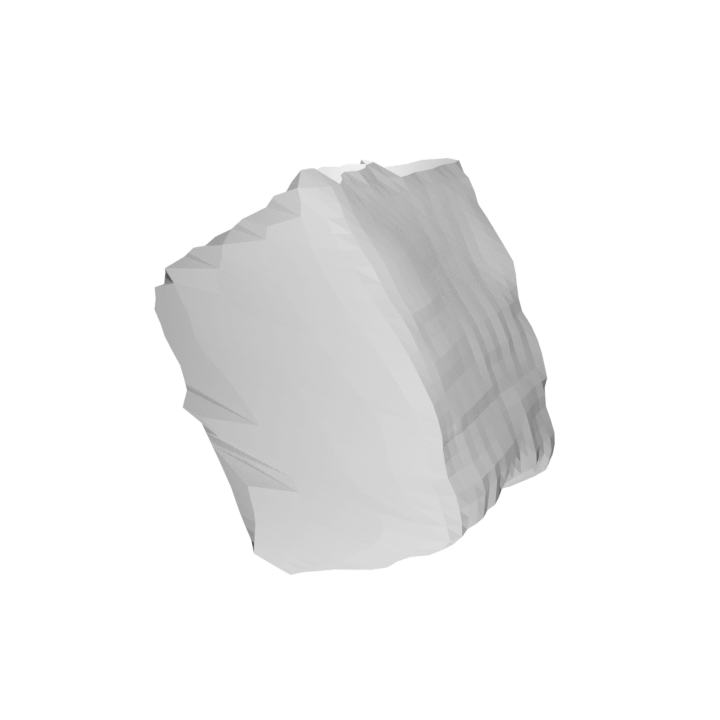}\\
\cline{2-5}
 &  $180^{\circ}$ & \includegraphics[width=\psdfw]{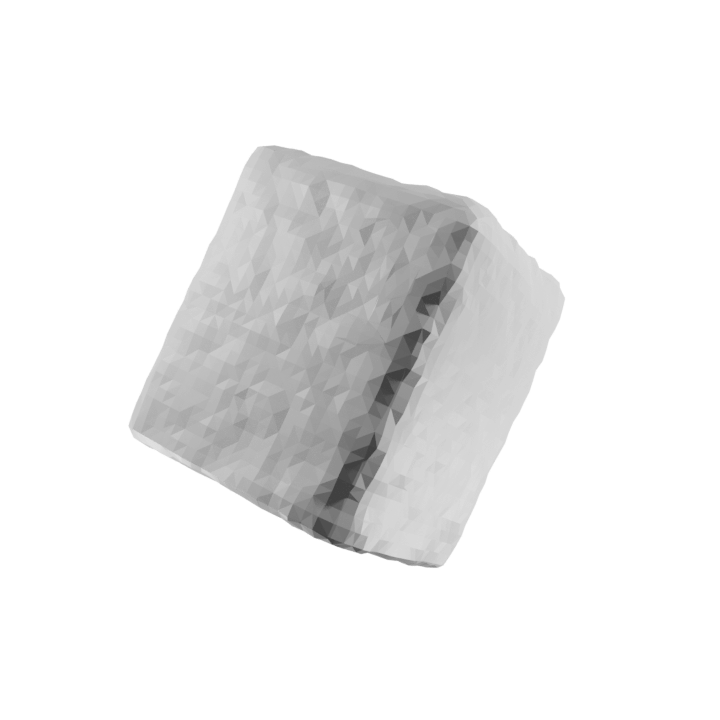} & \includegraphics[width=\psdfw]{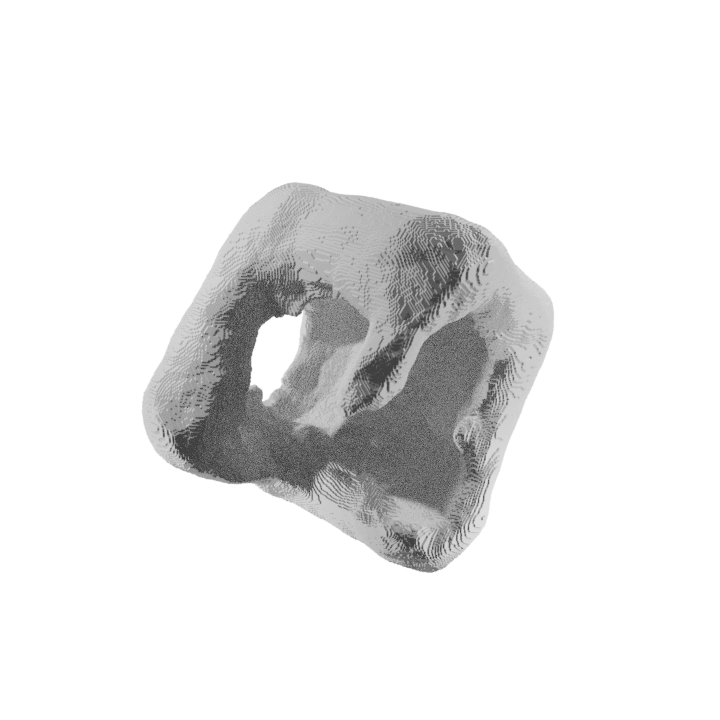} & \includegraphics[width=\psdfw]{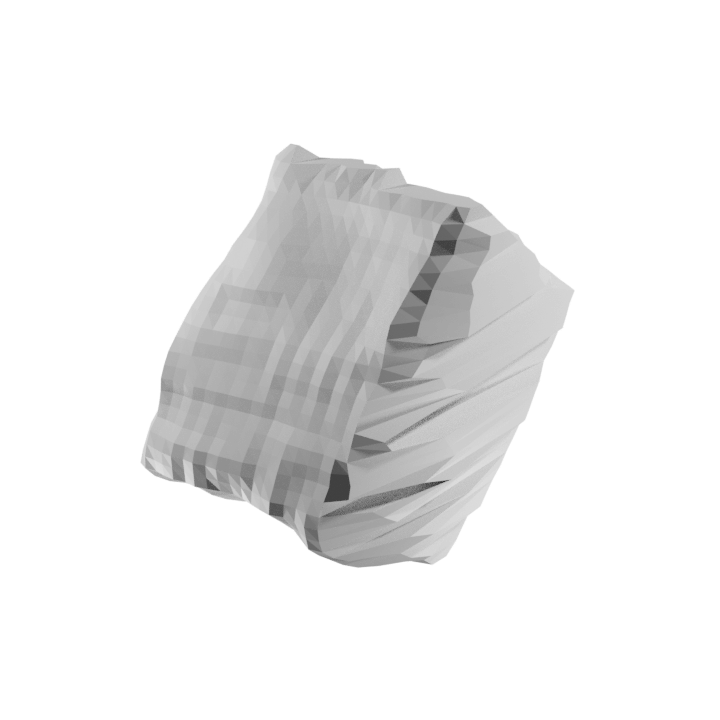}\\
\cline{2-5}
 &  $270^{\circ}$ & \includegraphics[width=\psdfw]{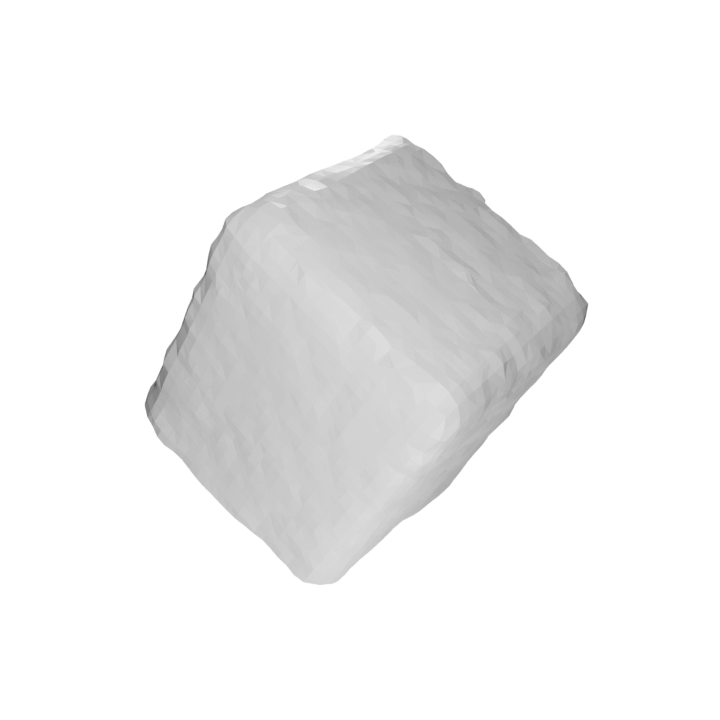} & \includegraphics[width=\psdfw]{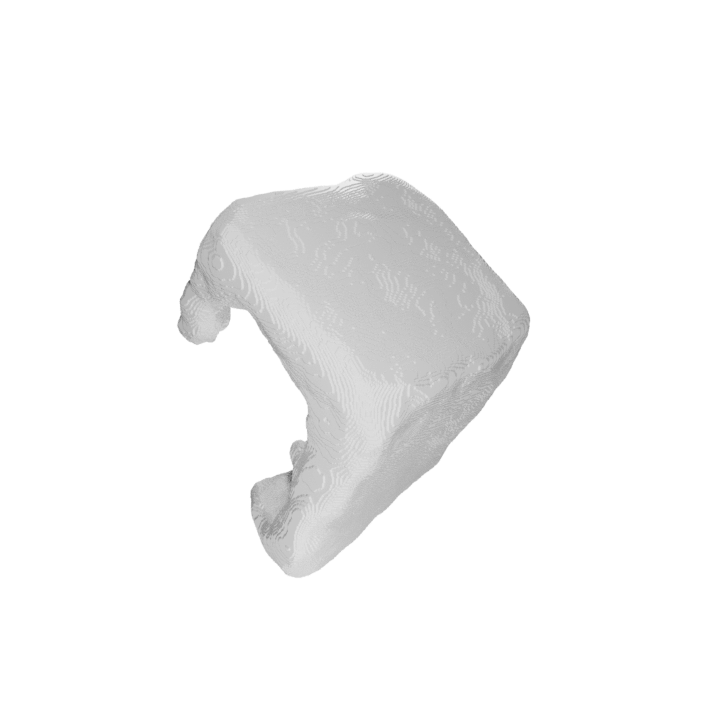} & \includegraphics[width=\psdfw]{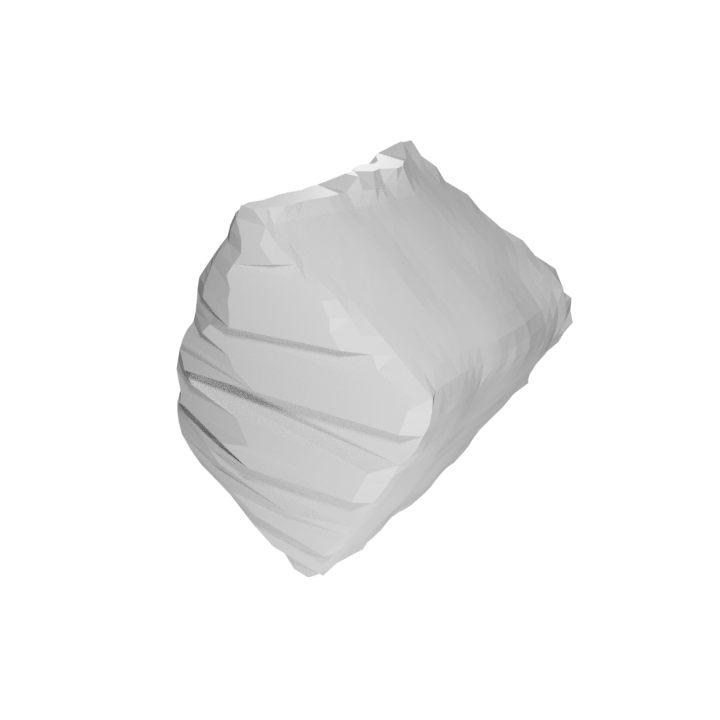}\\
\cline{2-5} & Chamfer &  & 2.35E-02 & 9.39E-03 \\
\hline

\end{tabular}
\end{table*}

    \begin{table*}
    \centering
    \begin{tabular}{c|c|c|c|c} 
    \hline
     &   & \multicolumn{1}{c|}{GT} & \multicolumn{1}{c|}{PointSDF} & \multicolumn{1}{c}{Shell} \TBstrut \\ 
    \hline
    \multirow{4}{*}{ \rotatebox[origin=c]{90}{Swiffer}} &  $90^{\circ}$ & \includegraphics[width=\psdfw]{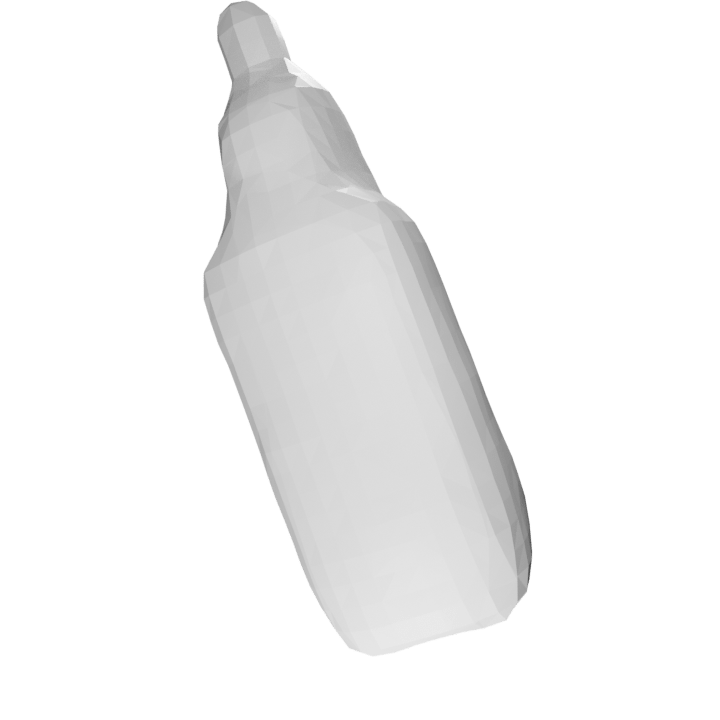} & \includegraphics[width=\psdfw]{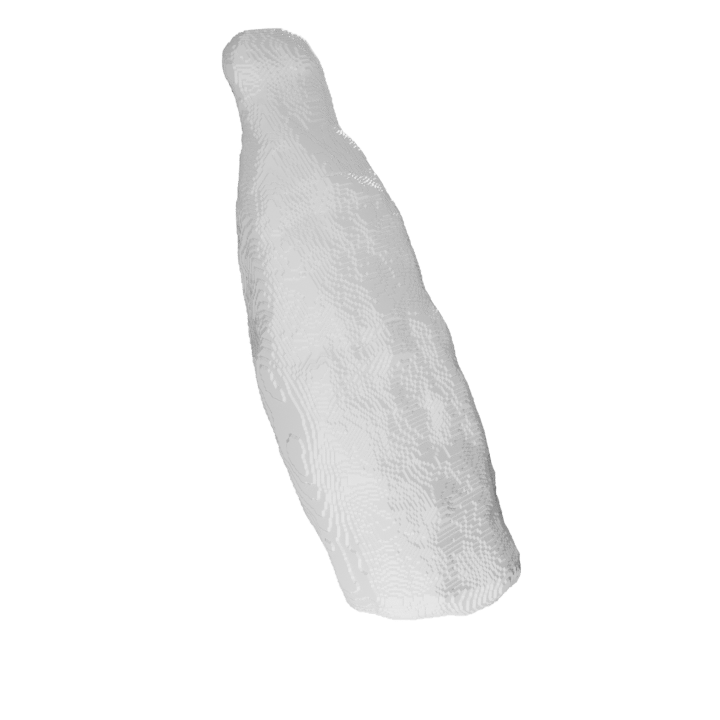} & \includegraphics[width=\psdfw]{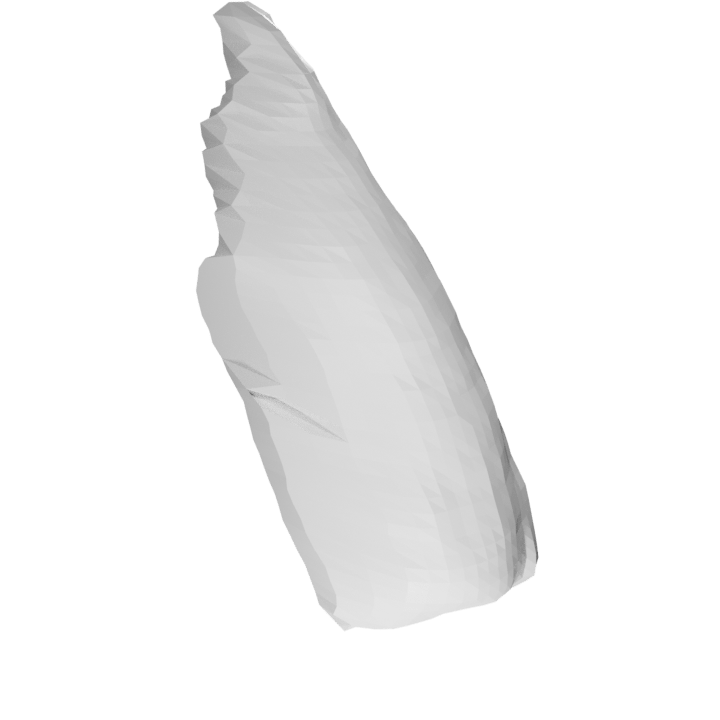}\\
\cline{2-5}
 &  $180^{\circ}$ & \includegraphics[width=\psdfw]{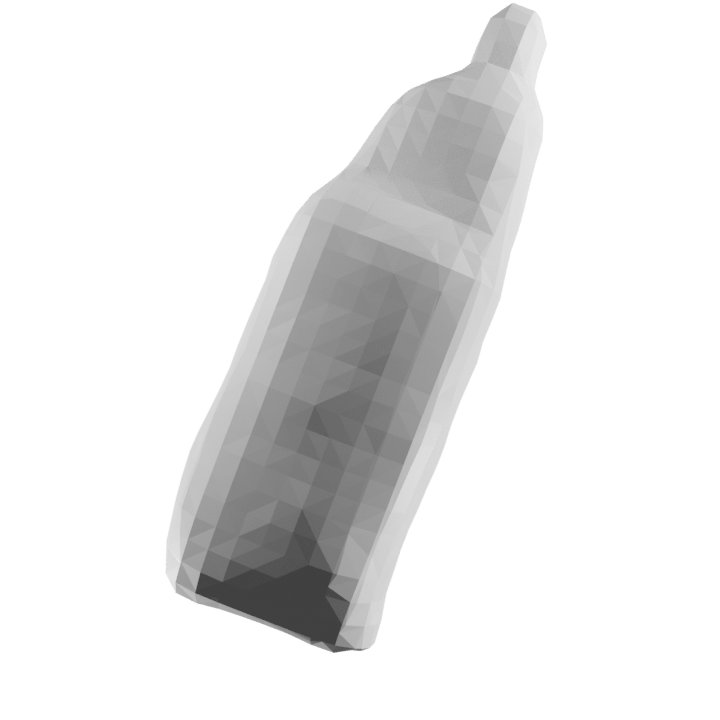} & \includegraphics[width=\psdfw]{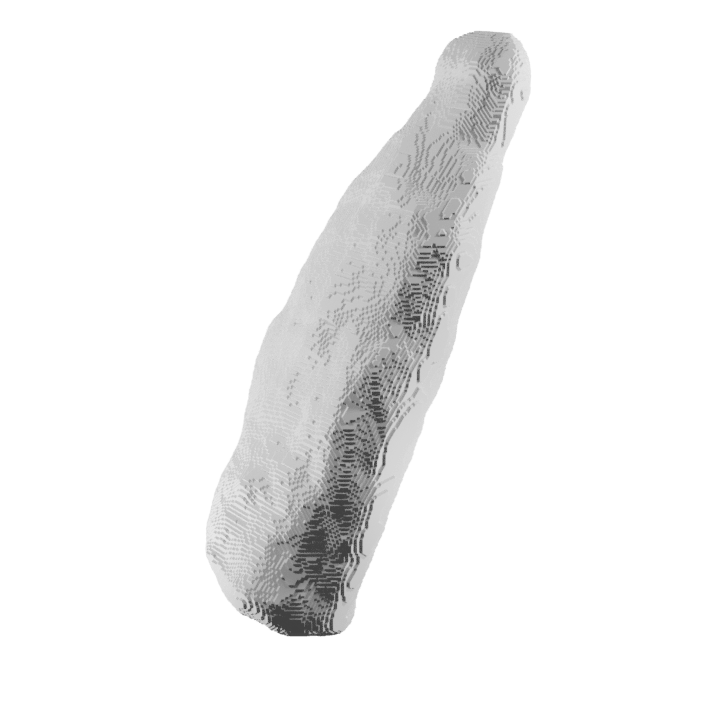} & \includegraphics[width=\psdfw]{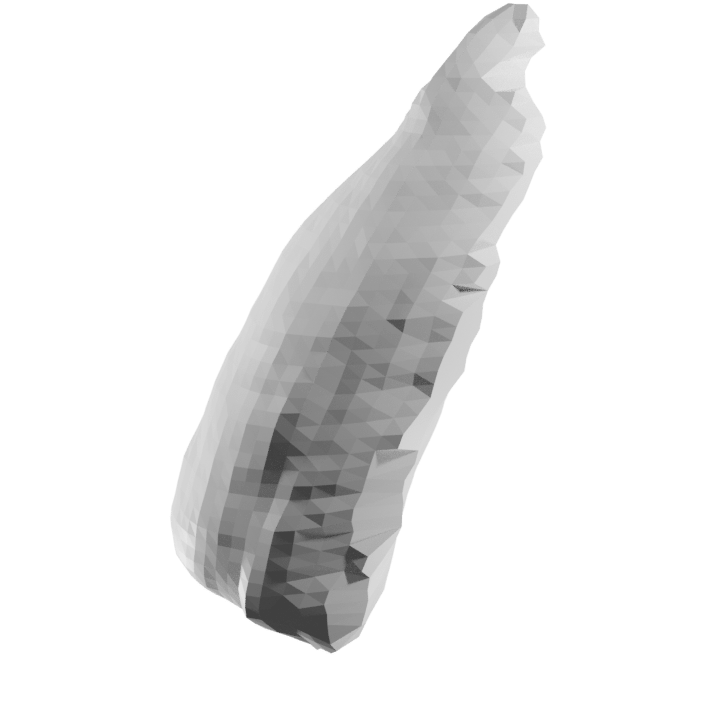}\\
\cline{2-5}
 &  $270^{\circ}$ & \includegraphics[width=\psdfw]{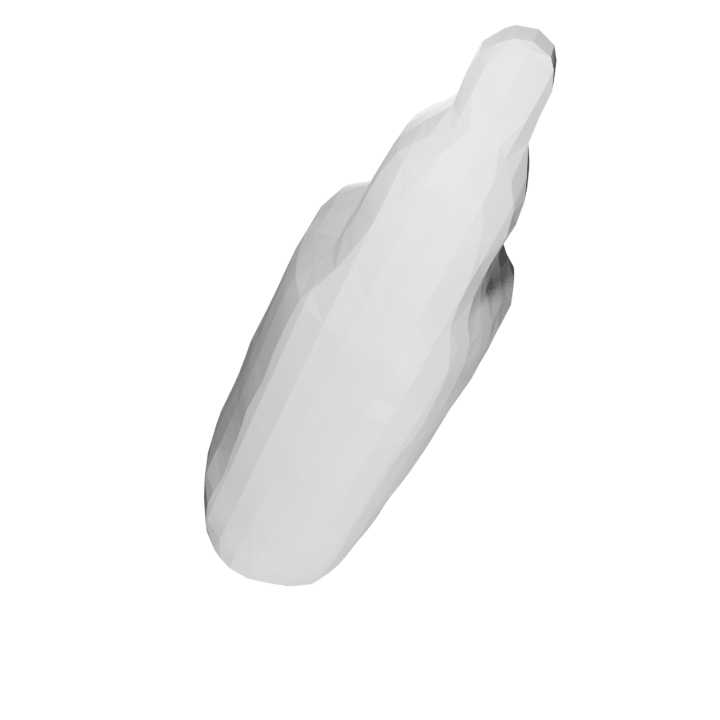} & \includegraphics[width=\psdfw]{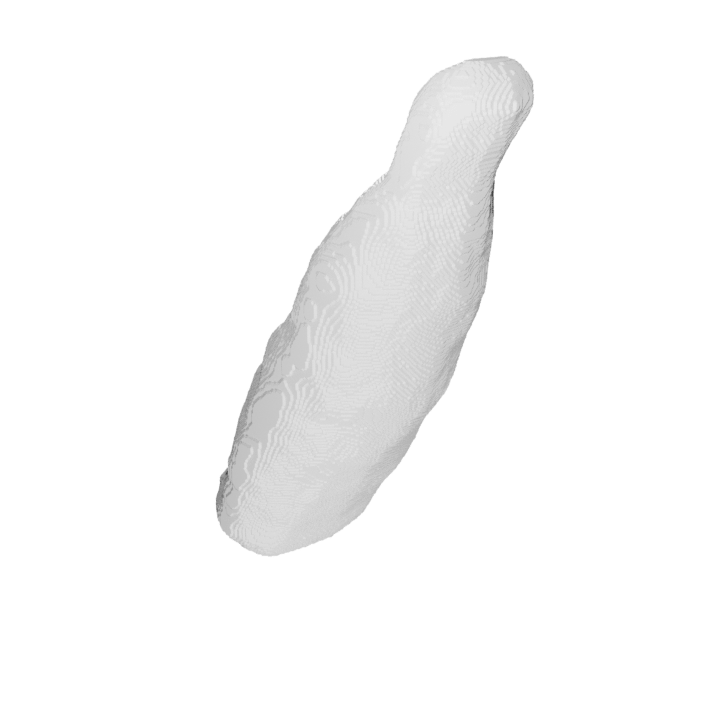} & \includegraphics[width=\psdfw]{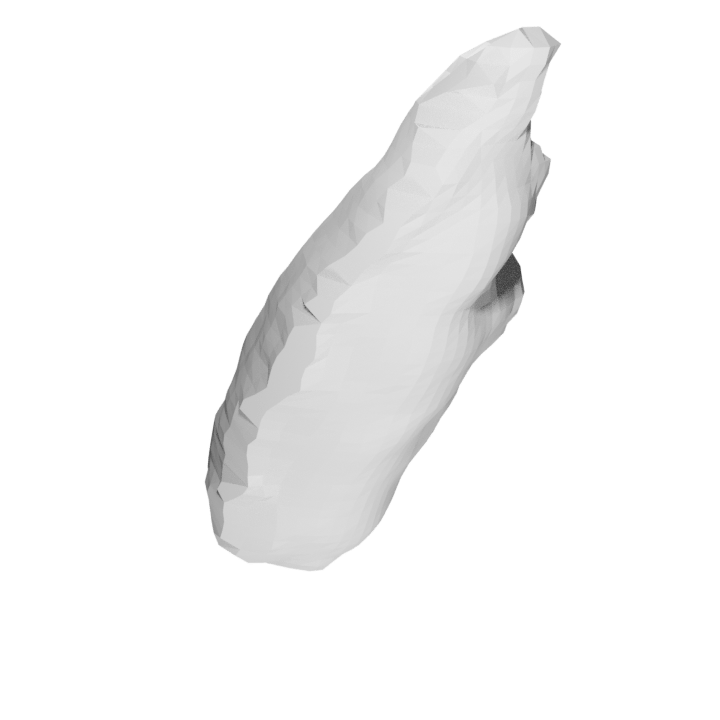}\\
\cline{2-5} & Chamfer &  & 2.16E-02 & 7.75E-03 \\
\hline
\multirow{4}{*}{ \rotatebox[origin=c]{90}{Soup}} & $90^{\circ}$ & \includegraphics[width=\psdfw]{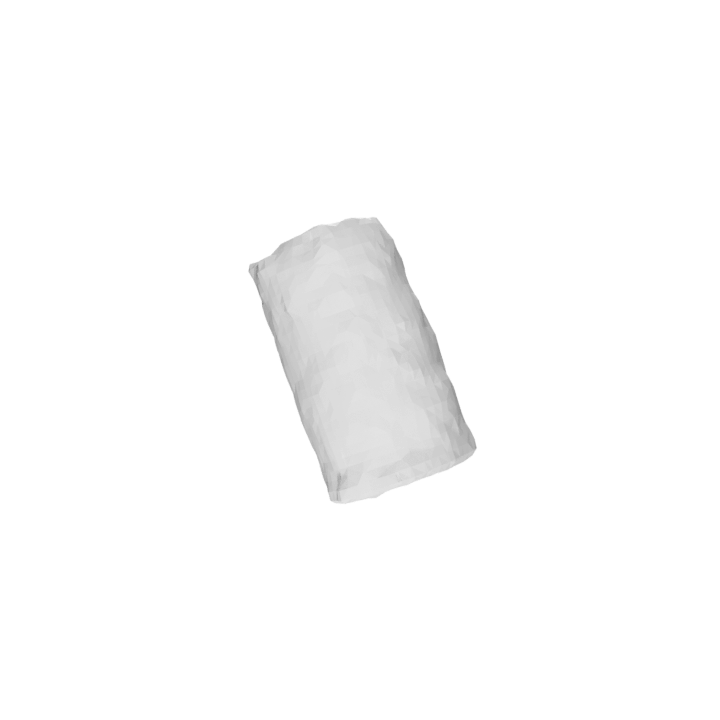} & \includegraphics[width=\psdfw]{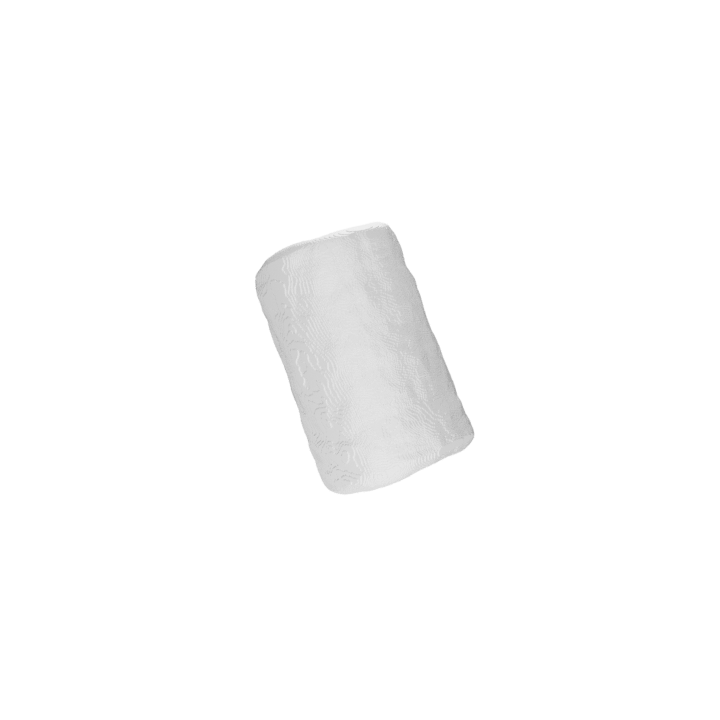} & \includegraphics[width=\psdfw]{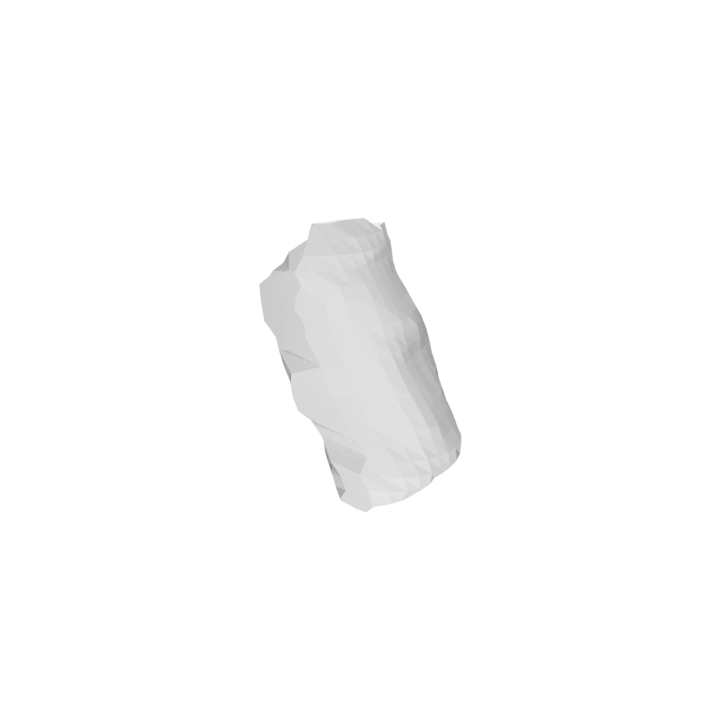}\\
\cline{2-5}
 & $180^{\circ}$ & \includegraphics[width=\psdfw]{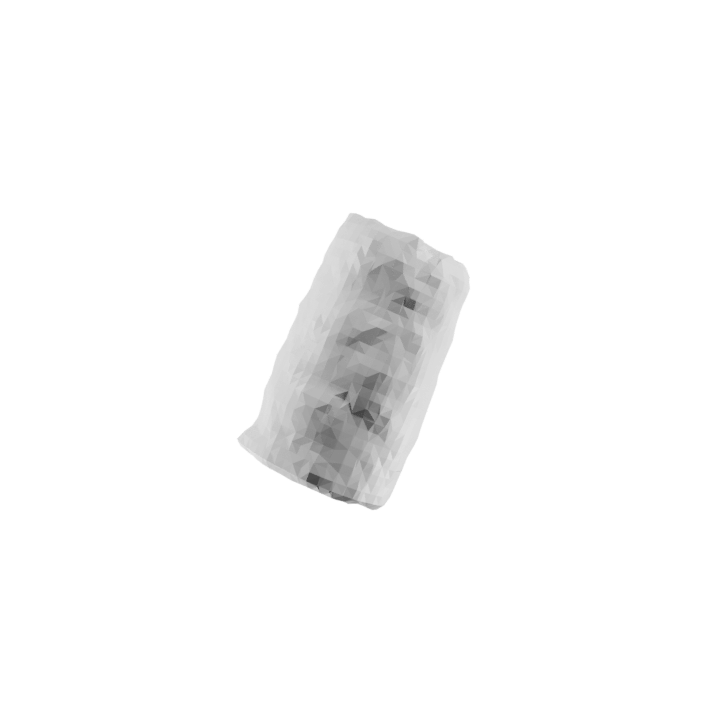} & \includegraphics[width=\psdfw]{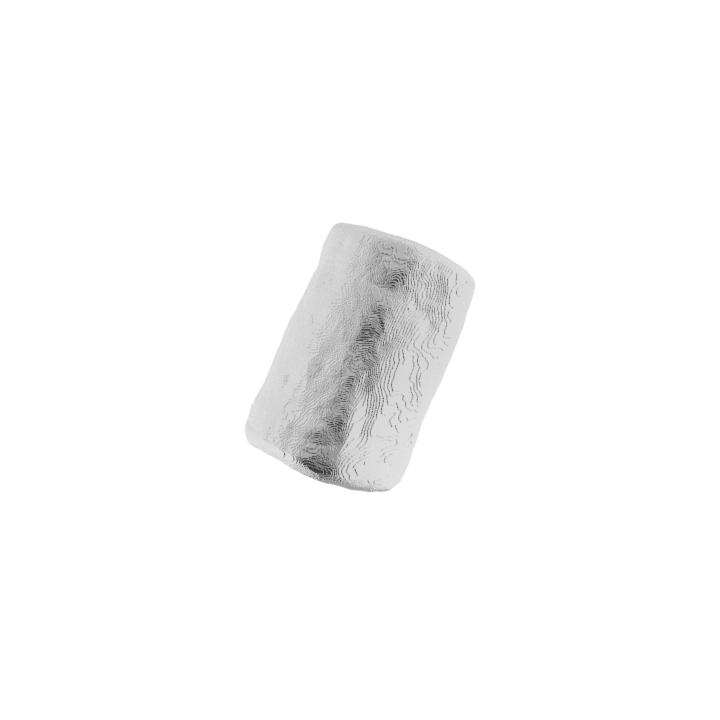} & \includegraphics[width=\psdfw]{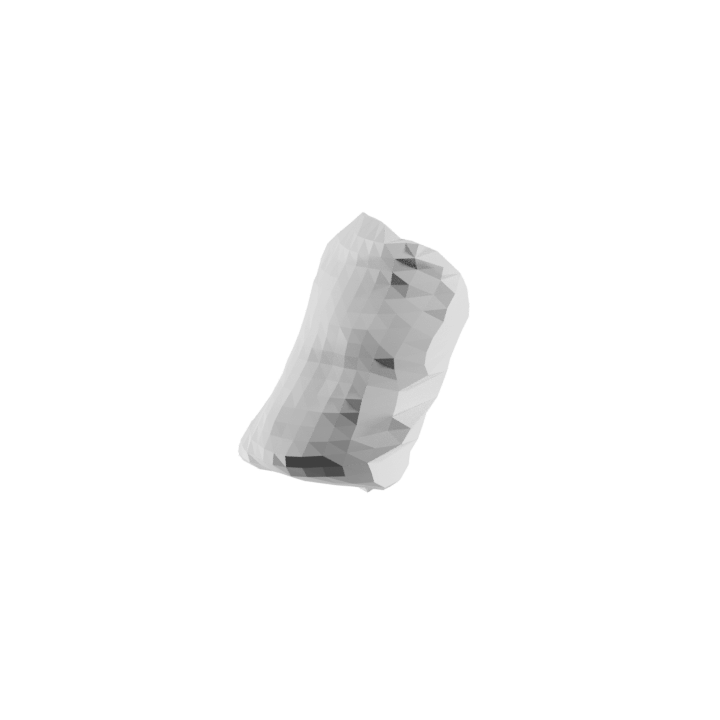}\\
\cline{2-5}
 & $270^{\circ}$ & \includegraphics[width=\psdfw]{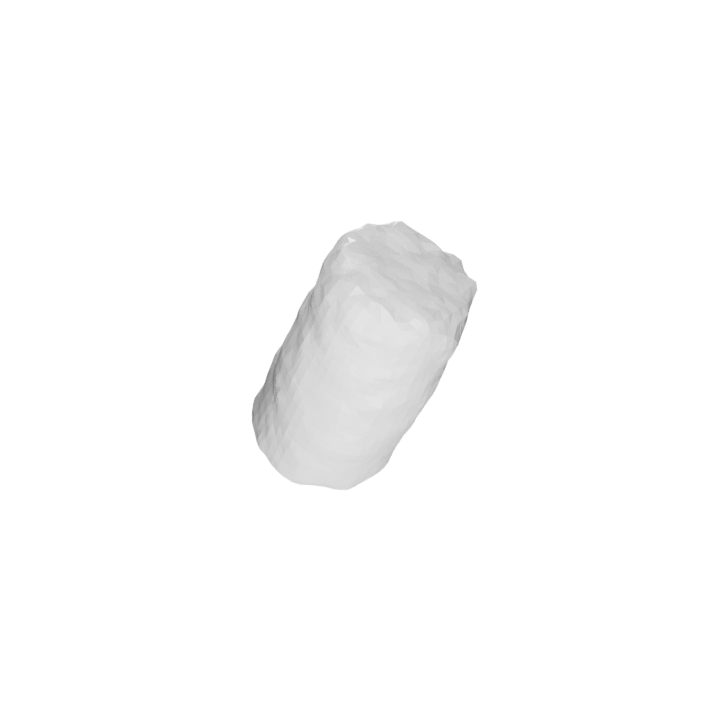} & \includegraphics[width=\psdfw]{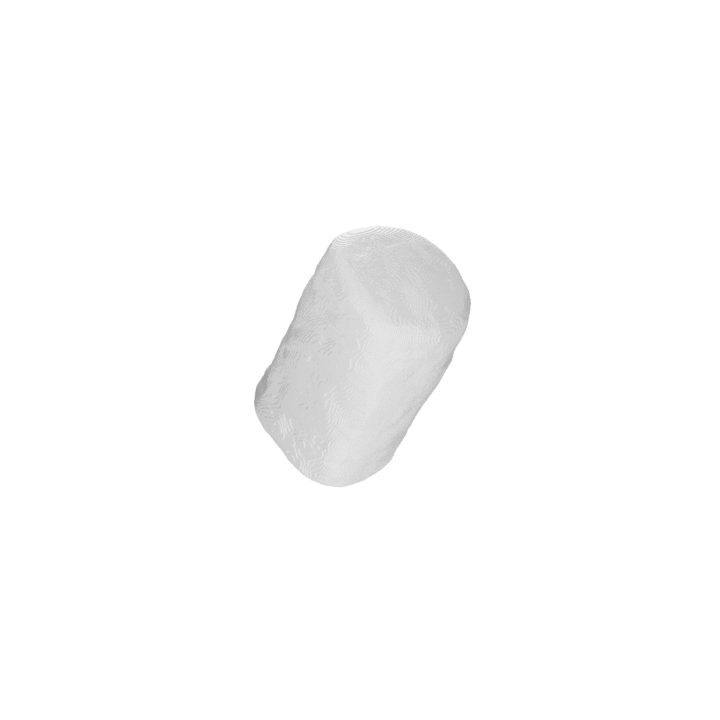} & \includegraphics[width=\psdfw]{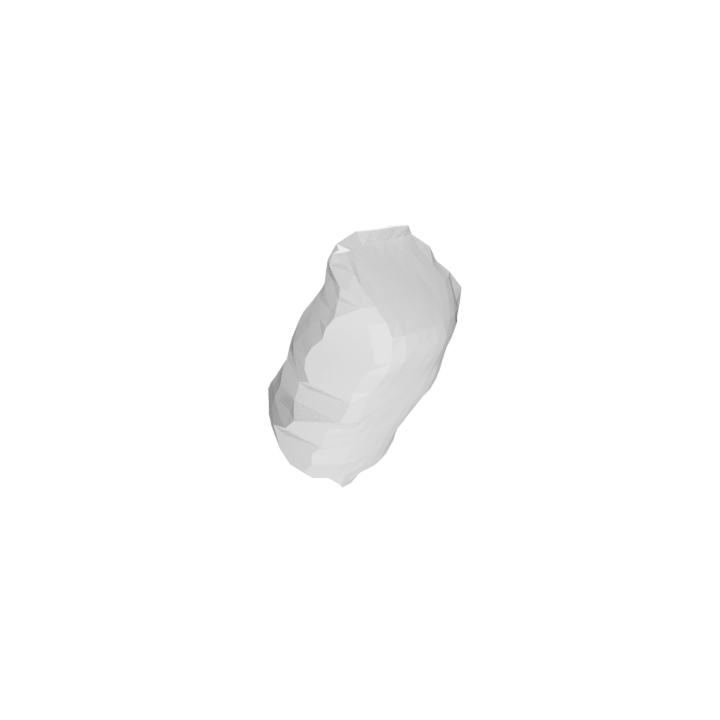}\\
\cline{2-5} & Chamfer &  & 5.83E-03 & 6.65E-03 \\
\hline

\end{tabular}
\end{table*}

\begin{table*}
\centering
\begin{tabular}{c|c|c|c|c} 
    \hline
     &   & \multicolumn{1}{c|}{GT} & \multicolumn{1}{c|}{PointSDF} & \multicolumn{1}{c}{Shell} \TBstrut \\ 
    \hline
    \multirow{4}{*}{ \rotatebox[origin=c]{90}{Bottle}} &  $90^{\circ}$ & \includegraphics[width=\psdfw]{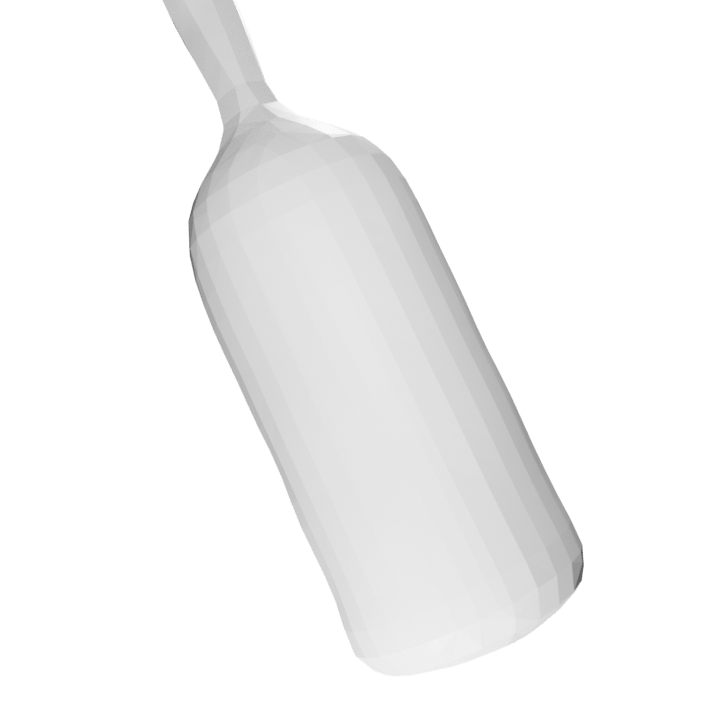} & \includegraphics[width=\psdfw]{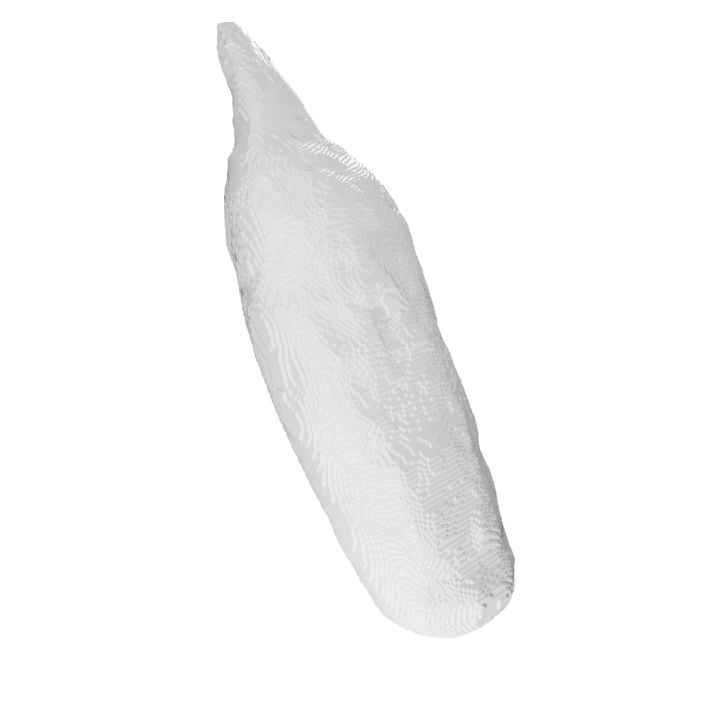} & \includegraphics[width=\psdfw]{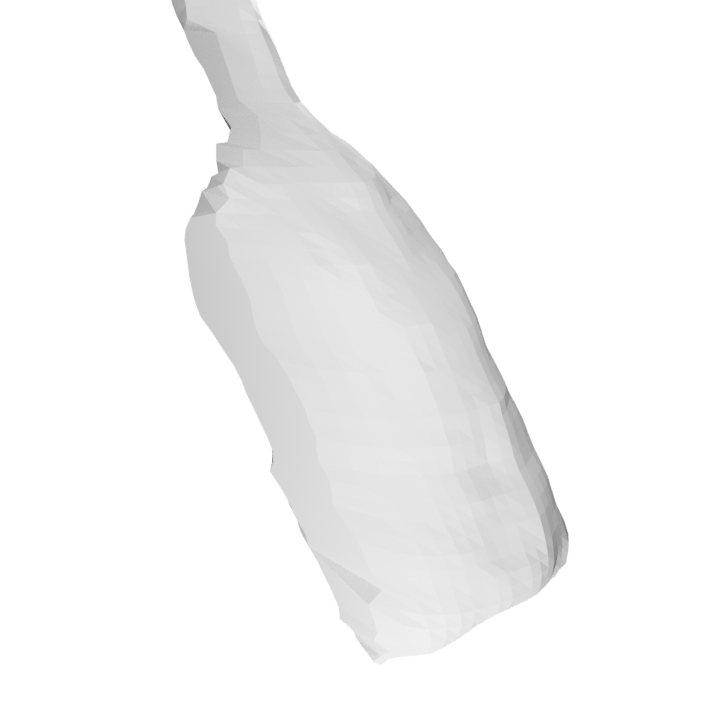}\\
\cline{2-5}
 &  $180^{\circ}$ & \includegraphics[width=\psdfw]{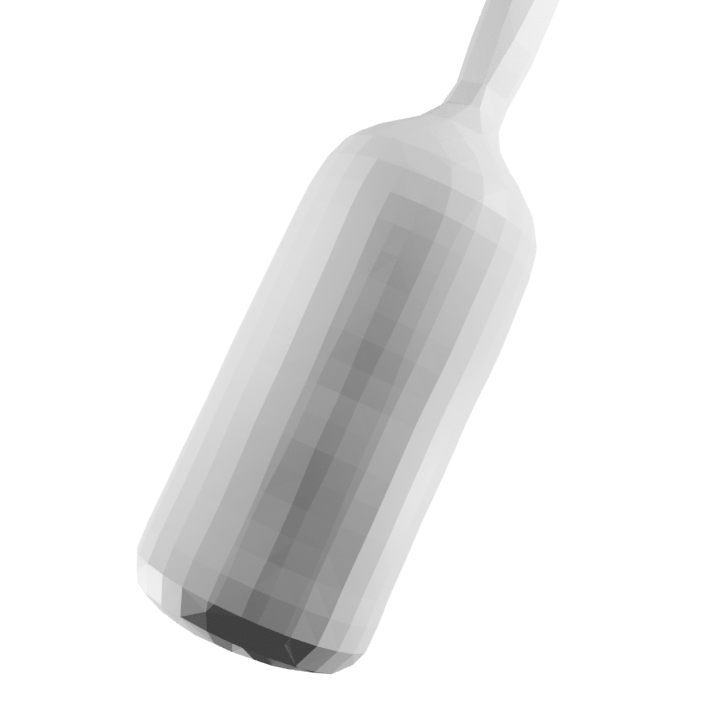} & \includegraphics[width=\psdfw]{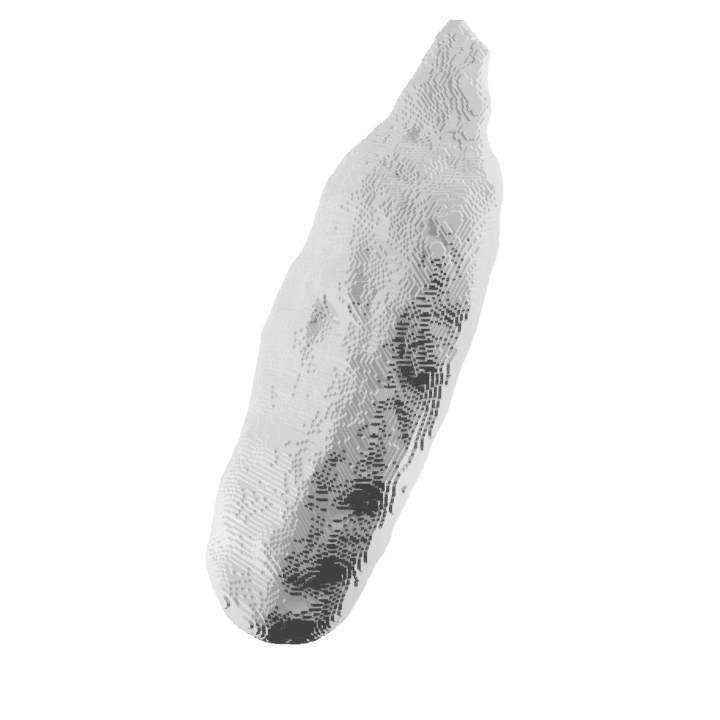} & \includegraphics[width=\psdfw]{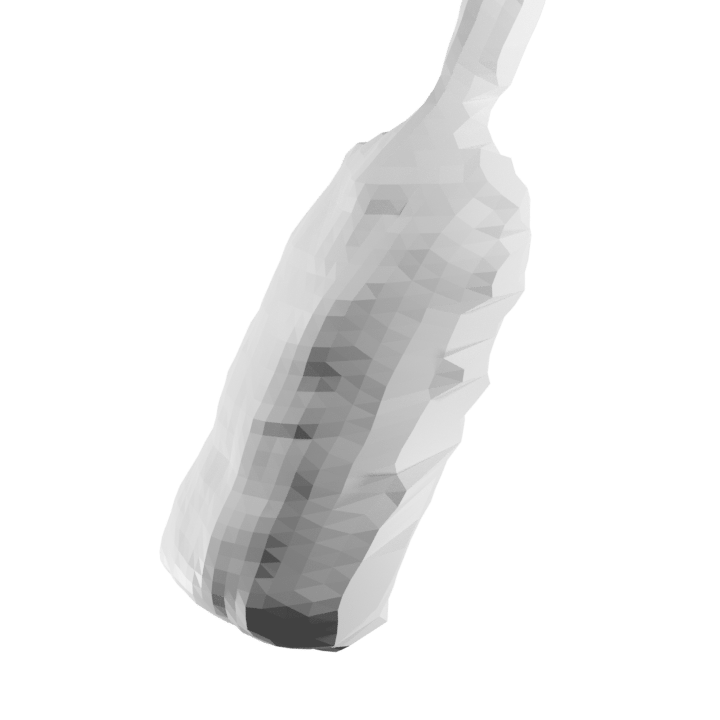}\\
\cline{2-5}
 &  $270^{\circ}$ & \includegraphics[width=\psdfw]{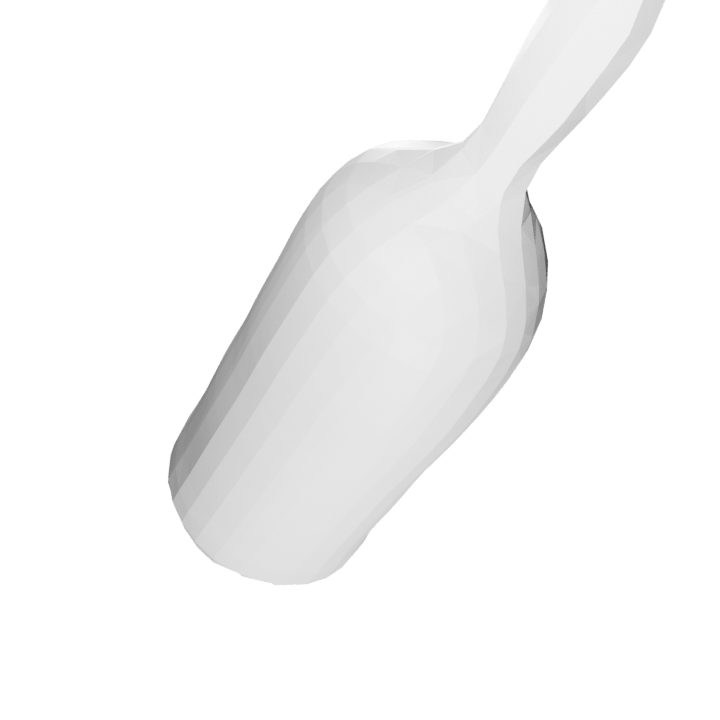} & \includegraphics[width=\psdfw]{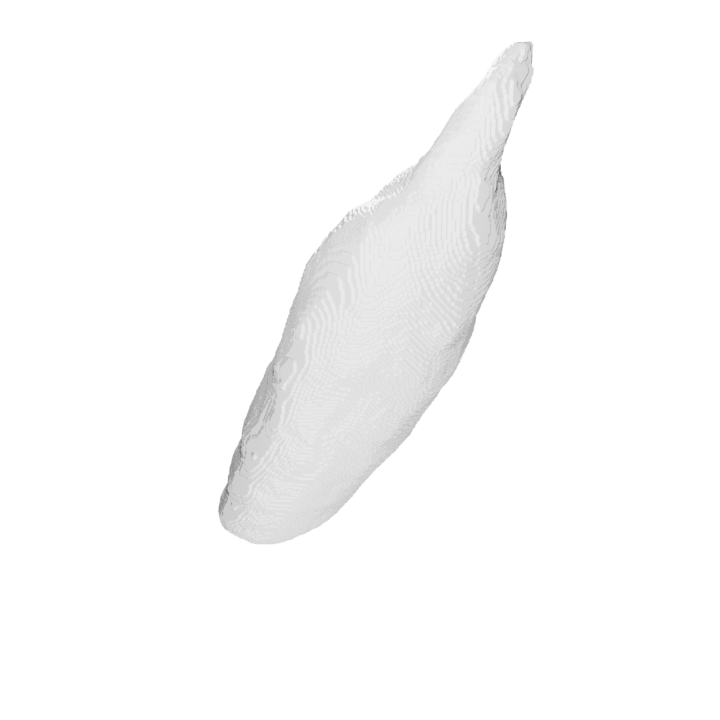} & \includegraphics[width=\psdfw]{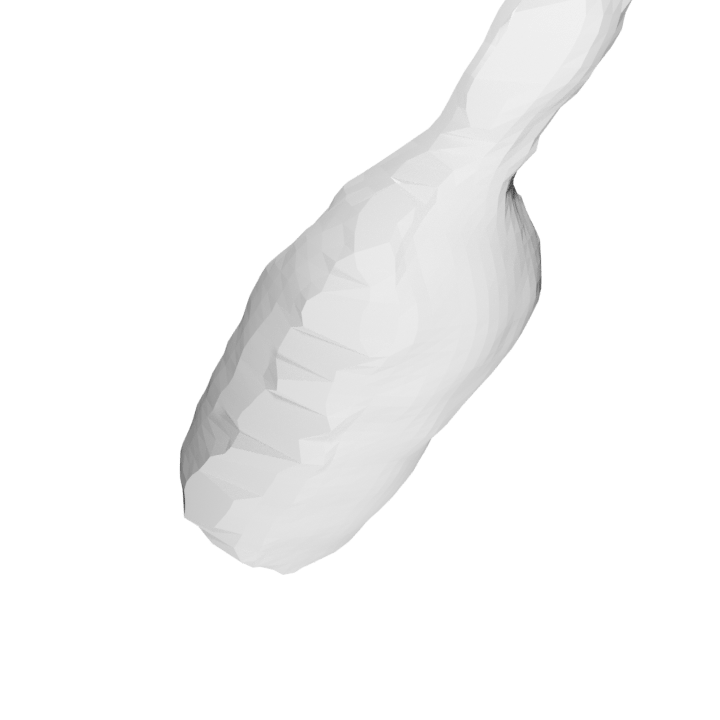}\\
\cline{2-5} & Chamfer &  & 3.10E-02 & 1.08E-02 \\
\hline

\end{tabular}
\end{table*}

\begin{table*}
    \centering
    \captionof{table}{\textbf{Qualitative comparison of object reconstructions from Shell and PointSDF methods for adversarial objects and object views.} Since these methods are not trained for specific classes of objects, they struggle to reconstruct the objects from a very limited information in the input. However, the shell reconstructions look qualitatively better since they do not loose the geometric details in the input.}
    \label{tab:shell_vs_ptsdf_adv}
    \begin{tabular}{l|l|l|l|l} 
        \hline
        & & Visible PCL & PointSDF & Shell  \\ 
        \hline
        \multirow{2}{*}{\rotatebox[origin=c]{90}{Clorox}} & View 0 & \includegraphics[height=\psdfw]{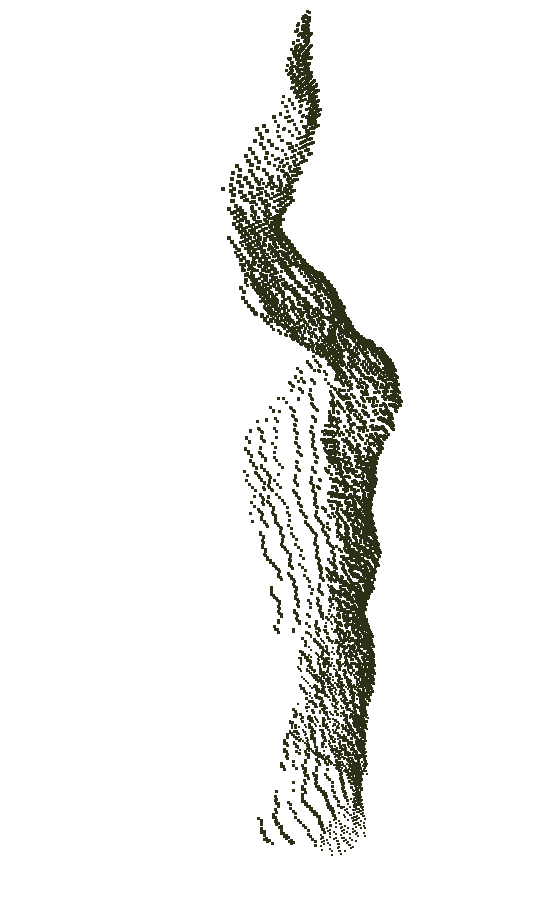} & \includegraphics[height=\psdfw]{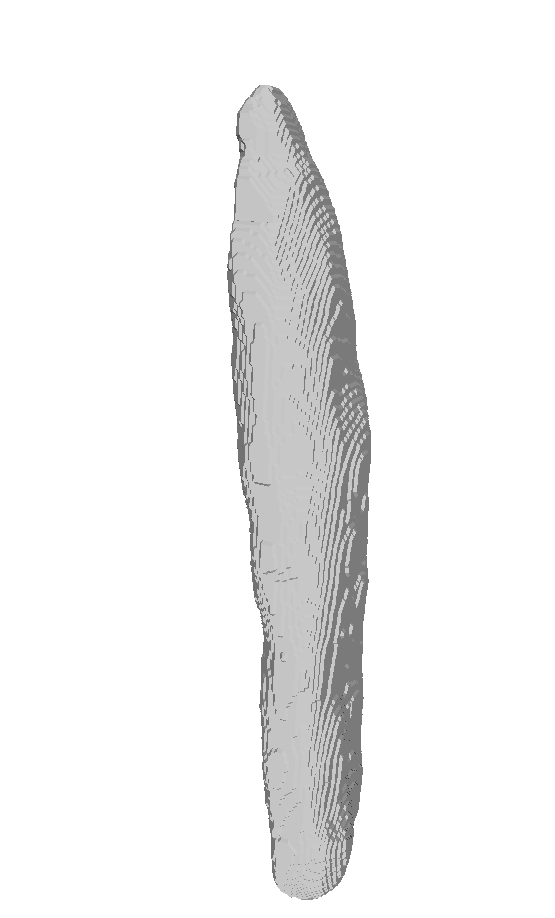} & \includegraphics[height=\psdfw]{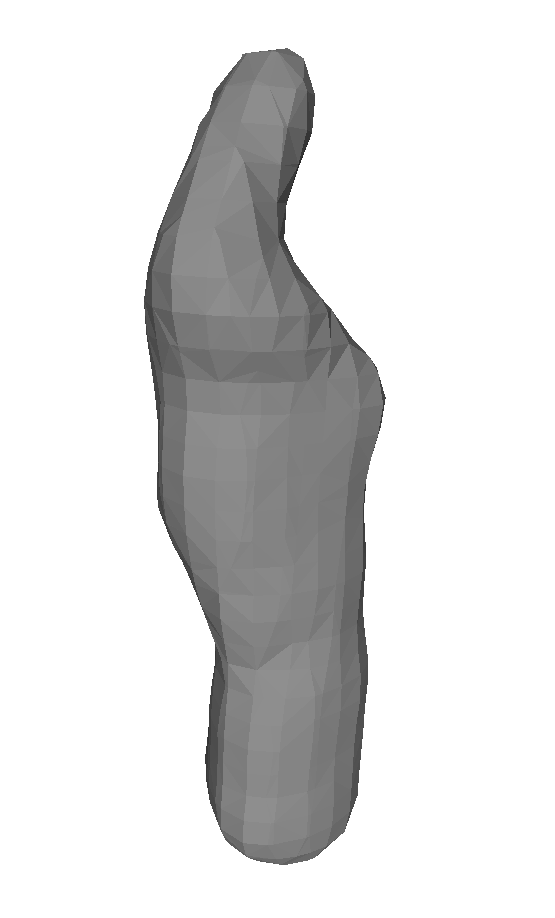} \\ 
        \cline{2-5} &  View 1 & \includegraphics[height=\psdfw]{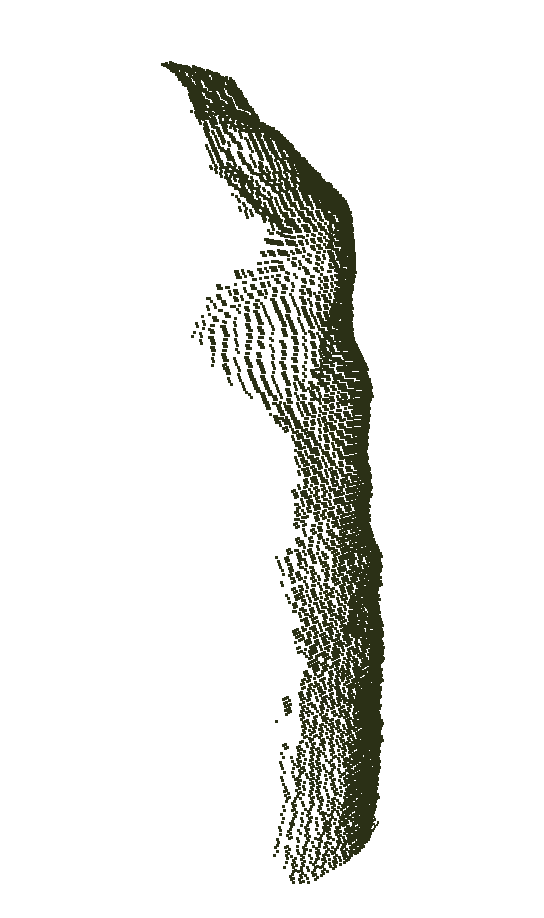} & \includegraphics[height=\psdfw]{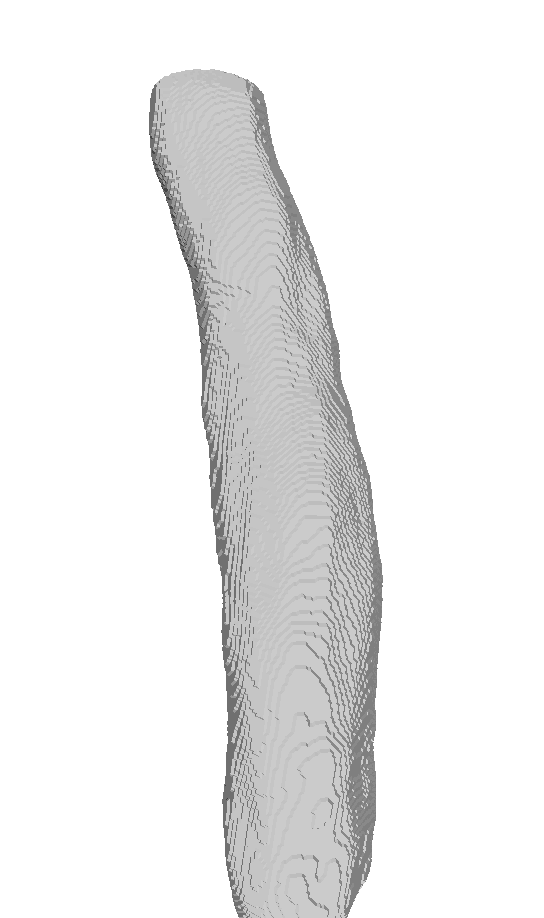} & \includegraphics[height=\psdfw]{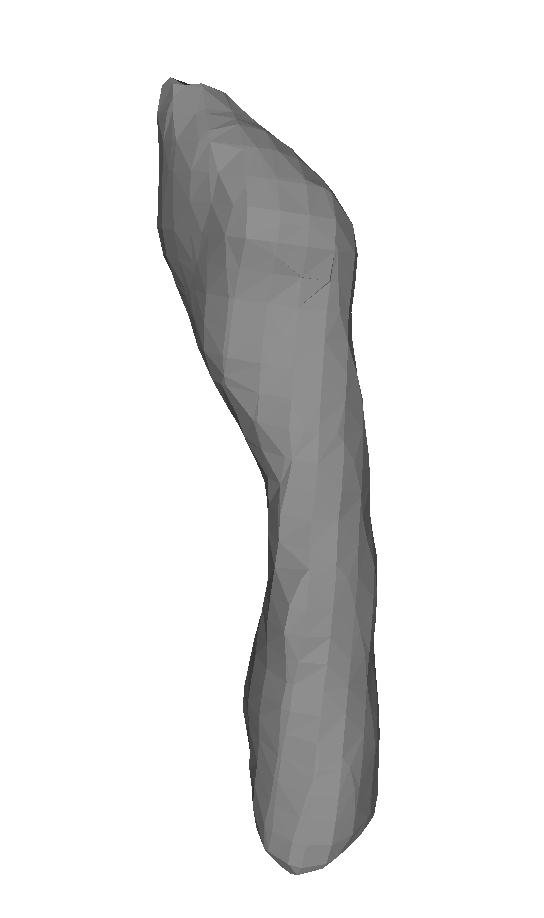}        \\
        \hline

        \multirow{2}{*}{\rotatebox[origin=c]{90}{Cup}} & View 0 & \includegraphics[height=\psdfw]{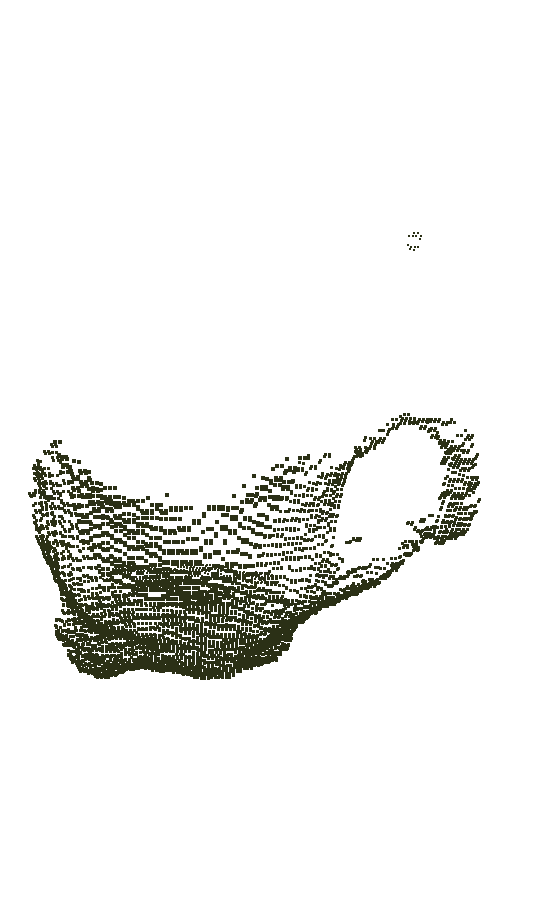} & \includegraphics[height=\psdfw]{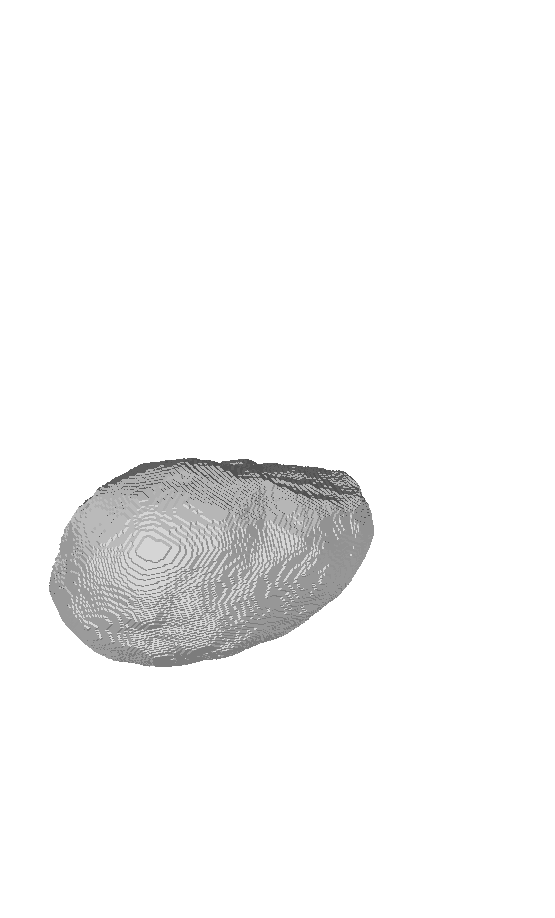} & \includegraphics[height=\psdfw]{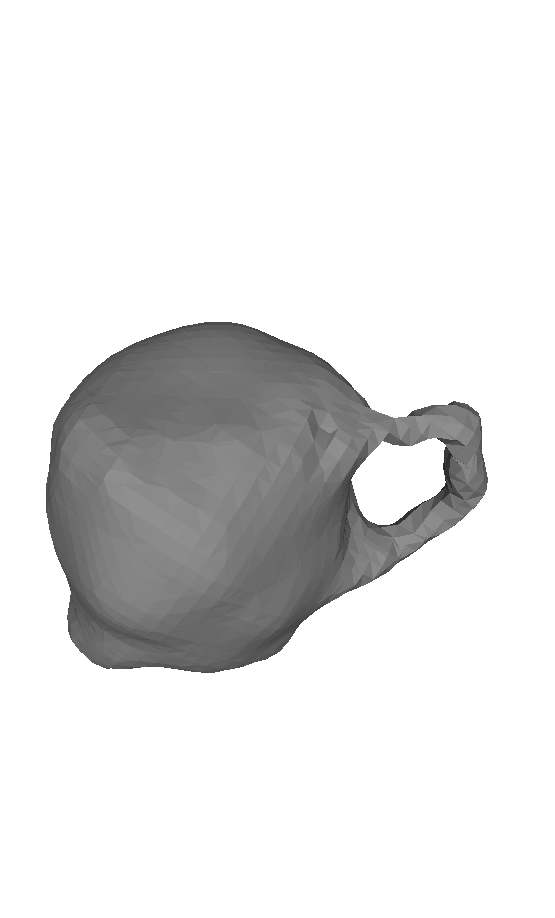} \\ 
        \cline{2-5} & View 1 & \includegraphics[height=\psdfw]{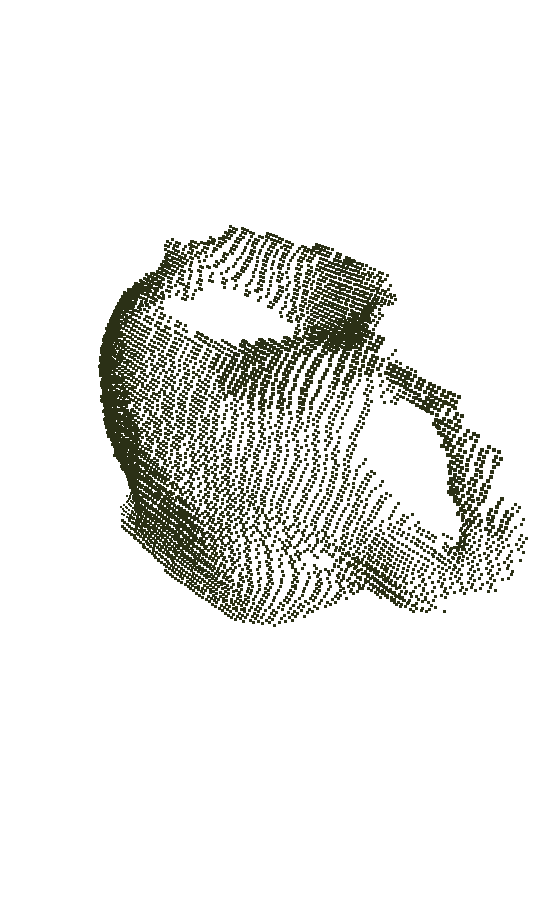} & \includegraphics[height=\psdfw]{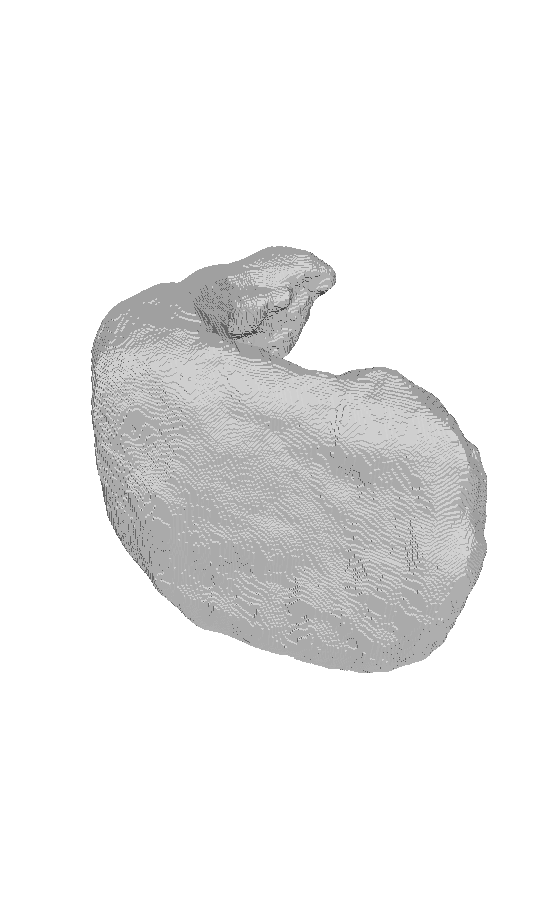} & \includegraphics[height=\psdfw]{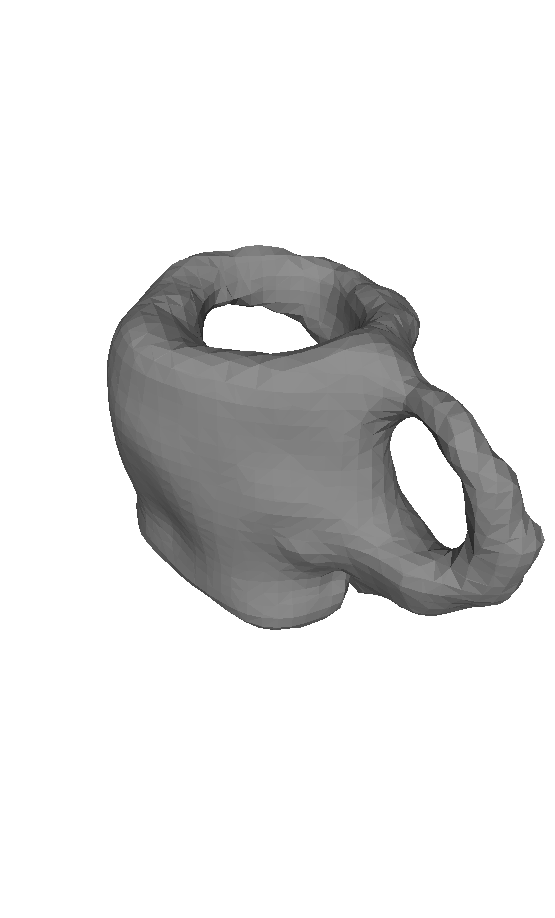}        \\
        \hline
        
    \end{tabular}
\end{table*}

\begin{table*}
    \centering
    \begin{tabular}{l|l|l|l|l} 
        \hline
        & & Visible PCL & PointSDF & Shell  \\ 
        \hline
        \multirow{2}{*}{\rotatebox[origin=c]{90}{Weiman}} & View 0 & \includegraphics[height=\psdfw]{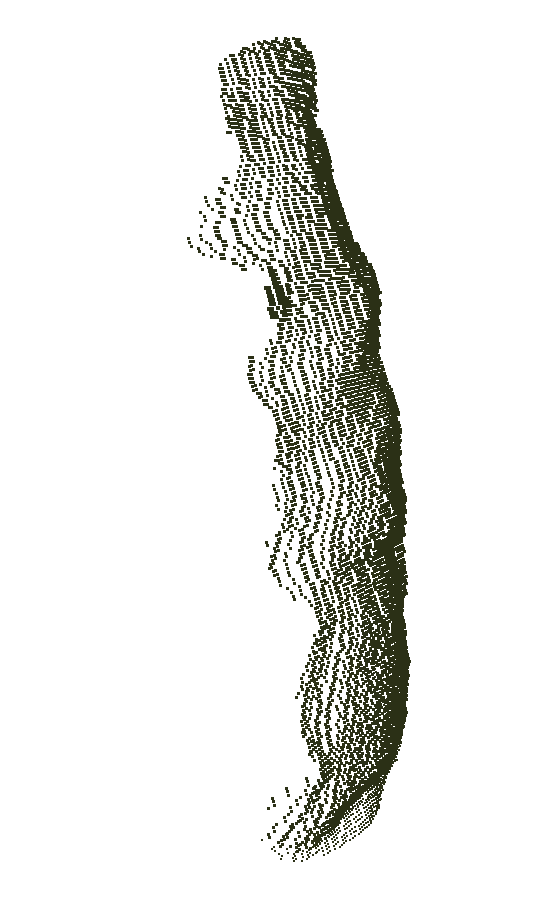} & \includegraphics[height=\psdfw]{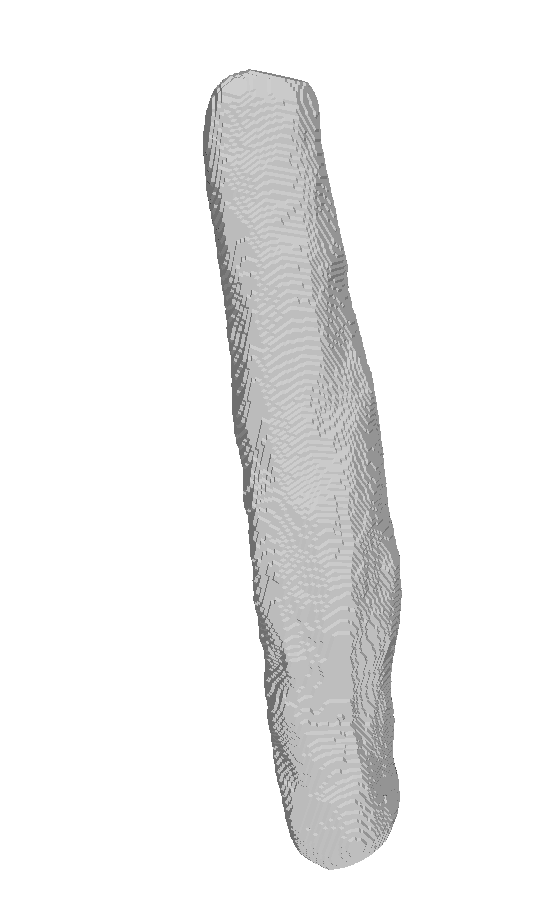} & \includegraphics[height=\psdfw]{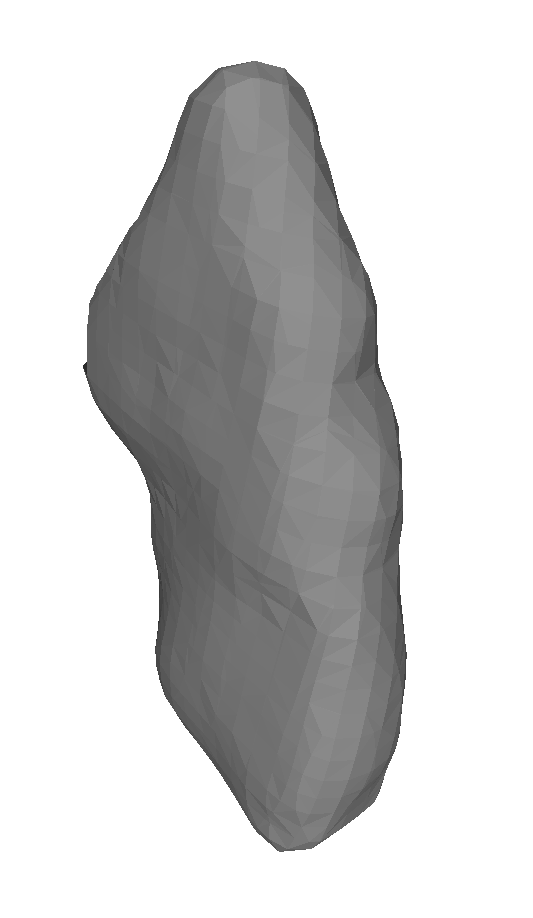} \\ 
        \cline{2-5} & View 1 & \includegraphics[height=\psdfw]{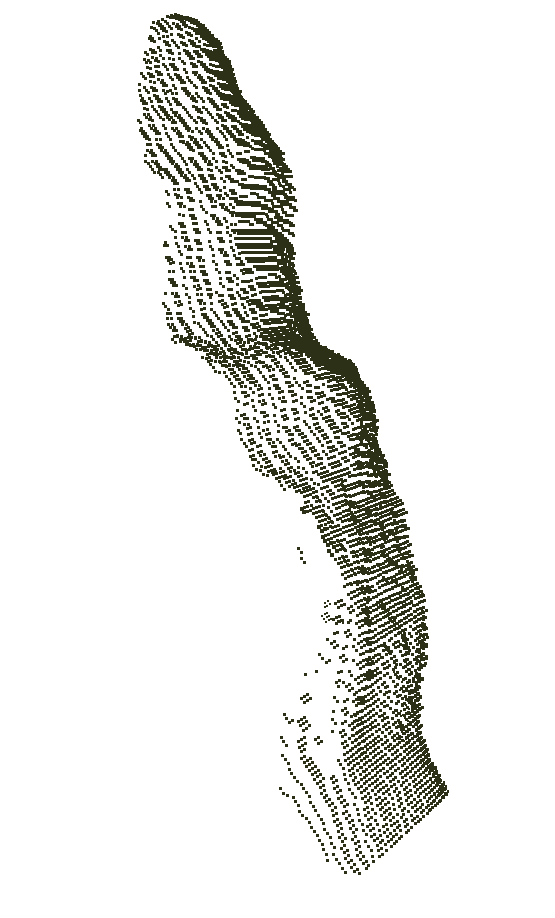} & \includegraphics[height=\psdfw]{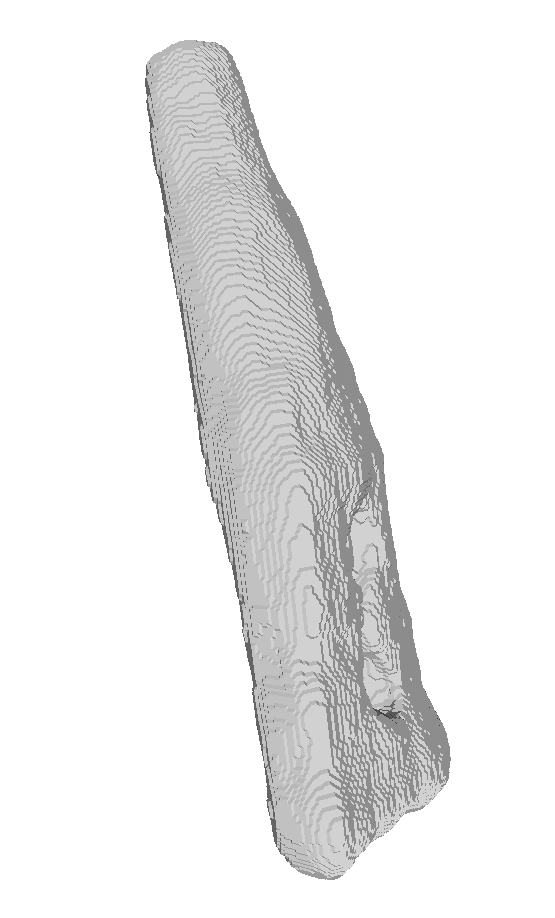} & \includegraphics[height=\psdfw]{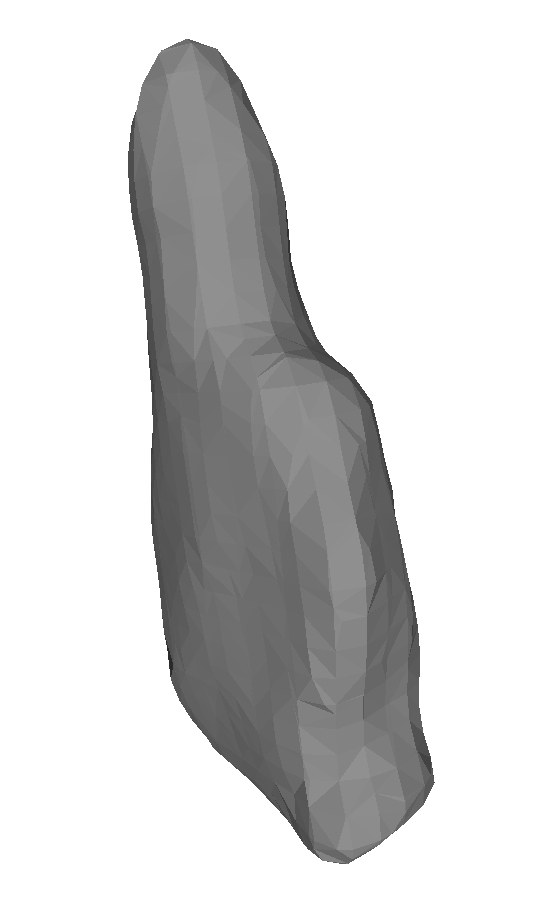}        \\
        \hline
        
        \rotatebox[origin=c]{90}{Cheeze-it} & View 0 & \includegraphics[height=\psdfw]{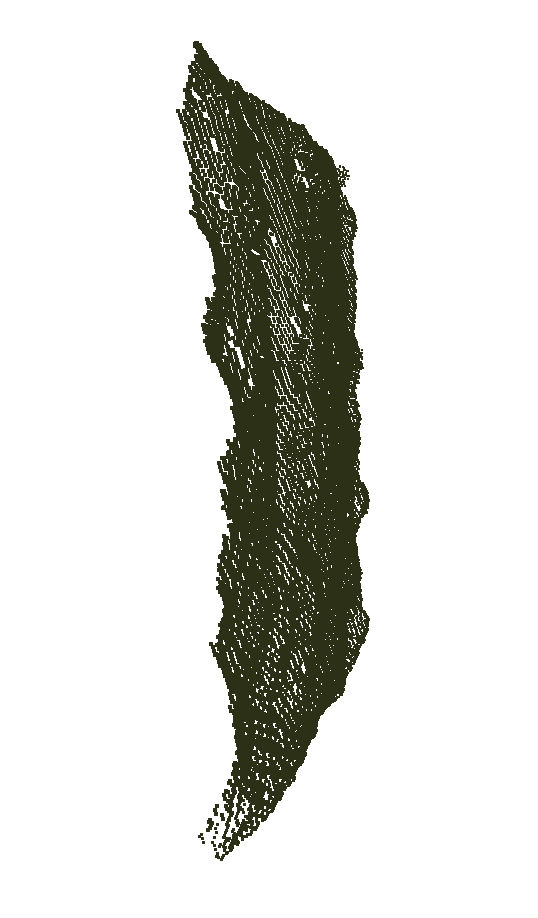} & \includegraphics[height=\psdfw]{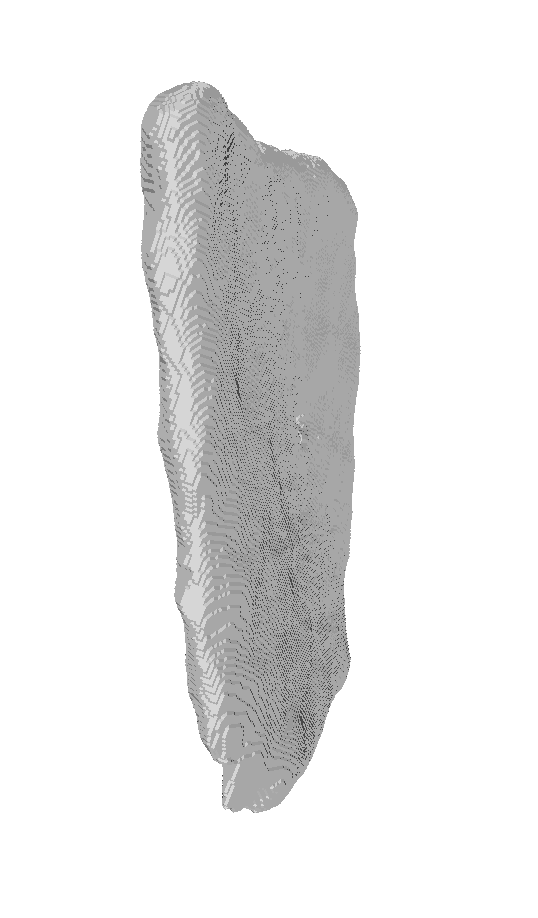} & \includegraphics[height=\psdfw]{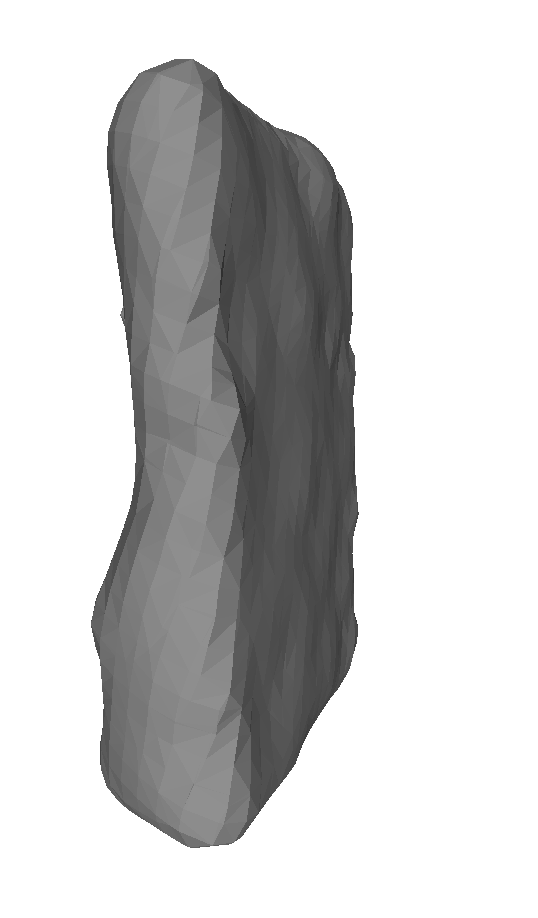} \\
        \hline

        \rotatebox[origin=c]{90}{ Ziploc} & View 0 & \includegraphics[height=\psdfw]{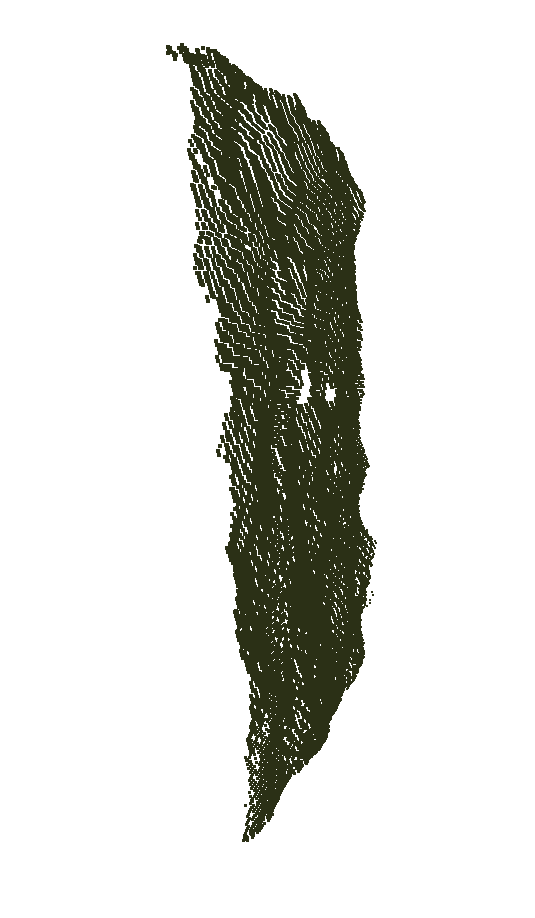} & \includegraphics[height=\psdfw]{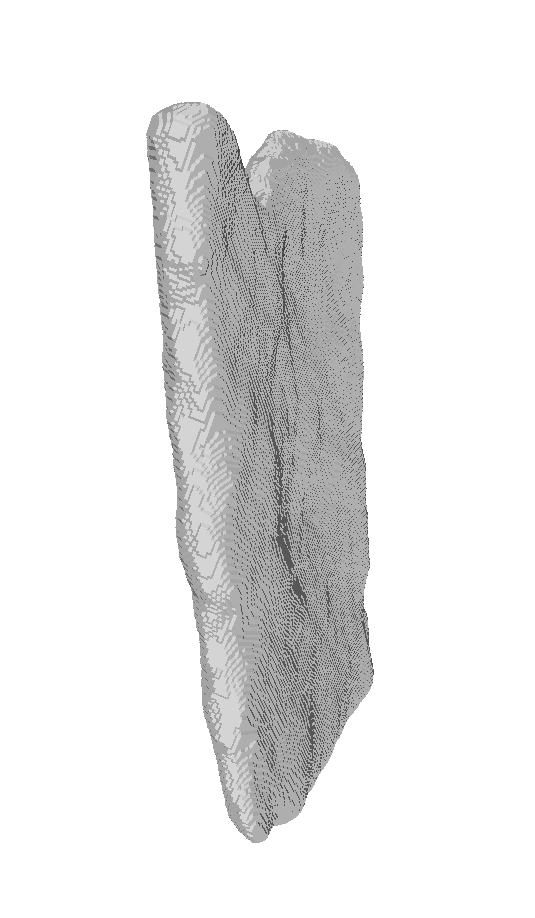} & \includegraphics[height=\psdfw]{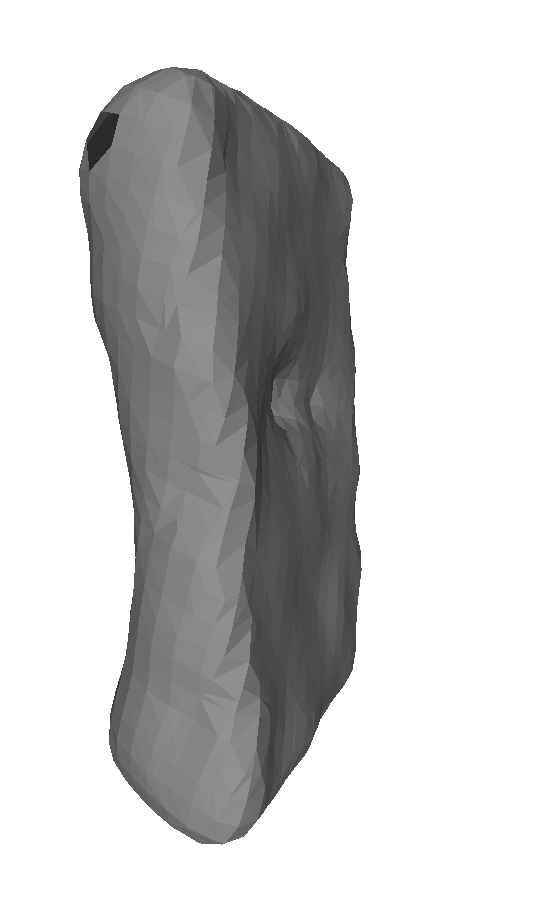} \\ 
        \hline

    \end{tabular}
\end{table*}

\end{document}